\documentclass{article}

\usepackage{PRIMEarxiv}

\usepackage[utf8]{inputenc} 
\usepackage[T1]{fontenc}    
\usepackage{hyperref}       
\usepackage{url}            
\usepackage{booktabs}       
\usepackage{amsfonts}       
\usepackage{nicefrac}       
\usepackage{microtype}      
\usepackage{lipsum}
\usepackage{fancyhdr}       
\usepackage{graphicx}       

\usepackage{amssymb}
\usepackage{amsthm}
\usepackage{amsmath} 
\usepackage{longtable}
\usepackage{natbib}

\usepackage{lineno}
\usepackage{url} 
\usepackage{booktabs,tabularx}
\usepackage{multirow}
\usepackage{tabularray}
\usepackage{xcolor}
\usepackage{soul}
\usepackage{tcolorbox}
\usepackage[tight,footnotesize]{subfigure}
\usepackage{subcaption}
\usepackage{stfloats}
\usepackage{setspace}
\usepackage{adjustbox}
\usepackage{enumitem}

\usepackage{lmodern}

\usepackage{cleveref}

\newcommand{\HSCBM}{{SCBM}}
\newcommand{\HSCBMT}{{SCBMT}}
\newcommand{\textSet}{{\mathcal{D}}}
\newcommand{\conceptSet}{{\mathcal{C}}}
\newcommand{\adjSet}{{\mathcal{A}}}
\newcommand{\M}{{\mathbf{M}}}
\newcommand{\Q}{{\mathbf{Q}}}
\newcommand{\K}{{\mathbf{K}}}
\newcommand{\V}{{\mathbf{V}}}
\newcommand{\T}{{\mathcal{T}}}
\newcommand{\token}{{\psi}}
\newcommand{\LLM}{{L_\theta}}
\newcommand{\cVec}{\mathbf{v}}
\newcommand{\z}{\mathbf{z}}
\newcommand{\y}{\mathbf{y}}
\newcommand{\haty}{\mathbf{\hat{y}}}
\newcommand{\Response}{{\mathcal{O}_{\LLM}^+}}
\newcommand{\tildeResponse}{\overline{\mathcal{O}}_{\LLM}^+}
\newcommand{\paratitle}[1]{\smallskip\noindent\textbf{#1}}
\definecolor{highlightblue}{HTML}{C4FCEF}
\colorlet{highlightblue}{highlightblue!90}

\definecolor{highlightred}{HTML}{FF6F91}
\colorlet{highlightred}{highlightred!50}

\definecolor{cs_color}{HTML}{005B44}
\definecolor{hs_color}{HTML}{845EC2}
\definecolor{ns_color}{HTML}{4B4453}

\colorlet{mygray}{gray!30}

\sethlcolor{highlightred} 
\newcommand{\redhighlight}[1]{\sethlcolor{mygray}\hl{#1}}
\sethlcolor{highlightblue} 

\newcommand{\revised}[1]{\textcolor{black}{#1}}

\def\eg{e.g.,}
\def\ie{i.e.,}

\title{Distilling Knowledge from Large Language Models: A Concept Bottleneck Model 
for Hate and Counter Speech Recognition}

\author{
  Roberto Labadie-Tamayo\textsuperscript{1}\thanks{These authors contributed equally to this work. Corresponding authors: \texttt{\{roberto.labadie-tamayo,djordje.slijepcevic\}@fhstp.ac.at}} \\
  \And
  Djordje Slijepčević\textsuperscript{1}$^{*}$ \\
  \And
  Xihui Chen\textsuperscript{1} \\
  \And
  Adrian Jaques Böck\textsuperscript{1} \\
  \And
  Andreas Babic\textsuperscript{1} \\
  \And
  Liz Freimann\textsuperscript{2} \\
  \And
  Christiane Atzmüller\textsuperscript{2} \\
  \And
  Matthias Zeppelzauer\textsuperscript{1}
  \\
  \textsuperscript{1}St.Pölten University of Applied Sciences, Institute of Creative Media Technologies, St.Pölten, Austria \\
  \texttt{\{roberto.labadie-tamayo,djordje.slijepcevic,xihui.chen, adrian.boeck,}\\
  \texttt{andreas.babic,matthias.zeppelzauer\}@fhstp.ac.at} \\
  \\
  \textsuperscript{2}University of Vienna, Department of Sociology, Vienna, Austria \\
  \texttt{\{liz.freimann,christiane.atzmueller\}@univie.ac.at}
}

\begin{document}
\maketitle

\begingroup
\renewcommand\thefootnote{}\footnotetext{
This is a preprint of a manuscript accepted for publication in \textit{Information Processing \& Management} (Elsevier). The final version will be available via the journal's website.}
\addtocounter{footnote}{-1}
\endgroup

\begin{abstract}
The rapid increase in hate speech on social media has exposed an unprecedented 
impact on society, making automated methods for detecting such content important. 
Unlike prior black-box models,  we propose a novel transparent method for 
automated hate and counter speech recognition, 
\ie~``Speech Concept Bottleneck Model'' (\HSCBM),  using 
adjectives as human-interpretable bottleneck concepts. 
\HSCBM~leverages large language models (LLMs) to map input texts to an 
abstract adjective-based representation, which is then sent to  
a light-weight classifier for downstream tasks. 
Across five benchmark datasets spanning multiple languages and platforms (\eg~Twitter, Reddit, YouTube), \HSCBM~achieves an average macro-F1 score of 
$0.69$ which outperforms the most recently reported results from the 
literature on four out of five datasets.
Aside from high recognition accuracy, \HSCBM~provides a high level of both local and global interpretability.
Furthermore, fusing our adjective-based concept representation with 
transformer embeddings, leads to a 1.8\% performance increase on average 
across all datasets, showing that the proposed representation captures complementary information.
Our results demonstrate that adjective-based concept representations can serve 
as compact, interpretable, and effective encodings for hate and counter speech recognition. With adapted adjectives, our method can also be applied to 
other NLP tasks. 
\end{abstract}



\section{Introduction}
Online hate speech has rapidly increased on social media platforms, including even mainstream ones like X, Instagram, and 
TikTok, and has become a prevalent social problem~\citep{guillaume2022hate,MathewDGM19,MasudBKA22,MasudDMJGD021}. 
This new form of online abuse negatively impacts a wide range of populations, not only causing emotional harm but also inciting social discrimination 
and even physical criminal violence~\citep{BS20,RHMS2023,MasudBKA22}. 
As a response, mechanisms such as manual content moderation have been implemented. 
However, the large volume of online discourse makes this approach 
resource-intensive, time-consuming, and hardly feasible in practice. 
Furthermore, prolonged exposure to 
harmful content can adversely affect the mental health of 
moderators~\citep{cobbe_algorithmic_2021}. Consequently, there is a pressing need for reliable automated recognition of toxic content~\citep{MasudBKA22,SunOUW24}. 
Community-driven initiatives like counter speech represent an important means to combat hate speech by providing responses to hateful content, such as offering facts, warning of potential consequences, and promoting empathy. Automatically identifying counter speech is important for increasing its visibility and thereby encouraging online civil courage.

Analysing online hate and counter speech presents unique challenges 
in natural language processing (NLP) due to its nuanced and context-dependent 
nature~\citep{yu-etal-2022-hate}. 
Recent advancements in NLP, particularly through the use of  transformers and large language models (LLM), such as 
BERT~\citep{devlin-etal-2019-bert,zhou2023comprehensive} and GPT~\citep{kalyan2023survey}, 
have led to significant improvements in the accuracy of hate speech 
recognition~\citep{LeeNSSP22, vidgen-etal-2020-detecting}. 
These models are trained on large-scale datasets, which enhances their ability to model context and capture nuanced language. As a result, they can be applied in a zero-shot setting or fine-tuned with annotated data for specific downstream tasks, further optimizing their performance for particular applications~\citep{LiCKYCJT23}.
However, a key limitation of these models is their high complexity and black-box nature, which results in a lack of transparency. While there have been attempts to explain them through explainable artificial intelligence (XAI) approaches, these explanations are often questionable, revealing potential biases and spurious correlations~\citep{wilming2024gecobenchgendercontrolledtextdataset}.
Especially post-hoc explanations~\citep{DalviDSBBG19,DurraniSDB20}, which are derived from pre-trained models, have been shown to sometimes provide misleading results~\citep{slack2020fooling}. For this reason, more recent efforts have moved towards providing explanations rooted in the inherent mechanisms of LLMs, such as assessing attention values of tokens, 
exploring counterexamples, or generating free-text explanations~\citep{HanWT20,DasGKLL22}. These approaches, however, primarily provide \emph{local} explanations for individual input texts and fail to capture the global reasoning principles of the models.

In contrast, \emph{concept bottleneck models} (CBMs)~\citep{KohNTMPKL20}, 
initially developed for vision tasks~\citep{OikarinenDNW23,YangPZJCY23}, offer both global and local explanations. 
CBMs achieve this by defining a set of high-level, human-interpretable 
\emph{concepts} and leveraging LLMs to encode input texts through these concepts. 
This encoding forms a \emph{bottleneck layer}, which serves as input for downstream models, typically implemented as a stack of dense layers on top of it. 
While CBMs have been widely explored in computer vision tasks, their application in NLP remains widely unexplored, with only a handful of pioneering studies~\citep{SunOUW24,LudanLYD24}. 
This gap is particularly evident for hate and counter speech recognition.


\paratitle{Objectives and contribution.}
In this paper, we aim to \emph{develop an effective and efficient human interpretable approach based on CBMs for detecting hate and counter speech}.
The approach leverages the rich text modelling capabilities of LLMs while simultaneously ensuring human interpretability and repeatability in the resulting representation. To achieve this, we address three key challenges commonly encountered in CBM-based approaches:
\begin{itemize}[leftmargin=*]
\item \textit{Defining format and semantics of the bottleneck  concepts.} \newline Existing CBM approaches in NLP typically extract concepts from LLM responses to zero-shot or in-context learning prompts~\citep{LudanLYD24} (\eg~for the prompt ``How would you describe the film [`film name']?'' extracted concepts might include terms 
like ``acting'', ``thrilling'', or ``emotional''). 
While these methods work well for domains such as music, films, or restaurant reviews, they are less suited for hate speech and counter speech recognition, which typically involves more subtle nuances, contextual dependencies, as well as social and cultural considerations. Furthermore, the concepts extracted by these methods can represent abstract or unrelated semantic aspects, making them less interpretable for human users.

\item \textit{Computing concept representations of a
given input text.} \newline While LLMs are frequently used for this task, their responses are conditioned by their inherent nature of randomness, even when presented with identical prompts. A desirable property for constructing semantic representations is repeatability and deterministic behaviour.

\item \textit{Mapping the computed bottleneck embeddings 
to predictions.} \newline Existing approaches typically apply a dense layer that utilises all extracted concepts~\citep{YangPZJCY23,SunOUW24,LudanLYD24}, which leads to over-complex explanations for a given prediction~\citep{miller2019explanation} hindering interpretability
~\citep{RKFR23}. 
However, a few key determinant concepts are usually sufficient to make a prediction.
\end{itemize}


To address these challenges, we introduce the “Speech Concept Bottleneck Model” (\HSCBM). 
This model provides a high level of interpretability while achieving performance comparable to, or even surpassing, state-of-the-art methods. 
To capture the complexities (e.g., contextual dependencies) associated with hate speech and counter-speech in a more transparent way, we introduce for the first time the use of adjectives as bottleneck concepts. Adjectives are particularly well-suited for this purpose, as they offer a descriptive linguistic layer that aligns naturally with how people interpret language, including emotional tone, intent, and attitude that shape how a message is perceived by the reader. By grounding the model’s internal reasoning in such interpretable linguistic units, we enable more transparent model behaviour and explanations that are easier for users to understand and evaluate.
These adjective-based concepts provide an intuitive way for users to understand the emotional cues that support or oppose the recognition and classification of online hate and counter speech.

To handle the stochastic nature of LLM responses, we propose a probabilistic approach with deterministic behaviour to assess the relation between selected adjectives and input texts. 
Finally, we design \HSCBM~as a light-weight classifier to map concept representations to 
predictions, explicitly modelling the varying importance of adjectives 
for different classes. This ensures that key determinant concepts are prioritised by the classifier, enabling sparse and focused explanations while maintaining 
computational efficiency.

\noindent
Our main contributions are summarised as follows:
\begin{itemize}[leftmargin=*]
    \item We propose \HSCBM, a model designed to recognise hate speech and counter speech by integrating 
    a bottleneck layer of concepts represented as \emph{adjectives}. 
    Leveraging adjectives provides an expressive representation of the emotions and intentions exhibited in hate and counter speech. 
    Encoding texts with adjectives can effectively complement traditional transformer-based approaches, resulting in improved performance. 
    \item We propose a novel method to harness the inherent stochasticity of LLMs -- 
    a limitation often overlooked in prior work -- by mapping input texts into 
    reproducible probabilistic representations. This approach is generalisable to scenarios where LLMs are used to construct input encodings.
    \item We develop a light-weight classifier tailored to hate and counter speech recognition that integrates a masking function to learn the importance of individual adjectives, enabling more focused and human-interpretable explanations compared to the state-of-the-art.
    \item We conduct comprehensive experiments to empirically evaluate the proposed \HSCBM~model. Our results show that it achieves state-of-the-art performance and, at the same time, yields transparency at local and global levels by identifying relevant subsets of adjectives for the respective downstream tasks. 
\end{itemize}

\paratitle{The structure of the paper.}
We give a brief overview of related works and emphasise the identified
challenges in Section~\ref{sec:related work}.
We describe our methodology in Section~\ref{sec:method} and 
present our experiments and results in Section~\ref{sec:experiments}. We conclude our findings in Section~\ref{sec:conclusion}.

\section{Related Work}
\label{sec:related work}

Our research is located at the intersection of different research topics and fields, briefly reviewed in the following.

\paratitle{Hate speech and counter speech recognition.}
Hate speech recognition, a classic NLP task, has gained significant importance in recent years. 
Early approaches relied on traditional statistical methods, such as 
bag-of-words~\citep{PawarAJGR18} and tf-idf~\citep{10605280}. 
Despite their computational efficiency, they
lack the ability to capture context and abstract semantics. These limitations have been
addressed with deep learning models, such as LSTMs~\citep{OusidZSY19}, 
and later transformer-based models like BERT~\citep{WangLOS20,Saleh31122023}. 
These models advanced NLP by capturing complex semantics and temporal dependencies. Their success primarily lay in the availability of large annotated 
datasets~\citep{JahanO23}.

Counter speech recognition has emerged as an increasingly relevant area of research commonly linked to hate speech and toxic language detection. Related works range from traditional approaches that utilise conventional representations, \eg~bag-of-words, to train support vector machines, logistic regressors, and other classical ML methods~\citep{mathew2018thou,garland-etal-2020-countering} to more advanced approaches that leverage recurrent neural networks and transformer-based models~\citep{yu-etal-2022-hate}.

Today's LLMs leverage extensive pre-training with unsupervised learning, enabling them to capture semantic features while 
taking into account contextual information. LLMs have demonstrated remarkable strengths in hate speech recognition~\citep{RHMS2023,giorgi2024human,GuoHMS2024,sen2024hatetinyllm,jin-etal-2024-gpt} as well as detecting and generating counter speech~\citep{poudhar-etal-2024-strategy, hengle-etal-2024-intent}. 
\cite{LiFAH24} showed that prompting ChatGPT using
in-context learning can lead to an accuracy comparable to expert 
annotations in general tasks such as classifying harmful content, 
including toxic and hateful speech. \cite{GuoHMS2024} investigated various LLMs and prompting strategies for hate speech recognition and demonstrated that they significantly outperform transformer-based models.
In contrast, \cite{hengle-etal-2024-intent} utilise LLMs for counter speech classification by employing a one-shot prompting approach. Instead of relying on extensive fine-tuning, their method involved providing a single labelled example as a prompt to guide the model in categorising counter speech texts. 

Despite these advancements, the high complexity of LLMs makes it difficult to understand their internal functioning. Furthermore, LLMs suffer from limitations such as inconsistent responses, sensitivity to prompt design, and lack of reproducibility~\citep{TjuatjaCWTN24}. 
These challenges undermine their trustworthiness and reliability. 

\paratitle{Explaining LLMs.}
To explain LLMs, researchers typically explore local explanations, which provide insights into individual predictions. In the XAI literature, methods are commonly categorised based on whether the model is \emph{self-explaining} or requires external \emph{post-hoc} explanations~\citep{arya2019one}. Post-hoc methods rely on auxiliary techniques to interpret predictions obtained once training is finished. Examples of post-hoc approaches for explaining complex models like LLMs are proposed by~\cite{DalviDSBBG19} and \cite{AntvergB22}. Post-hoc methods are often model-agnostic and useful for debugging but are criticised for their lack of faithfulness to the model's internal reasoning, as they can be influenced by biases and spurious correlations~\citep{wilming2024gecobenchgendercontrolledtextdataset, slack2020fooling}. Alternatively, LLMs can be regarded as self-explaining models due to their ability to generate textual human-understandable explanations for their predictions. While these explanations enhance interpretability and allow users to assess their plausibility, they do not necessarily provide full transparency into the underlying decision-making process. A major challenge is that LLM-generated explanations can suffer from hallucination~\citep{HaoWP24}, where the model fabricates information that seems reasonable but is actually nonsensical. This limitation raises concerns about their faithfulness, as the generated explanations may not necessarily reflect the true reasoning behind a prediction.

Recent efforts have also investigated explanations rooted in the inherent mechanisms of LLMs~\citep{HanWT20,DasGKLL22}, including attention values for important tokens~\citep{BastingsAT19}, neuron activations~\citep{DalviDSBBG19}, 
and example-based reasoning~\citep{HanWT20,DasGKLL22}. 
However, attention-based approaches have been criticised for lacking faithfulness and not always providing plausible explanations~\citep{bibal2022attention, jain2019attention}.

These challenges in explaining complex LLMs highlight an existing research gap. While current approaches primarily provide local explanations that enhance interpretability at the instance level, they fail to capture broader model behaviour and decision patterns~\citep{LudanLYD24}. Our research does not solve the lack of transparency of LLMs. Instead, we leverage LLMs to build a semantically rich and interpretable representation and train transparent downstream models from it, which provide local and global interpretability.


\paratitle{Concept bottleneck models (CBMs) in NLP.}
CBMs provide a methodology that leverages the advantages of LLMs while providing human interpretability. 
In CBMs~\citep{KohNTMPKL20}, the model is trained to first map the input data to a set of concepts (\ie~referred to as \emph{bottleneck concepts}) that can be interpreted by humans. These concepts are then used by the model to make predictions. As humans can evaluate these concepts, they represent an interpretable connection between the model's predictions and the underlying reasoning process.
Originally, CBMs were designed for image classification and 
are an end-to-end solution that requires expensive training with annotations for selected concepts~\citep{OikarinenDNW23,YangPZJCY23}. 
Despite the extension to label-free classification~\citep{OikarinenDNW23}, only a handful of works have been proposed for NLP tasks.

\cite{LudanLYD24} proposed Text Bottleneck Models (TBMs), which use GPT-4 to infer concepts without relying on human-annotated labels for the concepts. Through experimentation with multiple text understanding tasks, they found TBMs to enhance the interpretability of the model predictions. \cite{TanCWYLL24} introduced a model called $\mbox{C}^3\mbox{M}$ and studied its impact on interpretability in sentiment analysis tasks. 
Compared to TBMs, it uses a set of mixed concepts consisting of both 
human-annotated concepts and ones generated from ChatGPT. \cite{SunOUW24} presented CB-LLM, which 
leverages ChatGPT to generate concepts. Compared to $\mbox{C}^3\mbox{M}$ and 
TBMs, CB-LLM leverages a text embedding model that encodes both concepts and input texts, ensuring that semantically compatible concepts and inputs receive similar embeddings. These pioneering approaches have demonstrated promising potential for applying CBMs for text classification. However, 
the concepts extracted typically are in the form of short free-text descriptions without a 
unified or standardised format. A concept, for example, can correspond to 
topics or targets (\eg~``personal threats'' or  
``targeted group discrimination'') or linguistic patterns (\eg~``attacking 
language'' and ``irony or sarcasm'').
%

In contrast to previous works, we introduce a CBM architecture for NLP that relies on simple attributes (\ie~adjectives). Such a representation can either be mined from the data or simply curated a priori (as performed in this work to provide a proof-of-concept). One advantage of using simple adjectives as concepts is that they can be well-evaluated for a given input text by LLMs due to their strong text comprehension capabilities. Furthermore, adjectives are easy to understand and verify by humans, making the bottleneck representation inherently explainable. 

\section{\HSCBM:  Speech Concept Bottleneck Models}
\label{sec:method}

In this section, we describe our methodology, which introduces adjectives as intermediate bottleneck concepts. 
We also present a light-weight network architecture designed for downstream tasks that utilises adjective-based concepts to enable inherently transparent and human-centric predictions.

\subsection{Problem definition and methodology overview}
\label{ssec:problem-definition}
To achieve our goal, we start by constructing a simple but effective human-interpretable layer of bottleneck concepts.
Encoding input texts using the proposed adjective-based concepts allows us to obtain a representation that more effectively captures hidden meanings, intentions, and emotions. We then develop a downstream classifier that generates explanations by highlighting the most relevant concepts (\ie~adjectives) for a given prediction.

Let $\textSet$ be a set of input texts and $\conceptSet$ be the set of 
bottleneck concepts. Every $t\in\textSet$ is a text sample analysed to determine whether it contains, \eg~hate speech or counter speech. To achieve this, we follow the 
CBM framework~\citep{KohNTMPKL20} and use the 
function $f_\conceptSet:\textSet\to\mathbb{R}^{\vert\conceptSet\vert}$
to encode a given input text into a vector embedding. Each feature of this vector corresponds to one of the predefined concepts, representing the degree of its relevance to the given input text.
We define a classifier $g:\mathbb{R}^{\vert\conceptSet\vert}\to \mathcal{L}$
where $\mathcal{L}$ is the set of labels used for the downstream task. The overall function to recognise hate speech can thus be
formalised as the composition of the above two functions, \ie~$g \circ f_\conceptSet:\textSet\to\mathcal{L}$.

As discussed previously in Section~\ref{sec:related work}, we address the challenge of providing bottleneck concepts that facilitate human-interpretable explanations.
Considering the specific characteristics of hate and counter speech, we introduce a lexicon of descriptive adjectives as the set of concepts.  This set is denoted by $\adjSet$ and 
aims to characterise the intentions expressed in input texts. 
Note that such lexicons can also be determined automatically, \eg~by 
prompting LLMs and selecting related responses like in~\citep{DELAPENASARRACEN2023103433}. 
As a proof-of-concept, we leverage manually defined adjectives
(see Appendix~\ref{app:lexicon_list}) provided by social scientists with respective domain knowledge. As a starting point, we extracted a set of definitions and adjectives used to describe different types of hate and counter speech from the literature~\citep{benesch2016counterspeech, mathew2018thou, LeeNSSP22}. This set was subsequently extended and refined in collaboration with domain experts to obtain a richer data representation, which allows for capturing the nuanced strategies observed in real-world hate and counter speech.  

\begin{figure}[h!]
    \centering
    \includegraphics[width=1.0\textwidth]{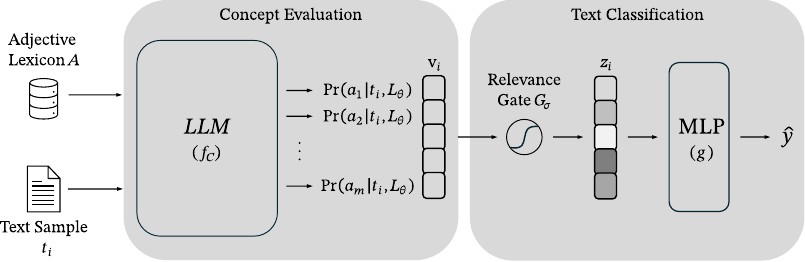}
    \caption{Overall architecture of the proposed Speech Concept Bottleneck Model (\HSCBM). 
    It consists of two sequential stages. The first, called concept evaluation, computes the encoding of the given input text in terms of the bottleneck concepts (i.e., adjectives), using a large language model (LLM). This 
    encoding is passed to the second stage, called text classification, which performs the final prediction. A relevance gate in this stage enables the model to dynamically learn the relative importance of the adjectives.}
    \label{fig:class-arch}
\end{figure}

Figure~\ref{fig:class-arch} depicts our proposed methodology.
The methodology consists of two components corresponding to two sequential steps: \emph{concept evaluation} and \emph{text classification}. 
The first component assesses the relevance of the adjectives in $\adjSet$ to a given input text and produces a vector of relevance scores as a concept bottleneck representation. 
The second component maps these relevance scores to a probability distribution over the pre-defined labels and outputs the most likely one as the final prediction.
In the concept evaluation step, instead of relying on syntactic representations (\eg~BoW or 
tf-idf), 
we propose a semantically rich and context-aware representation. 
Specifically, we distil knowledge from a pre-trained LLM to derive 
a representation in a zero-shot fashion. 
In the text classification step, we introduce a \textit{relevance gate} that enables the model to dynamically learn the relative importance of adjectives. 
Additionally, we incorporate a regularization in the loss function to constrain the number of adjectives used in the final classification. 
%
In the remainder of this section, we will describe the detailed design of these
two components.
\subsection{Concept evaluation} 
\label{ssec:concept-evaluation}
To motivate our evaluation method, 
we begin by briefly describing the straightforward and traditional approach to leverage LLMs for concept evaluation and discuss its shortcomings. 
For a dataset $\textSet$ of text samples and a lexicon of adjectives $\adjSet$, 
an LLM is prompted to determine one-by-one the elements from an incidence matrix 
$\M\in \{0,1\}^{\vert\textSet\vert\times\vert\adjSet\vert}$. 
An element of $\M$, denoted by $\M_{ij}\in\{0,1\}$,
indicates whether, according to the LLM's response, 
the $i$-th text sample in $\textSet$, \ie~$t_i$, can be characterised by the $j$-th adjective in $\adjSet$, denoted by $a_j$.
This established method is actually widely used to extract knowledge from LLMs in a zero-shot fashion. However, this method can only provide a coarse-grained evaluation, which is binary for each adjective in $\adjSet$. 
More importantly, this binary evaluation ignores the non-deterministic nature of LLM responses that rely on stochastic token sampling, 
even with temperature-smoothing (\ie~by modifying the entropy of 
the token distribution). This can lead to inconsistent values in $\M_{ij}$ in repeated concept evaluations. 

To tackle the above shortcomings, we propose a new method to address this limitation by taking into account the non-deterministic responses of LLMs.
In this way, we enable $\M_{ij}$ to quantify the probability 
that adjective $a_j$ can be used to describe a given text sample $t_i$. 
Formally, let $\LLM$ be the utilised LLM with parameters $\theta$, 
then we have: 

\begin{equation}
    \M_{ij} = \Pr(a_j\mid t_i,\LLM).
\end{equation}
%

In the following, we will take $a_j$ and $t_i$ as an example to illustrate our 
concept evaluation method. We construct a prompt for $a_j$ and $t_i$ with a 
template. Although responses given by an LLM are impacted by the formulation of prompts~\citep{RHMS2023},  we decided to use a simple template as  
proof-of-concept. If simplified prompts can lead to promising performance, then fine-tuned prompts can lead to further improvements. 
Our prompt template is simply:
\begin{center}
    \textit{``Tell me if the adjective [adjective] \\
    describes the content of the following text: [text]?''} 
\end{center}
where [\emph{adjective}] and [\emph{text}] are two placeholders for $a_j$ and 
$t_i$, respectively. Note that, if additional contextual information is available for the input texts (as provided by some benchmark datasets), we add it to the prompt before the above template. 
In addition, we explicitly provide information about specific personas, such as ``psychologists'' or ``social scientists'', in the prompt to further guide the LLMs, inspired by related works~\citep{giorgi2024human}, see also Appendix~\ref{app:templatep}. 
To assess the impact of different personas, we conduct a sensitivity analysis in our experiments. 
We use $T(a_j, t_i)$ to represent the prompt instantiated with our template $T$ when adjective $a_j$ and text sample $t_i$ are queried.

Given that the response of off-the-shelf LLMs to a given prompt is stochastic, each response $o$ is implicitly sampled from an underlying probability distribution.
We denote the probability that an LLM $\LLM$ outputs $o$ for the prompt $T(a_j, t_i)$ as $\Pr\left(o\mid T(a_j,t_i),\LLM\right)$.
We further assume that the probability of $a_j$ being able to characterise $t_i$ is the 
marginalised probability over all the \emph{positive} responses generated by $\LLM$.
Let $\Response$ be the set of all possible positive responses given by $\LLM$, then we have:
\begin{equation}
    \Pr(a_j\mid t_i,\LLM) = \sum_{o\in\Response} \Pr(o\mid T\left(a_j, r_i),\LLM\right).
\end{equation}
As the complete set of all possible positive responses $\Response$ are unavailable in practice, it is intractable to compute the marginalised probability as above. 
As a result, we focus on a subset of $\Response$, namely $\tildeResponse$, 
which consists of responses starting with the token \emph{`yes'}
\revised{(see \tableautorefname~\ref{tab:yes_tokens})}, 
with special characters and cases of letters ignored. 
Formally, we have the following approximate calculation:
\begin{equation}
\label{eq:prob_formula}
    \Pr(a_j\mid t_i,\LLM) = \sum_{o\in\tildeResponse} \Pr(o\mid T\left(a_j, r_i),\LLM\right) +\varepsilon_\LLM
\end{equation}
where $\varepsilon_\LLM$ corresponds to the residue probability mass of those
positive responses absent in $\tildeResponse$, \ie~$\Response/\tildeResponse$. 
Since $\tildeResponse\subseteq\Response$, $\varepsilon_\LLM$ is the sum of all responses not starting with \textit{`yes'} (or a variant of it). 

\begin{table}[t]
    \fontsize{8.3pt}{9pt}\selectfont
    \centering
    \caption{\revised{Sets of tokens to determine the subset of positive responses $\tildeResponse$ for the different LLMs employed in this work.}}
    \label{tab:yes_tokens}
    \begin{tabular}{l|p{10.5cm}}
         \hline
         \textbf{LLM $\LLM$} & \textbf{Token IDs \& strings} \\
         \hline
         Llama2-7b & 3869 (``\_Yes''), 4874 (``\_yes''), 8241 (``Yes''), 3582(``yes''), 22483 (``\_YES''), 10134 (``\_Ye'')\\
         \hline
         LeoLM-7b & 29967 (``Ja''), 12337(``\_ja''), 14021 (``\_Ja''), 1764 (``ja'') \\
         \hline
         Llama 3.1-8b & 9642 (``Yes''), 9891 (``yes''), 14331 (``YES''), 20137 (``:YES''), 41898 (``.Yes''), 58841 (``\_Yes''), 60844 (``\_yes''), 77830 (``\_YES''), 85502 (``.YES''), 98171 (``,Yes''), 5697 (``ja''), 45280 (``JA''), 53545 (``Ja'')\\
         \hline
    \end{tabular}
    \vspace{-3mm}
\end{table}

We have two challenges to solve in calculating Equation~\ref{eq:prob_formula}. The first is to estimate the value of $\varepsilon_\LLM$. Since the directly proportional relation
$\Pr(a_j\mid t_i,\LLM)\sim \sum_{o\in\tildeResponse}\Pr(o\mid T\left(a_j, r_i),\LLM\right)$ holds, we can omit the probability mass $\varepsilon_\LLM$ to evaluate our concepts. 
The second challenge is to collect the responses in $\tildeResponse$. This is hard because it is intractable to determine whether our collection of positive responses is complete.
To make this computation feasible, instead of sampling and traversing all
sequences starting with `yes' and its variants, we compute the probability of the LLM predicting `yes' as the first token.  
This reduces the summation of all free-text responses to adding up the number of `yes' tokens listed in Table~\ref{tab:yes_tokens} given by the LLM's tokenizer. 
Each of these terms corresponds to the residual probability of any sequence starting 
with those `yes' tokens. Let $\T_{\it yes}$ be the utilised list of  
`yes' tokens. Formally, the equation can be approximated as follows:
\begin{equation}
\label{eq:prob_formula_reduced}
    \Pr(a_j\mid t_i,\LLM) \approx \sum_{\token\in\T_{\it yes}} \Pr(\token\mid T\left(a_j, r_i),\LLM\right).
\end{equation}

\noindent
Finally, for the input text $t_i$, we concatenate all the calculated probabilities for the 
adjectives in $\adjSet$ and get a vector $\cVec_i$ of length $\vert\adjSet\vert$. This vector constitutes our bottleneck representation.

\subsection{Classification model}
\label{ssec:classification}
The encoding vector $\cVec_i$ can be interpreted as a semantically enriched 
abstraction of input text $t_i$ that captures its underlying intentions and emotions. 
We postulate that this encoding alone is sufficient to classify
the original input texts. We design a model as an implementation of  
the classification function $g$ taking $\cVec_i$ as input and generating a
probability distribution over the predefined labels $\mathcal{L}$.

Our general goal is to obtain a classifier that is light-weight, transparent, and flexible enough to be adapted to various downstream tasks considering the variety of definitions of labels $\mathcal{L}$.  
As depicted in Figure~\ref{fig:class-arch}, the main component of the proposed
classifier is a multi-layer perceptron~(MLP) composed of a single ReLU-activated hidden layer.
However, unlike prior CBM-based methods that rely on simple MLPs or single linear layers, our approach incorporates an additional masking function that aims to 
learn the importance of individual adjectives in $\cVec_i$, thereby enhancing the interpretability of the intermediate model outputs. 
To this end, we utilise a sigmoid-activated dense layer that is jointly learned to function as a relevance gate that weights the scores of individual adjectives in the bottleneck representation. By scaling $\cVec_i$, we have an intermediate latent representation of $\cVec_i$, denoted by $\z_i$:

\begin{equation}\label{equ:masked}
    \z_i = {\sf sigmoid}(\mathbf{W}\cdot \cVec_i)\odot \cVec_i,
\end{equation}
where $\mathbf{W}\in\mathbb{R}^{\vert\adjSet\vert\times\vert\adjSet\vert}$ is 
a matrix of parameters to learn.
In the following, we refer to $\z_i$ as the \emph{latent encoding} of text $t_i$.
Finally, $\z_i$ is input into the MLP and the final output 
$\haty_i$ can be computed as follows:
\begin{equation}
\haty_i= {\sf softmax}\left({\sf MLP}(\z_i)\right).
\end{equation}

We note that the latent encoding $\z_i$ can complement other forms of 
text encodings to enable more accurate predictions. 
We propose a simple but effective extension by integrating $\z_i$ in the fine-tuning of a transformer-based model, such as XLM-RoBERTa~\citep{liu2019robertarobustlyoptimizedbert}. 
To this end, we stack an MLP composed of a single ReLU-activated hidden layer on top of the transformer model and feed it with both $\z_i$ and 
the representation from the transformer encoder. 
Given an input text $t_i$, let ${\sf transform}(t_i)$ represent the $d$-dimensional encoding generated by a transformer encoder and 
$\parallel$ 
denote the concatenation operation. Then, we define:
\begin{equation}
\haty_i= {\sf softmax}\bigl({\sf MLP}\left(\mathbf{W}'\cdot\z_i\parallel {\sf transform}(t_i)\right)\bigr)    
\end{equation}

where $\mathbf{W}'\in\mathbb{R}^{d\times\vert\adjSet\vert}$ is learnt 
to align the values of the latent encoding $\z_i$ to the scale and 
distribution of the transformer encoding space.
According to our experiments, employing the latent encoding $\z_i$ 
with the transformer-based representation can effectively 
improve the performance compared to
solely fine-tuning the transformer-based model.
To avoid confusion, we refer to this extended model as \HSCBMT.

\paratitle{Model complexity.}
Although our model relies on LLMs to compute the 
initial concept-based representation, our classification model is \emph{light-weight} in terms of 
model complexity. As a comparison, the SCBM model involves only about 0.076 million trainable 
parameters, which is smaller by several orders than conventional transformer-based classifiers 
like XLM-RoBERTa (278 million) or BERT-large (335 million). 
This reduced model complexity leads to more efficient training and inference, 
especially in resource-constrained environments. 
It is important to note that the LLM used for concept extraction is employed in a frozen, inference-only 
manner, meaning that it does not contribute to the learnable parameters of the model. 
To conclude, our classification model is compact and computationally efficient. 
\subsection{Class-discriminative concept prioritisation}
\label{ssec:regularization}
One common challenge for CBMs is to specify the number of needed concepts. More concepts can lead to better expressiveness, but they also make decisions more complex and harder for human users to understand~\citep{miller2019explanation}.
Compared to existing methods working on selection strategies for concepts~\citep{LudanLYD24,SunOUW24,TanCWYLL24}, we propose a novel regularisation term that guides the model to learn a selection of the most discriminative concepts during training.

Assume that we have an ordered set of labels $\mathcal{L}$, with  $|\mathcal{L}| = n$.
For an adjective $a_i\in \adjSet$, we construct a vector $\bar{\mathbf{c}}_i\in \mathbb{R}^{n}$ whose features indicate the expected value of the masked relevance of this adjective for all text samples of class $j$.
Let $\z_t(i)$ be the masked relevance of adjective $a_i$ for 
text $t$ in its latent encoding $z_t$ (see \equationautorefname~\ref{equ:masked}). Then we have: 
\begin{equation}
    \mathbf{\bar{c}}_{ij} = \mathbb{E}_{{t\sim D_j}} \z_t(i),
\end{equation}

where $D_j$ denotes the distribution of text samples in the $j$-th class. 

To guide the model during training to identify and prioritise the most discriminative adjectives, \ie~those showing high relevance for only \textit{one} class, we introduce a \emph{class-discriminative regularisation} term. 
This regularisation influences the learnable parameters in the relevance gate, encouraging adjectives to exhibit strong alignment with a specific class by penalizing overlaps in their activations across multiple classes. Since this penalization relies on adjective activations aggregated over different classes, it cannot be computed for individual text samples. Instead, the class-discriminative regularisation is calculated considering the entire training set as follows:

\begin{equation}\label{eq:discriminativeloss}
    {\it Loss}_{\it CD} = \frac{1}{\vert\adjSet\vert}\sum_{a_i\in\adjSet}\sum_{j=1}^n 
    \sum_{k=1}^{j-1}\mathbf{\bar{c}}_{ij}\mathbf{\bar{c}}_{ik}. 
\end{equation}

This class-discriminative (CD) regularisation term iteratively computes pairwise products of expected importance values $\mathbf{\bar{c}}_{ij}$ and $\mathbf{\bar{c}}_{ik}$ for all class pairs $(j,k)$, thereby penalizing shared relevance across classes.
This results in a sparser latent representation used for the final prediction, thereby enhancing interpretability~\citep{miller2019explanation}.

The final loss function is defined as a weighted sum of the cross-entropy and the proposed class-discriminative regularisation term: 
\begin{equation}
    {\it Loss} = \sum_{t_i}-\y_i\log\haty_i + \lambda {\it Loss}_{\it CD}
\end{equation}

where $\lambda\in\mathbb{R}$ is a hyper-parameter that controls the importance of the regularization.
During training, we optimise the model's parameters using the RMSProp algorithm~\citep{hinton2012rmsprop} and process the dataset in mini-batches. For each mini-batch, the expected importance of an adjective $a_i$ in a class $j$, denoted as $\mathbf{\bar{c}}_{ij}$, is dynamically estimated based on the current mini-batch data distribution when computing ${\it Loss}_{\it CD}$.

%
\subsection{Runtime optimisations}
\label{ssec:runtime}
As described in \sectionautorefname~\ref{ssec:concept-evaluation}, we estimate the relevance of an individual adjective for an input text as the probability of sampling a variant of the `yes' token when prompting the LLM $\LLM$.
Our approach involves a forward pass through multiple decoder blocks in  $\LLM$. This forward pass includes calculating the self-attention values of the input tokens, which exhibits a computational complexity that grows quadratically with the prompt length. 
In the self-attention calculation, the embeddings of the input tokens 
are sequentially transformed into three vectors: query, key, and value, 
stored as rows in three matrices $\Q$, $\K$, and $\V$, respectively. 
Each query vector $\Q_i$ is compared against all keys $\K_j$ to compute 
pairwise attention scores, which quantify the relative relevance of tokens. 
These scores are then used to weight the value vectors in $\V$, producing a weighted average that encodes the most relevant information for the output representation of the tokens to feed the next decoder block of $\LLM$.

To estimate the relevance of an adjective to an input text, we split the prompt fed into the LLM into a fixed prefix and a suffix. The prefix is invariant to each combination of adjectives and text samples, and it is mainly composed of the persona injected into the LLM via the prompt. The suffix comprises the remaining text in the prompt, including the combination of adjective and input sample (see Appendix~\ref{app:prefix+suffix}).

To optimise computation, we pre-compute and cache the $\K$ and $\V$ matrices corresponding to the prefix’s forward pass through the transformer blocks. 
These cached matrices are reused during the attention computations for each pair of adjectives and text samples.  
Similar optimisation is also used to improve the efficiency of LLMs in handling large context windows~\citep{chan2024don}.

Provided that different pairs of adjectives and texts lead to sequences of different lengths, we employ padding tokens to ensure uniform batch sizes during the computation. However, as autoregressive models are not pre-trained with embeddings for padding tokens, their hidden states are excluded during the forward pass using masked attention. 
The attention mask also ensures the correct extraction of the 
final hidden state, which corresponds to the last non-padding token in the sequence. The softmax probabilities derived from this hidden state in the next token prediction are then used to estimate the desired concept evaluation. 
This runtime optimisation streamlines the process while efficiently handling varying sequence lengths.


\section{Experimental Setup}
\label{sec:experiments}

\subsection{Datasets}
We employ five publicly available and frequently utilised datasets from our related works. To demonstrate the versatility of our approach, each dataset is selected to represent a specific task related to recognising hate speech (HS), toxic and offensive speech, as well as counter speech (CS). 
Detailed information on all selected datasets is summarised in Table~\ref{tab:datasets_summary}.
We provide a brief description of each dataset in the  following:
\begin{itemize}[leftmargin=*]
\item \textbf{GermEval}~\citep{wiegand2018overview} consists of a collection of posts in German crawled from X (previously Twitter) annotated with respect to the presence of offensive language. 
We examined the binary classification of \texttt{offensive} vs. \texttt{other} speech.
\item  \textbf{ELF22}~\citep{LeeNSSP22} consists of pairs of troll comments and responses, supplemented by additional ``responses'' created by annotators. We use the dataset for binary classification of the responses, \ie~\texttt{troll} vs. \texttt{covert troll}, while considering the comments as context.
\item  \textbf{HS-CS}~\citep{yu-etal-2022-hate} consists of pairs of comments and responses from Reddit annotated as \texttt{hate speech}, \texttt{neutral}, or \texttt{counter hate}. 
\item \textbf{TSNH} (Thou shalt not hate)~\citep{mathew2018thou} comprises manually annotated comments from YouTube with hateful content, labelled as \texttt{counter speech} or \texttt{non-counter speech}. Note that, since no validation and test sets are available for TSNH, we apply cross-validation on the training set.
\item  \textbf{CONAN}~\citep{Chung_2019} is a multilingual collection of hate speech and counter-narratives (\ie~counter speech), including annotations on hate speech subtopics and counter-narrative types. 
We use its English portion to perform multi-class classification of counter-narratives with the following labels: 
\texttt{facts}, \texttt{denouncing}, \texttt{hypocrisy}, \texttt{question}, \texttt{unrelated}, \texttt{support}, and \texttt{humor} (the same as in~\citep{chung-etal-2021-multilingual}).  
\end{itemize}

The prompts we use to query the LLMs vary across datasets due to the nature and structure of the target texts (\eg~conversations or provision of context). We list the templates of prompts for each dataset in Appendix~\ref{app:templatep}.

\begin{table}[t]
\caption{Overview of all utilised datasets in our experiments. The datasets cover different tasks as well as different languages and stem from different social media platforms. Abbreviations: HS: hate speech, CH: counter hate, CS: counter speech.}
\label{tab:datasets_summary}
\centering
\resizebox{\linewidth}{!}{
\begin{tabular}{lllll}
\toprule
\multirow{2}{*}{\text{\bf Name}} & \multirow{2}{*}{\text{\bf Lang.}} & \multirow{2}{*}{\text{\bf Labels}} & \text{\bf Partitioning} & \multirow{2}{*}{\text{\bf Data source}}\\ 
& & & \text{(Train/Test/Val)} \\
\midrule
{\bf GermEval}~\citep{wiegand2018overview} & DE & Offensive, Other & 5,009/3,532/- & X / Twitter \\
{\bf ELF22}~\citep{LeeNSSP22} & EN & Overt Troll, Covert Troll & 5,351/762/573  & Reddit \& Human\\
{\bf HS-CS}~\citep{yu-etal-2022-hate} & EN & HS, Neutral, CH & 3,325/713/713 & Reddit \\
{\bf TSNH}~\citep{mathew2018thou} & EN & CS, Non-CS & 13,924/-/- & YouTube\\
{\bf CONAN}~\citep{Chung_2019} & EN & {Facts, Denouncing, Hypocrisy,} & 2,452/613/- & Human \\
 & & {Question, Unrelated, Support,} & &  \\
  & & {Humor} & &  \\

\bottomrule
\end{tabular}
}
\end{table}
\subsection{Evaluation design and model selection} 
\label{subsec:evalDesign}
In our experiments, we evaluate five categories of models which 
differ from each other in terms of the compositions of classifiers and 
text encoding methods: 
\begin{itemize}[leftmargin=*]
\item Category I -- conventional ML models based on conventional text representations, to provide a conventional baseline for reference.
\item Category II -- conventional ML models applied on the proposed bottleneck concept representations 
to investigate the effectiveness of our new representation compared with 
conventional text encodings. 
\item Category III -- fine-tuned transformer models followed by an MLP classifier to 
provide respective baselines for transformer-based encodings. 
\item Category IV -- various LLMs and prompting techniques, i.e., in-context learning and chain-of-thoughts)  to obtain performance baselines for state-of-the-art language models.
\item Category V -- the proposed approaches \HSCBM~and \HSCBMT~including their variants 
using the proposed class-discriminative regularisation term.
\end{itemize}

To ensure a comprehensive comparison, we select not only state-of-the-art methods but also classical ML approaches as our baselines. 
For the first two categories, we employ support vector machine (SVM), 
logistic regression (LR), random forest (RF), gradient boosting (GB), and multi-layer perceptron (MLP) as classifiers. 
For category III, we incorporate  pre-trained instances of a BERT-base 
model\footnote{Weights from \url{https://huggingface.co/google-bert/bert-base-uncased} (English) and \url{https://huggingface.co/google-bert/bert-base-german-cased} (German).} 
and XLM-RoBERTa\footnote{\url{https://huggingface.co/FacebookAI/xlm-roberta-base}}. 
In category IV, we evaluate various LLMs. 
We evaluate Llama 3.1-8b-instruct as an open-source LLM, along with three proprietary models via the official OpenAI API, i.e., GPT-3.5-turbo, GPT 4o, and o3-mini. Llama 3.1-8b-instruct and
GPT-3.5-turbo are evaluated in a zero-shot setting. GPT 4o is evaluated in zero-shot and in-context learning (ICL) settings.
GPT o3-mini is a reasoning model that internally generates a chain of thoughts (CoT) during inference. \revised{For the proprietary models accessed through APIs, we use the default settings for all parameters 
such as temperature and top-k, except the maximum number of output tokens, which was limited to 10. 
Additionally, these models are prompted and evaluated in a zero-shot approach.}

Category V includes our proposed \HSCBM~and its extension with transformer-based models, \ie~\HSCBMT. 
Our \HSCBM~model relies on an LLM to evaluate the concept abstraction.
Without loss of generality, we utilise frozen 8-bit quantised pre-trained instances of the Llama 2-7b and Llama 3.1-8b-instruct\footnote{Weights from \url{https://huggingface.co/meta-llama/Llama-2-7b-chat-hf} and \url{https://huggingface.co/meta-llama/Llama-3.1-8B-Instruct}}~\citep{touvron2023llama,dubey2024llama} models as $\LLM$. Llama is a family of open-source language models that have achieved competitive results compared to proprietary and closed-source models.
Among the available versions, we employ Llama 2 with 7 billion parameters, optimised with Reinforcement Learning from Human Feedback (RLHF), and Llama 3.1-8b-instruct, a further refined model designed for enhanced reasoning and instruction-following capabilities in conversational settings.
The same LLMs are also used in 
the models of the second category to generate text encodings.
For our \HSCBMT~model, we additionally use XLM-RoBERTa-base\footnote{\url{https://huggingface.co/FacebookAI/xlm-roberta-base}}~\citep{Conneau2019UnsupervisedCR}, which is a multilingual instance of RoBERTa~\citep{liu2019robertarobustlyoptimizedbert} as the transformer backbone. 
In order to test the effectiveness of our proposed class-discriminative loss ${\it Loss}_{\it CD}$, we also include the corresponding variants with loss regularisation, denoted as \HSCBM-R~and \HSCBMT-R.
Note that when employing different LLMs, the prediction performance may vary. Intuitively, stronger LLMs give a more accurate evaluation of descriptive adjectives, which will subsequently improve the final recognition accuracy.  
We opt for an open-source LLM despite the frequently reported superior performance of proprietary models in the literature, as our approach requires generating a soft probability for an adjective describing a given text. Closed-source LLMs typically provide only hard predictions along their logits, which limits their suitability for our approach.
We emphasise that the goal of our experiments is to demonstrate the effectiveness of our proposed concept bottleneck model. An exhaustive comparison of the performance of available LLMs would exceed the scope of this paper.
Another influencing factor on the performance is the selection of prompting techniques. Our main results are based on zero-shot prompting; however, we also conduct additional experiments to investigate the 
impact of different prompt engineering techniques for closed-source models (see \sectionautorefname~\ref{sec:performance_eval}) and our \HSCBM~model (see \sectionautorefname~\ref{subsec:sensible}).

\subsection{Model training and evaluation}
For all investigated approaches except for our two CBM-based models, we set the hyperparameters to the values that are suggested or widely adopted in the literature. 
To train \HSCBM, we employ a fixed learning rate of $2e-3$ 
with RMSProp optimised over 300 training epochs, with early stopping based on validation set performance. 
For fine-tuning transformer-based models and training \HSCBMT, 
a fixed learning rate of $1e-5$ is used across all transformer blocks and MLP layers with RMSProp optimised over 16 epochs. 
We use different training settings for these models because fine-tuning the transformer backbone in \HSCBMT~requires lower learning rates to prevent overwriting the knowledge gained during pre-training.
In both cases, early stopping is applied based on the macro-$F_1$ performance on the validation set.

\revised{For LLMs used in Category IV as zero-shot classifiers, we define templates and generate  prompts from them to obtain the desired outputs from the LLM. 
Due to the non-deterministic nature of LLM-generated sequences, we apply majority voting over three repetitions of each input prompt to determine the final results. We provide the prompting templates and the definition of the in-context learning strategy 
in Appendix~\ref{app:template_zeroshot}.}

As some datasets have imbalanced class distributions, we use the macro-$F_1$ as the main performance metric, because it implicitly considers class imbalance in performance estimation. 
The partitioning of each dataset follows the original splits provided by the creators. All results are averaged over five runs to account for variability due to model parameter initialization. Note that as no validation set is available for the TSNH dataset, we carry out a stratified five-fold cross-validation and report the average performance across the five folds.

The implementation of the proposed methodology, baselines, and experiments was carried out using Python (version 3.11.9), the PyTorch framework (version 2.6.0) for neural networks, and the scikit-learn framework (version 1.6.1). 
The models were trained and evaluated on an NVIDIA H100 GPU with 80GB of memory.\revised{\footnote{The source code for our methods and experiment implementations are 
publicly available at \url{https://github.com/fhstp/SCBM}.}}

\section{Results}

\begin{figure}[h!]
    \centering
    \includegraphics[width=.7\linewidth]{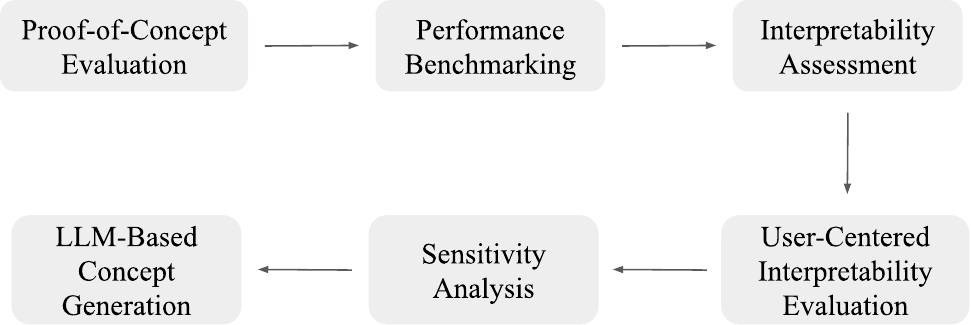}
    \caption{Flowchart of our experimental analysis. We design the experiments around six main research aims.}
    \label{fig:experiments}
\end{figure}

We design our experimental analysis around six main research aims, 
as shown in Figure~\ref{fig:experiments}. We start with a proof-of-concept analysis to
assess whether the bottleneck concept representations can capture hate and counter speech content. 
We conduct a comprehensive performance comparison across the five methodological categories (including our proposed approach) and compare the results to those reported in the literature for the investigated datasets. Furthermore, we examine various aspects of the inherent local and global interpretability of our approach, including the impact of the novel class-discriminative regularization. Next, we present the results of a user evaluation assessing the local interpretability of our approach. This is followed by a detailed sensitivity analysis to investigate the influence of different subsets of the proposed adjective set and the effect of various LLM prompt engineering techniques on our model. We conclude the section by investigating the potential of LLMs to automatically generate adjective concepts.

\subsection{Concept evaluation} 

\begin{figure}[!b]
    \centering
    \includegraphics[scale=0.45]{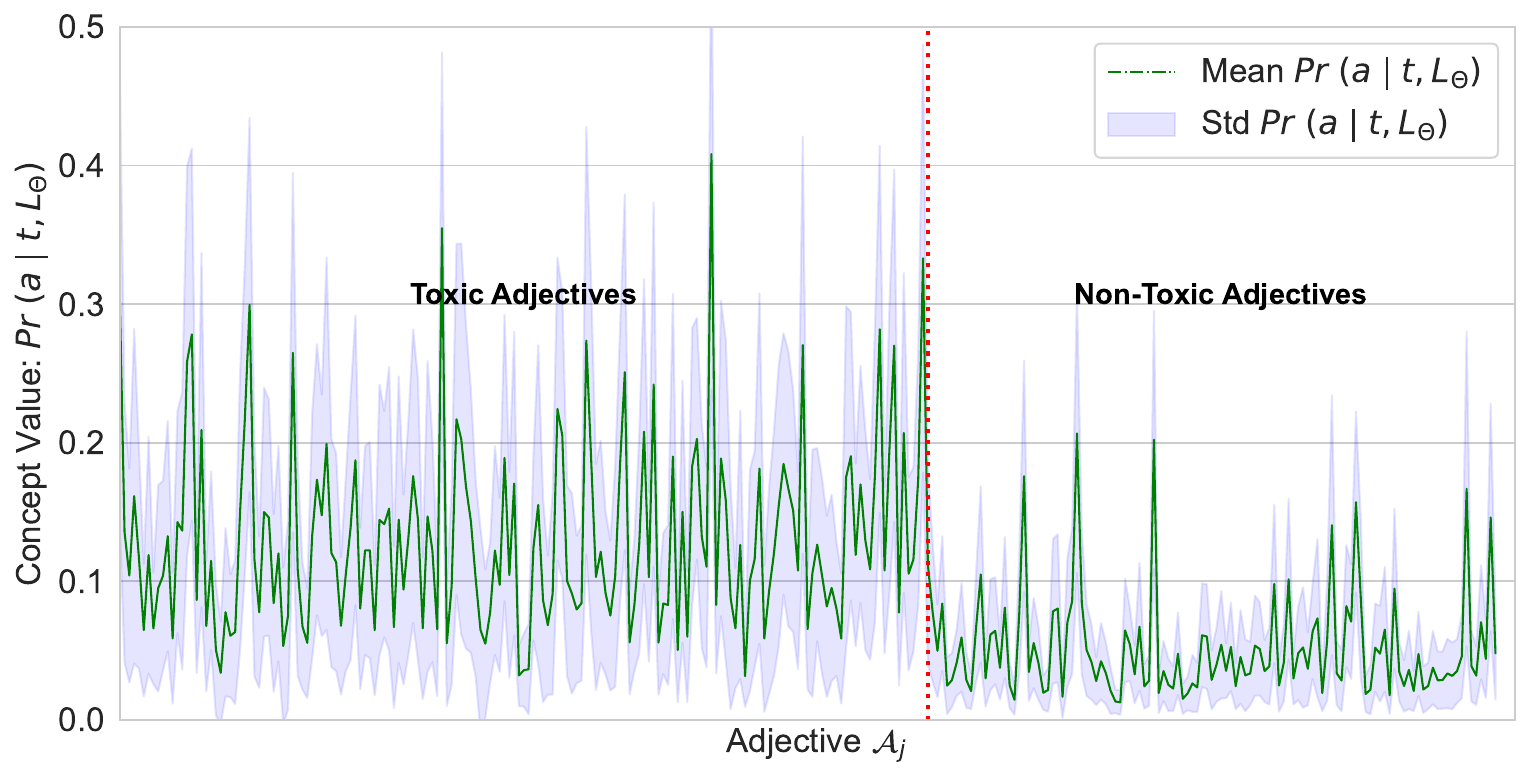}
    \caption{Visualisation of the bottleneck representation, \ie~$\Pr(a\mid t, \LLM)$, as defined in \equationautorefname~\ref{eq:prob_formula_reduced} for a subset of 100 comments from the \texttt{hateful speech} class in the GermEval dataset. The x-axis represents the adjectives, and the y-axis represents their relevance value. The green line indicates the average relevance across these 100 comments, while the purple area represents the variance. 
    }
    \label{fig:meanprob}
\end{figure}

As a preliminary proof-of-concept, we randomly select 100 \emph{offensive} samples from the GermEval dataset and determine for each the bottleneck representation (as defined in \equationautorefname~\ref{eq:prob_formula_reduced}) to assess whether this representation can capture toxic content. 
In Figure~\ref{fig:meanprob}, we show the average values of relevance for each adjective in green and the corresponding variances in purple across these 100 samples. The adjectives to the left of the vertical red line characterize hateful and offensive speech. The adjectives to the right of the red line characterize rather counter speech-related content. 
We observe an obvious difference in the relevance values between the two subsets of adjectives. Hate speech-related adjectives are assigned much higher probabilities 
compared to those related to non-toxic and counter speech content. 
The top-10 most relevant adjectives across the 100 samples are listed in Table~\ref{tab:highestval}.  
We can see that all the top-10 adjectives are appropriate to describe toxic and hateful content. This experiment offers initial empirical evidence supporting the effectiveness of the proposed knowledge distillation approach in evaluating adjective concepts.

\begin{table}[t]
    \fontsize{9.3pt}{10pt}\selectfont
    \centering
    \caption{Top-10 most relevant adjectives related to 100 sampled instances from the offensive class in the GermEval dataset.}
    \label{tab:highestval}
    \begin{tabular}{l|l||l|l}
        \hline
        \textbf{German} & \textbf{English} & \textbf{German} & \textbf{English}\\
        \hline\hline 
         abwertend&pejorative & entwertend&devaluing \\
         relativierend&relativizing & herabw\"{u}rdigend&degrading\\
         diffamierend&defamatory & vorurteilsbehaftet&prejudiced\\
         ettiketierend&eticizing & verleumderisch&slanderous\\
         abf\"{a}llig&derogatory & ehrverletzend&defamatory\\
         \hline
    \end{tabular}
\end{table}

\subsection{Performance evaluation}~\label{sec:performance_eval}
We evaluate the performance of all baselines and our proposed models on all five datasets. The results are summarised in \tableautorefname~\ref{tab:class_result}. 
The first row of the table provides the random baseline for reference, where labels are randomly assigned following a uniform distribution.
The runs are grouped into five categories (I, II, $\ldots$, V) as described in 
Section~\ref{subsec:evalDesign}. From the results, we have the following 
observations. 

\begin{table}[t]
\fontsize{8pt}{9pt}\selectfont
    \centering
    \caption{Performance of all employed approaches across all explored datasets in terms of macro-$F_1$ score. The best-performing approach in each category (I, II, ... V) is highlighted in italics, and the best-performing approach per dataset is highlighted in bold.} 
    \label{tab:class_result}
\resizebox{\linewidth}{!}{
\begin{tabular}{|l|ll|lllll|}
\hline
\multirow{2}{*}{\textbf{}}                 & \multicolumn{2}{l|}{\multirow{2}{*}{\textbf{Method}}}                                            & \multicolumn{5}{c|}{\textbf{Dataset}}                                                                                                                                                                                         \\ \cline{4-8} 
                                           & \multicolumn{2}{l|}{}                                                                            & \multicolumn{1}{l|}{\textbf{GermEval}}         & \multicolumn{1}{l|}{\textbf{ELF22}}            & \multicolumn{1}{l|}{\textbf{HS-CS}}            & \multicolumn{1}{l|}{\textbf{CONAN}}            & \textbf{TSNH}             \\ \hline
                                           & \multicolumn{2}{l|}{Random}                                                                      & \multicolumn{1}{l|}{0.488}                     & \multicolumn{1}{l|}{0.515}                     & \multicolumn{1}{l|}{0.347}                     & \multicolumn{1}{l|}{0.109}                     & 0.503                     \\ \hline
\multicolumn{1}{|c|}{\multirow{5}{*}{I}}   & \multicolumn{2}{l|}{SVM}                                                                         & \multicolumn{1}{l|}{$\mathit{0.648_{\pm 0.000}}$}  & \multicolumn{1}{l|}{$0.553_{\pm 0.000}$}          & \multicolumn{1}{l|}{$\mathit{0.426_{\pm 0.000}}$}  & \multicolumn{1}{l|}{$0.364_{\pm 0.000}$}          & $\mathit{0.696_{\pm 0.007}}$ \\ \cline{2-8} 
\multicolumn{1}{|c|}{}                     & \multicolumn{2}{l|}{LR}                                                                          & \multicolumn{1}{l|}{$0.586_{\pm 0.000}$}          & \multicolumn{1}{l|}{$\mathit{0.556_{\pm 0.000}}$}  & \multicolumn{1}{l|}{$0.413_{\pm 0.000}$}          & \multicolumn{1}{l|}{$0.322_{\pm 0.000}$}          & $0.693_{\pm 0.007}$          \\ \cline{2-8} 
\multicolumn{1}{|c|}{}                     & \multicolumn{2}{l|}{RF}                                                                          & \multicolumn{1}{l|}{$0.535_{\pm 0.009}$}          & \multicolumn{1}{l|}{$0.531_{\pm 0.027}$}          & \multicolumn{1}{l|}{$0.323_{\pm 0.014}$}          & \multicolumn{1}{l|}{$0.259_{\pm 0.005}$}          & $0.689_{\pm 0.005}$          \\ \cline{2-8} 
\multicolumn{1}{|c|}{}                     & \multicolumn{2}{l|}{GB}                                                                          & \multicolumn{1}{l|}{$0.571_{\pm 0.002}$}          & \multicolumn{1}{l|}{$0.547_{\pm 0.027}$}          & \multicolumn{1}{l|}{$0.374_{\pm 0.008}$}          & \multicolumn{1}{l|}{$0.368_{\pm 0.005}$}          & $0.668_{\pm 0.008}$          \\ \cline{2-8} 
\multicolumn{1}{|c|}{}                     & \multicolumn{2}{l|}{MLP}                                                                         & \multicolumn{1}{l|}{$0.648_{\pm 0.003}$}          & \multicolumn{1}{l|}{$0.542_{\pm 0.010}$}          & \multicolumn{1}{l|}{$0.398_{\pm 0.003}$}          & \multicolumn{1}{l|}{$\mathit{0.386_{\pm 0.011}}$}  & $0.672_{\pm 0.005}$          \\ \hline
\multicolumn{1}{|c|}{\multirow{10}{*}{II}} & \multicolumn{1}{l|}{\multirow{2}{*}{SVM}}                                             & Llama 2   & \multicolumn{1}{l|}{$0.695_{\pm 0.029}$}          & \multicolumn{1}{l|}{$0.356_{\pm 0.000}$}          & \multicolumn{1}{l|}{$0.504_{\pm 0.000}$}          & \multicolumn{1}{l|}{$0.593_{\pm 0.000}$}          & $0.637_{\pm 0.090}$          \\ \cline{3-8} 
\multicolumn{1}{|c|}{}                     & \multicolumn{1}{l|}{}                                                                 & Llama 3.1 & \multicolumn{1}{l|}{$\mathit{0.779_{\pm  0.000}}$} & \multicolumn{1}{l|}{$0.669_{\pm 0.000}$}          & \multicolumn{1}{l|}{$\mathit{0.577_{\pm 0.000}}$} & \multicolumn{1}{l|}{$0.602_{\pm 0.000}$}          & $0.724_{\pm 0.010}$          \\ \cline{2-8} 
\multicolumn{1}{|c|}{}                     & \multicolumn{1}{l|}{\multirow{2}{*}{LR}}                                              & Llama 2   & \multicolumn{1}{l|}{$0.693_{\pm 0.029}$}          & \multicolumn{1}{l|}{$0.356_{\pm 0.000}$}          & \multicolumn{1}{l|}{$0.504_{\pm 0.000}$}          & \multicolumn{1}{l|}{$0.593_{\pm 0.000}$}          & $0.646_{\pm 0.093}$          \\ \cline{3-8} 
\multicolumn{1}{|c|}{}                     & \multicolumn{1}{l|}{}                                                                 & Llama 3.1 & \multicolumn{1}{l|}{$0.777_{\pm 0.000}$}          & \multicolumn{1}{l|}{$0.671_{\pm 0.000}$}          & \multicolumn{1}{l|}{$\mathit{0.577_{\pm 0.000}}$} & \multicolumn{1}{l|}{$0.602_{\pm 0.000}$}          & $0.723_{\pm 0.009}$          \\ \cline{2-8} 
\multicolumn{1}{|c|}{}                     & \multicolumn{1}{l|}{\multirow{2}{*}{RF}}                                              & Llama 2   & \multicolumn{1}{l|}{$0.689_{\pm 0.028}$}          & \multicolumn{1}{l|}{$0.646_{\pm 0.010}$}          & \multicolumn{1}{l|}{$0.466_{\pm 0.012}$}          & \multicolumn{1}{l|}{$0.394_{\pm 0.012}$}          & $0.604_{\pm 0.010}$          \\ \cline{3-8} 
\multicolumn{1}{|c|}{}                     & \multicolumn{1}{l|}{}                                                                 & Llama 3.1 & \multicolumn{1}{l|}{$0.757_{\pm 0.004}$}          & \multicolumn{1}{l|}{$\mathit{0.671_{\pm  0.003}}$} & \multicolumn{1}{l|}{$0.487_{\pm 0.009}$}          & \multicolumn{1}{l|}{$0.486_{\pm 0.012}$}          & $0.719_{\pm 0.005}$          \\ \cline{2-8} 
\multicolumn{1}{|c|}{}                     & \multicolumn{1}{l|}{\multirow{2}{*}{GB}}                                              & Llama 2   & \multicolumn{1}{l|}{$0.729_{\pm 0.019}$}          & \multicolumn{1}{l|}{$0.561_{\pm 0.000}$}          & \multicolumn{1}{l|}{$0.500_{\pm 0.001}$}          & \multicolumn{1}{l|}{$0.481_{\pm 0.000}$}          & $0.642_{\pm 0.092}$          \\ \cline{3-8} 
\multicolumn{1}{|c|}{}                     & \multicolumn{1}{l|}{}                                                                 & Llama 3.1 & \multicolumn{1}{l|}{$0.766_{\pm 0.000}$}          & \multicolumn{1}{l|}{$0.577_{\pm 0.001}$}          & \multicolumn{1}{l|}{$0.562_{\pm 0.002}$}          & \multicolumn{1}{l|}{$0.534_{\pm 0.002}$}          & $0.721_{\pm 0.006}$          \\ \cline{2-8} 
\multicolumn{1}{|c|}{}                     & \multicolumn{1}{l|}{\multirow{2}{*}{MLP}}                                             & Llama 2   & \multicolumn{1}{l|}{$0.743_{\pm 0.017}$}          & \multicolumn{1}{l|}{$0.396_{\pm 0.079}$}          & \multicolumn{1}{l|}{$0.481_{\pm 0.011}$}          & \multicolumn{1}{l|}{$\mathit{0.627_{\pm  0.011}}$} & $0.640_{\pm 0.096}$          \\ \cline{3-8} 
\multicolumn{1}{|c|}{}                     & \multicolumn{1}{l|}{}                                                                 & Llama 3.1 & \multicolumn{1}{l|}{$0.762_{\pm 0.018}$}          & \multicolumn{1}{l|}{$0.654_{\pm 0.017}$}          & \multicolumn{1}{l|}{$0.556_{\pm 0.014}$}          & \multicolumn{1}{l|}{$0.618_{\pm 0.018}$}          & $\mathit{0.728_{\pm  0.007}}$ \\ \hline
\multirow{4}{*}{III}                       & \multicolumn{2}{l|}{XLM-RoBERTa-base}                                                            & \multicolumn{1}{l|}{$0.747_{\pm 0.017}$}          & \multicolumn{1}{l|}{$0.645_{\pm 0.018}$}          & \multicolumn{1}{l|}{$0.524_{\pm 0.008}$}          & \multicolumn{1}{l|}{$0.729_{\pm 0.016}$}          & $0.747_{\pm 0.013}$          \\ \cline{2-8} 
& \multicolumn{2}{l|}{BERT-base}                                                                   & \multicolumn{1}{l|}{$0.654_{\pm 0.040}$}          & \multicolumn{1}{l|}{$0.670_{\pm 0.008}$}          & \multicolumn{1}{l|}{$0.543_{\pm 0.004}$}          & \multicolumn{1}{l|}{$0.721_{\pm 0.022}$}          & $0.752_{\pm 0.022}$          \\ \cline{2-8} 
                                           & \multicolumn{2}{l|}{XLM-RoBERTa-large}                                                           & \multicolumn{1}{l|}{$\mathit{0.786_{\pm  0.004}}$} & \multicolumn{1}{l|}{$0.680_{\pm 0.008}$}          & \multicolumn{1}{l|}{$\mathit{0.572_{\pm  0.021}}$} & \multicolumn{1}{l|}{$0.746_{\pm 0.020}$}           & $\mathbf{0.781_{\pm 0.009}}$  \\ \cline{2-8} 
                                           & \multicolumn{2}{l|}{BERT-large}                                                                  & \multicolumn{1}{l|}{$0.676_{\pm 0.014}$}          & \multicolumn{1}{l|}{$\mathit{0.683_{\pm  0.009}}$} & \multicolumn{1}{l|}{$0.545_{\pm 0.011}$}          & \multicolumn{1}{l|}{$0.744_{\pm 0.007}$}          & $0.773_{\pm 0.008}$          \\ \hline
\multirow{3}{*}{IV}                        & \multicolumn{2}{l|}{GPT 3.5}                                                                     & \multicolumn{1}{l|}{$0.686_{\pm 0.003}$}          & \multicolumn{1}{l|}{$0.469_{\pm 0.078}$}          & \multicolumn{1}{l|}{$0.247_{\pm 0.012}$}          & \multicolumn{1}{l|}{$0.291_{\pm 0.067}$}          & $0.508_{\pm 0.022}$          \\ \cline{2-8} 
& \multicolumn{2}{l|}{GPT 4o}                                                                      & \multicolumn{1}{l|}{$0.833_{\pm 0.025}$}           & \multicolumn{1}{l|}{$0.500_{\pm  0.039}$}          & \multicolumn{1}{l|}{$0.267_{\pm 0.014}$}          & \multicolumn{1}{l|}{$0.361_{\pm  0.140}$}          & $0.560_{\pm  0.017}$           \\ \cline{2-8} 
& \multicolumn{2}{l|}{GPT 4o (ICL)}                                                               & \multicolumn{1}{l|}{$\mathbf{0.854_{\pm 0.002}}$}  & \multicolumn{1}{l|}{$\mathit{0.651_{\pm 0.005}}$}  & \multicolumn{1}{l|}{$\mathit{0.390_{\pm 0.006}}$}  & \multicolumn{1}{l|}{$\mathbf{0.763_{\pm 0.007}}$}  & $\mathit{0.642_{\pm 0.026}}$  \\ \cline{2-8} 
& \multicolumn{2}{l|}{GPT o3-mini (CoT)}                                                                 & \multicolumn{1}{l|}{$0.666_{\pm 0.165}$}           & \multicolumn{1}{l|}{$0.606_{\pm 0.004}$}           & \multicolumn{1}{l|}{$0.301_{\pm 0.008}$}           & \multicolumn{1}{l|}{$0.542_{\pm 0.012}$}           & $0.503_{\pm 0.009}$           \\ \cline{2-8} 
& \multicolumn{2}{l|}{Llama 3.1}                                                                    & \multicolumn{1}{l|}{$0.700_{\pm 0.112}$}          & \multicolumn{1}{l|}{$0.510_{\pm 0.013}$}          & \multicolumn{1}{l|}{$0.270_{\pm 0.018}$}           & \multicolumn{1}{l|}{$0.203_{\pm 0.017}$}          & $0.438_{\pm 0.081}$          \\ \hline
\multirow{8}{*}{V}                         & \multicolumn{1}{l|}{\multirow{2}{*}{{\sf \HSCBM}}}                                  & Llama 2   & \multicolumn{1}{l|}{$0.746_{\pm 0.004}$}          & \multicolumn{1}{l|}{$0.673_{\pm 0.007}$}          & \multicolumn{1}{l|}{$0.536_{\pm 0.005}$}          & \multicolumn{1}{l|}{$0.616_{\pm 0.011}$}          & $0.705_{\pm 0.013}$          \\ \cline{3-8} 
& \multicolumn{1}{l|}{}                                                                 & Llama 3.1 & \multicolumn{1}{l|}{$\mathit{0.781_{\pm  0.003}}$} & \multicolumn{1}{l|}{$\mathbf{0.693_{\pm 0.011}}$}  & \multicolumn{1}{l|}{$\mathbf{0.581_{\pm 0.008}}$}  & \multicolumn{1}{l|}{$0.630_{\pm 0.006}$}          & $0.739_{\pm 0.008}$          \\ \cline{2-8} 
& \multicolumn{1}{l|}{\multirow{2}{*}{\sf \HSCBM-R}}  & Llama 2   & \multicolumn{1}{l|}{$0.745_{\pm 0.002}$}          & \multicolumn{1}{l|}{$0.638_{\pm 0.027}$}          
    & \multicolumn{1}{l|}{$0.523_{\pm 0.004}$}          & \multicolumn{1}{l|}{$0.611_{\pm 0.011}$}          & $0.705_{\pm 0.009}$          \\ \cline{3-8} 
& \multicolumn{1}{l|}{}                                                                 & Llama 3.1 & \multicolumn{1}{l|}{$0.779_{\pm 0.002}$}          & \multicolumn{1}{l|}{$0.683_{\pm 0.006}$}          & \multicolumn{1}{l|}{$0.574_{\pm 0.010}$}          & \multicolumn{1}{l|}{$0.610_{\pm 0.008}$}          & $0.735_{\pm 0.008}$          \\ \cline{2-8} 
                                           & \multicolumn{1}{l|}{\multirow{2}{*}{${\sf \HSCBMT}$}}                                 & Llama 2   & \multicolumn{1}{l|}{$0.766_{\pm 0.004}$}          & \multicolumn{1}{l|}{$0.658_{\pm 0.008}$}          & \multicolumn{1}{l|}{$0.542_{\pm 0.016}$}          & \multicolumn{1}{l|}{$\mathit{0.723_{\pm  0.016}}$} & $0.709_{\pm 0.104}$          \\ \cline{3-8} 
                                           & \multicolumn{1}{l|}{}                                                                 & Llama 3.1 & \multicolumn{1}{l|}{$0.768_{\pm 0.009}$}          & \multicolumn{1}{l|}{$0.685_{\pm 0.012}$}          & \multicolumn{1}{l|}{$0.551_{\pm 0.013}$}          & \multicolumn{1}{l|}{$0.714_{\pm 0.016}$}          & $\mathit{0.763_{\pm  0.011}}$ \\ \cline{2-8} 
                                           & \multicolumn{1}{l|}{\multirow{2}{*}{\sf HSCBMT-R}} & Llama 2   & \multicolumn{1}{l|}{$0.757_{\pm 0.009}$}          & \multicolumn{1}{l|}{$0.637_{\pm 0.003}$}          & \multicolumn{1}{l|}{$0.526_{\pm 0.023}$}          & \multicolumn{1}{l|}{$0.710_{\pm 0.013}$}          & $0.710_{\pm 0.107}$          \\ \cline{3-8} 
                                           & \multicolumn{1}{l|}{}                                                                 & Llama 3.1 & \multicolumn{1}{l|}{$0.769_{\pm 0.008}$}          & \multicolumn{1}{l|}{$0.666_{\pm 0.012}$}          & \multicolumn{1}{l|}{$0.540_{\pm 0.011}$}          & \multicolumn{1}{l|}{$1.710_{\pm 0.009}$}          & $0.760_{\pm 0.006}$          \\ \hline
\end{tabular}
}
\end{table}

First, when comparing the first two types of baselines (I and II) with traditional ML models, we observe that the latent representations in terms of our adjective concepts significantly improve the prediction performance. 
Surprisingly, with our concept bottleneck representation ($\mathbf{z}$), even traditional ML models can achieve nearly the same level of performance as more complex models, such as LLMs and transformers. The improvements by our representation are achieved for every employed classifier in I and II in every dataset. 
In the case of the CONAN dataset, which contains different types of semantically complex counter-narratives, the increase even reaches over 60\%. This indicates that our proposed concept bottleneck representation can effectively capture complex semantics, including different intentions expressed in the input texts.

Our second observation is that the choice of LLMs to evaluate concepts impacts the final performance. This can be seen in both the conventional ML models (II) and our \HSCBM~models (V). \revised{With a later version of Llama, \ie~Llama 3.1, which demonstrated stronger performance in benchmarks than Llama 2, we obtain overall better prediction accuracy. This improvement of Llama 3.1 can be attributed to architectural enhancements, potentially an increased parameter count, a new tokenizer, and access to more high-quality pretraining data.}
The difference in the capabilities of  LLMs of different versions has a significantly higher impact on conventional ML models compared to our CBM-based models.
The performance improvement, however, varies between datasets.
For instance, when applying SVMs, the performance differences between Llama 2 and Llama 3.1-based representations range from 12\% in GermEval to 31\% in ELF22.

Third, transformers fine-tuned on the target datasets (category III) for the respective tasks yield promising prediction performance for all datasets. Their performance is consistent across the datasets, which may stem from the similar architecture of the models. Their main drawback is (i) that they are opaque in their decision-making process and  (ii) that they are large in numbers of parameters (\eg~XLM-large has $\sim560$ million parameters compared to $\sim0,076$ million in \HSCBM). 
When applying even more powerful LLMs in a zero-shot setting (category IV), 
we observe a different behaviour. The LLMs do not produce consistent and reliable results. An outlier is GPT 4o on GermEval, which produces extraordinarily high results, which may suggest that GermEval is part of the GPT 4o training set (which cannot be clarified conclusively since GPT 4o is not open-source). For all other datasets, the examined LLMs achieve the worst overall results compared to the other categories in the zero-shot setting. 
In contrast, when using ICL, by adding five randomly sampled labelled data instances as contextual information, we observe a considerable improvement in prediction performance across all datasets.
These results indicate the ability of GPT 4o to capture the contextual information embedded in prompts.
In addition, we examine GPT o3-mini, a reasoning LLM trained with a built-in CoT mechanism that increases the likelihood of generating tokens leading to a correct answer.
The reasoning model outperforms GPT 4o in the zero-shot setting on three datasets but performs worse than GPT 4o with ICL across all datasets. 
Thus, introducing CoT reasoning does not lead to notable performance improvements for the investigated datasets.
Overall, these results suggest that, for NLP tasks such as hate and counter speech recognition, the performance of LLMs remains limited without fine-tuning or more advanced prompting strategies. \revised{An interesting observation is that, on the HS-CS dataset, zero-shot evaluation using LLMs perform even worse than the random baseline. 
One explanation for this result might be the nature of the HS-CS dataset, which comprises three classes (\texttt{counter hate,} \texttt{hate speech}, and \texttt{neutral}). Although these classes represent different communicative intentions, a deeper investigation of the data reveals that all (even the \texttt{neutral} class) partly contain  toxic and negative language. This leads to ambiguities that make it difficult for the models to accurately distinguish between the classes. This challenge is not unique to LLMs but is also evident across other methods evaluated on the dataset.
}

The results of our proposed models are summarised in category V in Table~\ref{tab:highestval}. For reference, overall, the highest average performance across all datasets in our experiments is achieved by the fully fine-tuned large-scale transformer model XLM-RoBERTa, with an average macro-F1 of 0.713. Our methods, \HSCBMT~and its regularised variant \HSCBMT-R, rank second, achieving average macro-F1 scores of 0.696 and 0.689 across all datasets, respectively. The much smaller and fully transparent \HSCBM~follows closely with a macro-F1 of 0.685 when using bottleneck representations derived from Llama 3.1. 
We note that the proposed \HSCBM~model, which is based only on an MLP, achieves peak performance on two of the five datasets (ELF22 and HS-CS) and, in these cases, outperforms the base and large transformer models. 
This is in contrast to the existing literature, which suggests that CBM models typically struggle to match or surpass the performance of fully fine-tuned transformer-based models~\citep{SunOUW24,TanCWYLL24}. The reason for this is that bottleneck concepts only provide an abstract representation of the original text based on predefined concepts, which inevitably leads to information loss compared to more complex representations, such as sentence embeddings learned by transformer models. While this information loss is also present in  \HSCBM, it still achieves comparable performance to the strongest transformer model, XLM-RoBERTa, and even outperforms it in two datasets. These results demonstrate the utility of adjective-based concept evaluation for text analysis in the investigated recognition tasks. They further confirm the robustness of the proposed adjective evaluation approach and the strong expressiveness of the resulting representation.

We further note that \HSCBMT~outperforms the \HSCBM~model by approximately $1.1\%$ in average F1-score across all datasets. 
This improvement can be attributed to additional information in the text embeddings generated by the fully fine-tuned transformers. Since this improvement is rather small, we can conclude that most of the useful information for the evaluated NLP tasks is captured by the adjective evaluations in our proposed representation in the \HSCBM~model. The raw text inputs actually add only very limited additional information. 

A related question is, how complementary our representation is to that of a transformer. We can investigate this by comparing the results of XLM-RoBERTa-base (directly fine-tuned on the datases, IV) and the \HSCBMT~and \HSCBMT-R~models, which combine XLM-RoBERTa-base embeddings with our representation (V). In four of five datasets (all except for CONAN), \HSCBMT~and \HSCBMT-R~models clearly outperform the fine-tuned XLM-RoBERTa-base model. This demonstrates that our proposed concept bottleneck representation provides complementary information to the transformer-based embedding and that this complementary benefit can be leveraged well by the \HSCBMT~architecture from Section~\ref{ssec:classification}.

We further compare the MLP model from category II  to the  \HSCBM~model to investigate the effect of the relevance 
gate added to \HSCBM, which refines the importance of the adjectives given by the LLM (see Section~\ref{ssec:classification}). The relevance gate clearly improves the model's ability to identify the most important attributes for classification, resulting in an average performance increase of 3.7\% across all datasets.

Finally, we investigate the effect of adding our class-discriminative regularisation term to the loss function. The term guides the model to rely on a rather small set of highly discriminative adjectives for encoding input texts to improve the compactness and human interpretability of the representation. The regularisation causes a slight decrease in recognition performance, which is actually expected, as increased interpretability typically goes along with a cost of performance.  A closer look, however, reveals that this loss in performance is rather negligible, with approximately 1\% on average for both models \HSCBM-R~and \HSCBMT-R.  

\begin{table}[t]
    \centering
    \caption{Performance comparison of our approaches, \HSCBM~ and \HSCBMT, against state-of-the-art (SOTA) methods for each evaluated dataset.}
\resizebox{\linewidth}{!}{
    \begin{tabular}{llllll}
        \hline
        \multirow{2}{*}{\textbf{Method}} & \multicolumn{5}{c}{\textbf{Dataset}}\\
        \cline{2-6}
        &\multicolumn{1}{l}{\textbf{GermEval}} & \multicolumn{1}{l}{\textbf{ELF22}}   & \multicolumn{1}{l}{\textbf{HS-CS}}   & \multicolumn{1}{l}{\textbf{CONAN}}   & \textbf{TSNH}    \\ \hline
         \HSCBM & \textbf{0.781}& \textbf{0.693}& 0.581 & 0.630 & 0.739\\
         \HSCBMT & 0.768 & 0.685 & 0.551 & \textbf{0.714} & \textbf{0.763}\\
         \textit{SOTA} & 0.718\textsuperscript{\citep{bornheim-etal-2021-fhac}} & 0.640\textsuperscript{\citep{LeeNSSP22}} & \textbf{0.610}\textsuperscript{\citep{yu-etal-2022-hate}} & 0.620\textsuperscript{\citep{poudhar-etal-2024-strategy}} & 0.717\textsuperscript{\citep{bock2024exploring}}\\
         \hline
    \end{tabular}
    }
    \label{tab:sota_comparison}
\end{table}
\paratitle{Comparison to the literature.} 
Due to the large number of prior works related to hate and counter speech recognition, it is not practical to conduct an exhaustive comparison. 
Instead, we focus on comparing our proposed models to the best reported performance in the literature for the investigated datasets, as summarized in Table~\ref{tab:sota_comparison}. 
The mainstream approach among the state-of-the-art (SOTA) models~\citep{labadie2023everybody,dimitrov2024semeval,kirk-etal-2023-semeval,yu-etal-2022-hate,LeeNSSP22,poudhar-etal-2024-strategy} typically involves fine-tuning transformer-based models and augmenting the training data to achieve the highest performance on these datasets.

Our \HSCBM~model clearly outperforms existing transformer-based baselines on four of the five investigated datasets. For instance, on the GermEval dataset, our model achieved a macro-F1 score of 0.781, outperforming the score of 0.718 reported by~\citep{bornheim-etal-2021-fhac}, who fine-tuned an ensemble of BERT-based models. 
For the ELF22 dataset, our model achieved a macro-F1 score of 0.693, surpassing the baseline of 0.64, obtained by a fine-tuned RoBERTa model in~\cite{LeeNSSP22}.
\subsection{Interpretability analysis}
\label{subsec:interpretabilityResult}
In this section, we analyse the interpretability offered by our proposed models. 
We apply \HSCBM~to the HS-CS dataset, which provides a sufficiently complex scenario thanks to the availability of contextual information. The three labels used to annotate the dataset are \texttt{hate speech}, \texttt{counter speech}, and \texttt{neutral speech}. 
One important contribution of this paper is the introduction of the class-discriminative regularisation term to improve the interpretability of our CBM approach. To show the effectiveness of this regularisation, we compare the level of interpretability of our \HSCBM~model in two scenarios: with regularisation and without. 
We explore interpretability at two levels: the global level, which aims to explain the learned class levels and at the local level, which aims to explain individual decisions by the model, \ie~which concepts (or adjectives in our case)  contribute to an individual predictions. 

\begin{figure}[t]
  \subfigure[Local explanations without class-discriminative regularisation.
  \label{fig:exp2a}]{%
    \includegraphics[width=\linewidth]{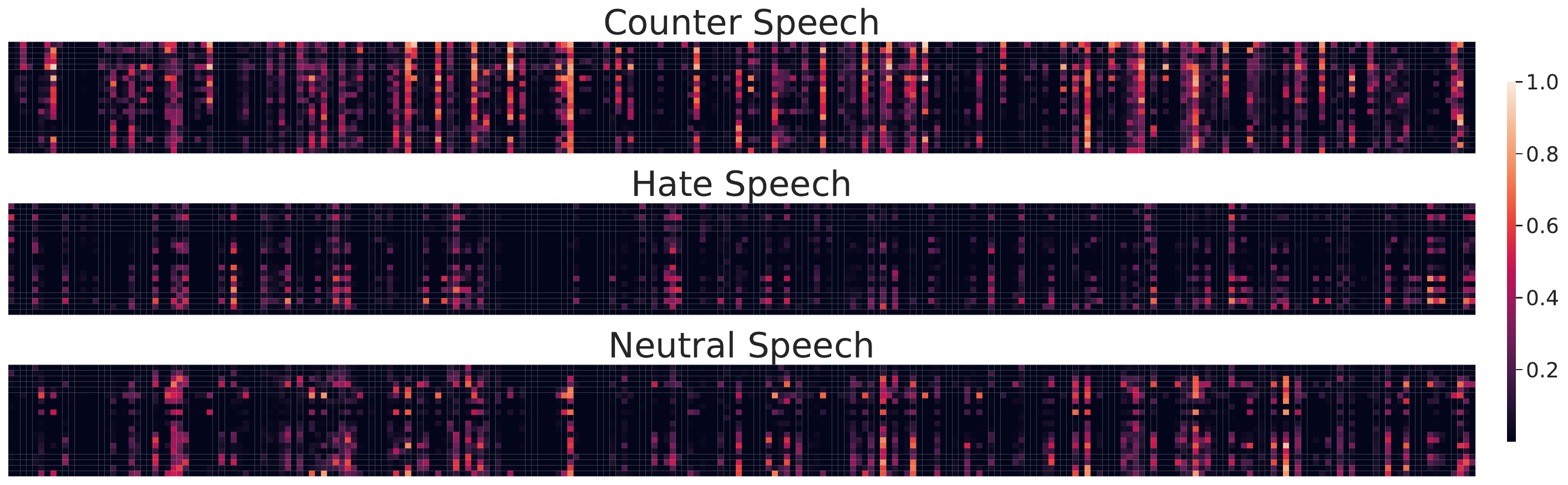}%
  }%
  \hfill
  \subfigure[Local explanations with class-discriminative regularisation\label{fig:exp2b}]{%
    \includegraphics[width=\linewidth]{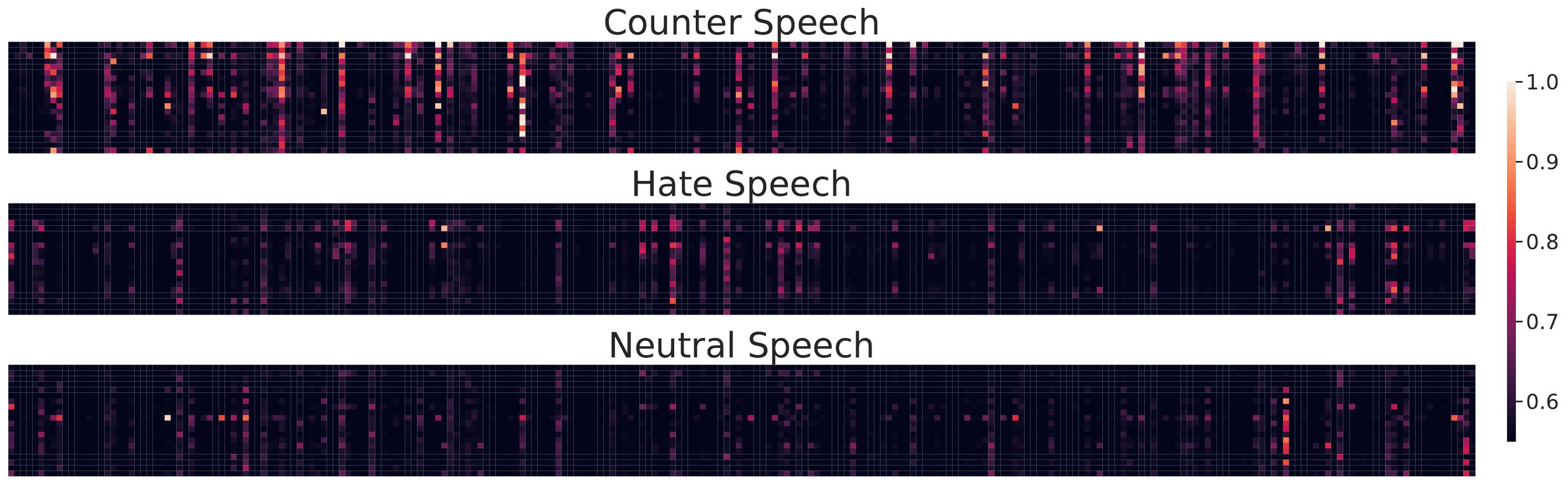}
  }
  \caption{Contribution of adjectives (x-axis) to the prediction of the top-20 (y-axis) most confidently predicted instances in each class. Darker colours indicate lower relevance, while lighter colours represent higher relevance scores.}
  \label{fig:top_activations_heatmap}
\end{figure}

\paratitle{Local (decision-level) interpretability.}
Since our \HSCBM~model is, per definition, self-explainable, we can simply derive local explanations directly from the latent encoding, \ie~$z_i$ for a given input text $t_i$. 
Figure~\ref{fig:top_activations_heatmap} shows the top-20 
most \emph{confidently} and \emph{correctly} predicted instances for each class with corresponding local explanations.
\figureautorefname~\ref{fig:exp2a} shows the local explanations of \HSCBM~trained without, and Figure~\ref{fig:exp2b} with, our proposed class-discriminative regularisation.

The vertical line-like structures in Figure~\ref{fig:top_activations_heatmap} occur in cases where the same set of adjectives is relevant for predicting the majority of the investigated text samples. These patterns indicate that the \HSCBM~model utilises consistent adjective sets to characterise the text samples of the same class.
Moreover, when comparing the explanations across different classes, we observe that each class has its own consistent adjective set. Similar to our analysis of global explanations, when the class-discriminative regularisation 
is applied, the latent text encoding becomes more sparse (see larger dark regions in Figure~\ref{fig:exp2b}) compared to the model trained just on the cross-entropy loss. Furthermore, the number of shared adjectives across the classes is strongly reduced.

\begin{table}[t]
    \fontsize{8pt}{9pt}\selectfont
    \centering
    \caption{Top-10 most relevant adjectives for individual input samples from each class of the HS-CS dataset provided by \HSCBM. For comparison, we provide LIME explanations for the same samples generated from the fine-tuned XLM-RoBERTa model. }
    \label{tab:instances_descriptor}
    \begin{tblr}{
                  colspec={Q[t, 3.8em]|p{7.2cm}|p{4cm}},
                  hline{1,3, 4, 5, 6},
                }
         \textbf{Class}& \textbf{Input} & \textbf{Adjectives}  \\
         \hline
         Counter- speech&{CONTEXT: From \colorbox{cs_color!10}{what} I \colorbox{cs_color!35}{read} the \colorbox{cs_color!10}{movie} \colorbox{cs_color!50}{is} severely \colorbox{cs_color!20}{inaccurrate} \colorbox{cs_color!10}{and} the only redeeming feature \colorbox{cs_color!50}{is} \colorbox{cs_color!30}{Rami} \colorbox{white}{Maleks performance.}\\
         \colorbox{white}{COMMENT:}  I wasn't \colorbox{cs_color!10}{familiar} \colorbox{cs_color!10}{enough} \colorbox{cs_color!25}{with} Queen to spot the \colorbox{cs_color!15}{inaccuracies}, so I \colorbox{cs_color!40}{enjoyed} \colorbox{cs_color!10}{it} a \colorbox{cs_color!10}{ton}.}&{\fontsize{8pt}{12}\selectfont differentiating,  moderating, \newline conciliatory,  amused, \newline promoting,  admiring, \newline exclusionary,  respectfully, \newline balancing,  emotionally} \\
         
         Hate Speech & {CONTEXT:  Damn, this is some \colorbox{hs_color!30}{cringey} \colorbox{hs_color!28}{neckbeard} shit  Y'all \colorbox{hs_color!10}{lived} up \colorbox{hs_color!10}{to my} expectations and \colorbox{white}{didn't} disappoint hit me up \colorbox{hs_color!10}{if} you \colorbox{hs_color!10}{wanna} know how \colorbox{hs_color!10}{a} \colorbox{hs_color!25}{vagina} feels\\
         COMMENT: You’re \colorbox{hs_color!10}{a} {fucking} \colorbox{hs_color!20}{retard}} & {\fontsize{8pt}{12}\selectfont hurtful,  gender discriminatory, \newline inappropriate,  unacceptable, \newline exclusionary,  hostile, \newline disrespectful,  abusive, \newline insensitive, sexist}\\
         
         Neutral Speech & {CONTEXT: Almost no \colorbox{ns_color!5}{one} on \colorbox{ns_color!3}{a} train \colorbox{ns_color!1}{or} \colorbox{ns_color!3}{subway} is displaying \colorbox{ns_color!5}{dominance}. \colorbox{ns_color!50}{You} \colorbox{ns_color!5}{are just} \colorbox{ns_color!7}{looking} for \colorbox{ns_color!2}a \colorbox{ns_color!50}{dumb} \colorbox{ns_color!10}{debate}.\\
         COMMENT:  \colorbox{ns_color!50}{You} \colorbox{ns_color!5}{my} \colorbox{ns_color!2}{friend} have \colorbox{ns_color!10}{never} \colorbox{ns_color!5}{been} on \colorbox{ns_color!5}{the red} line in Chicago \colorbox{ns_color!5}{south} of \colorbox{ns_color!5}{Roosevelt}.}&{\fontsize{8pt}{12}\selectfont worried, \newline refuting,  condescending, \newline unfair,  impious, \newline conciliatory,  exclusionary, \newline expressing concern, \newline unnecessary,  insulting} \\
    \end{tblr}
\end{table}

To investigate the local explanations in more detail, we generate explanations comprising the top-10 most relevant adjectives for a given input from the \HSCBM~and \HSCBM-R~models and compare them with explanations obtained via  LIME~\citep{molnar2022interpretable} for  
predictions made with XLM-RoBERTa. LIME is a widely used XAI method that provides model-agnostic post-hoc explanations, showing the relative contribution of each token to the prediction. 
In Table~\ref{tab:instances_descriptor}, we use one instance from each class of the HS-CS dataset as an example for the comparison.
Note that the LIME explanations are represented in the ``Input'' column by colour encodings that indicate the level of relevance for each token. Thereby, more saturated colours 
indicate stronger relevance for the decision. 
Recall that an instance in the HS-CS dataset consists of two parts denoted by ``CONTEXT'' 
and ``COMMENT'', respectively. The comment part contains the actual content that is finally classified.

Most of the adjectives considered relevant by the model are indeed consistent with the semantics and pragmatics of the content, especially when the provided context is taken into consideration when interpreting the statements. Compared to LIME explanations, the appropriateness of individual adjectives can easily be verified by a human by just interpreting the text.  
The LIME explanations, in contrast, reside on a lower level of abstraction (\ie~the level of the raw text), which makes them more challenging to interpret. Additionally, they are often questionable 
(\eg~highly relevant word ``is'' for counter speech and ``you'' for neutral speech), which represents a problem that has been previously reported in the literature~\citep{ribeiro2018anchors,madsen2022post}. 
The proposed approach provides explanations at a higher semantic level, making it easier for a human to assess. The adjectives enable us to quickly verify the correct reasoning behind an individual prediction so that falsely learned patterns can also be identified more easily. 
For further examples from this experiment, refer to Appendix~\ref{app:lime}.

We notice that the underlying LLM in \HSCBM~does not completely separate the information contained in the comment from the context, even though the prompt only asks for adjectives related to the comment. For example,  ``gender discriminatory'' and ``sexist'' in the hate speech example in Table~\ref{tab:instances_descriptor} actually describe the context, but they are attributed to the comment being classified. This may be due to the strong context awareness of LLMs, a property that can be highly 
useful when the proposed approach is applied to categorizing, for example, entire communication strands. 

\begin{figure}[t]
  \subfigure[\label{fig:exp1a}Average relevance scores without  class-discriminative regularisation.]{\includegraphics[width=\linewidth]{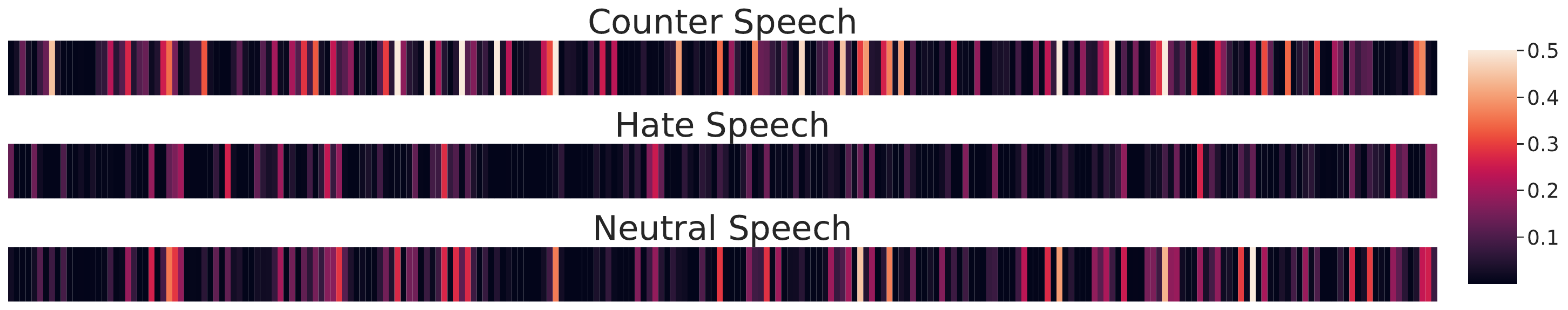}}
  \subfigure[\label{fig:exp1b}Average relevance scores with  class-discriminative regularisation.]{
    \includegraphics[width=\linewidth]{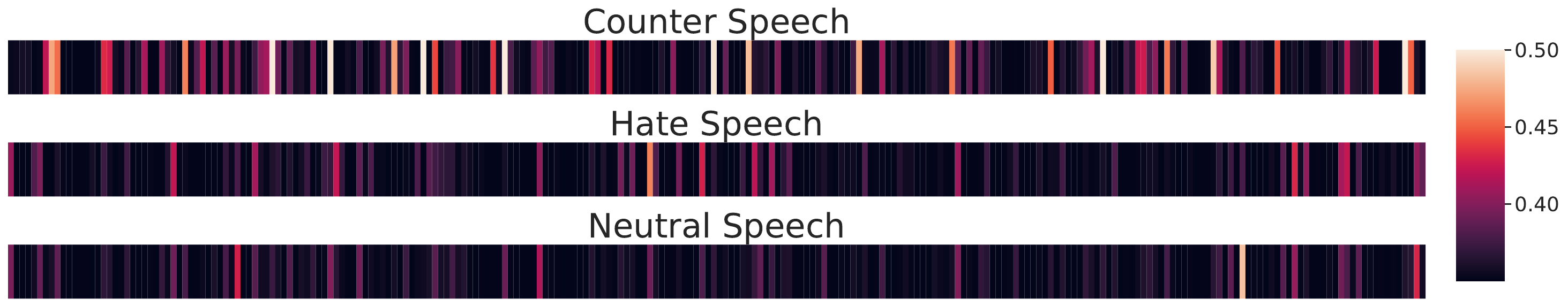}%
  }
  \caption{Mean contribution per adjective (x-axis) of correctly predicted instances of each class. Each vertical line represents an individual adjective, with colours indicating its relevance score. Darker colours indicate lower relevance, while lighter colours represent higher relevance scores. The distribution of relevance values across different classes suggests that without regularization, the model assigns importance to a wide range of adjectives, often overlapping between classes, rather than focusing on class-discriminative concepts. Regularisation strongly counteracts this circumstance.}
  \label{fig:activations_heatmap}
\end{figure}

\paratitle{Global interpretability.}
To provide a global explanation for \HSCBM, we propose aggregating local explanations to identify the adjectives that contribute most to predicting specific classes. The global contribution of an adjective to a particular class is computed as the mean value of its activations across all instances in the training set belonging to that class.
\figureautorefname~\ref{fig:activations_heatmap} comprises subfigures for each class, illustrating the average contribution of each adjective for correctly predicted instances in the training data. 
\figureautorefname~\ref{fig:exp1a} depicts these contributions when \HSCBM~is optimised only with the cross-entropy loss function, while \figureautorefname~\ref{fig:exp1b} shows the contributions when the class-discriminative regularization term is applied.

Our first observation is that the number of adjectives that are highly relevant to the predictions drastically decreases when class-discriminative regularization is applied.
The resulting latent encodings of input texts thus become more sparse.  
Another observation is that when trained with class-discriminative regularisation, fewer adjectives are shared by multiple classes, making the adjectives more discriminative, the representation sparser, and explanations derived from it more compact and easier to grasp. A positive effect of the regularisation is thus well visible.

\tableautorefname~\ref{tab:class_descritors} shows the top-8 most relevant adjectives for each class for the \HSCBM~model with and without the class-discriminative regularisation. We observe that the adjectives corresponding to hate speech and counter speech are reasonable and easy to interpret.
For the neutral class, we observe that the listed adjectives partly overlap with the other two classes (\eg~questioning, expressive, rough, incomprehensible, mocking and challenging). This overlap is expected because, in the given dataset, the neutral class also contains toxic and harmful language. 
This overlap, however, is reduced when utilizing the class-discriminative regularisation, leading to more expressive and distinct class models. 

\begin{table}[t]
    \centering
    \fontsize{8pt}{9pt}\selectfont
    \caption{Most relevant adjectives for each class (\ie~global explanation) in the HC-CS dataset for the \HSCBM~model, with and without class-discriminative regularization.}
    \label{tab:class_descritors}
    \begin{tabular}{c|p{3.1cm}p{3.1cm}p{3.1cm}}
        \hline
         \multirow{2}{*}{\bf Model} & \multicolumn{3}{c}{\bf Class} \\
         \cline{2-4}
         & {\bf Hate Speech} & {\bf Counter speech} & {\bf Neutral Speech}\\
         \hline\hline
         \HSCBM & disrespectful, \newline vile,  \newline controversial,  \newline  sexist,  \newline  repulsive,  \newline  condescending,  \newline  mocking,  \newline  degrading  & questioning,  \newline  expressive,  \newline  rough,  \newline  differentiating,  \newline  dialogue-oriented,  \newline  incomprehensible,  \newline  mockingly,  \newline  challenging & stirring up conflict,  \newline  mockingly,  \newline  questioning,  \newline  rough,  \newline  challenging,  \newline  neutralizing,  \newline  expressive,  \newline  incomprehensible\\
         \hline
         \HSCBM-R & hateful,  \newline  passive-aggressive,  \newline  defamatory,  \newline  skeptical,  \newline  discreditable,  \newline  prejudiced,  \newline  contrasting,  \newline  aggressive & refuting,  \newline  exclusionary,  \newline  worried,  \newline  sarcastic,  \newline  incomprehensible,  \newline  differentiating,  \newline  condescending,  \newline  neutralizing & exclusionary,  \newline  refuting,  \newline  worried,  \newline  impious,  \newline  promoting aggression,  \newline  expressive,  \newline  condescending, \newline unfair\\
         \hline
    \end{tabular}
\end{table}

\subsection{User-centered evaluation of local interpretability}
\label{subsec:userstudy}

To evaluate the usefulness of the local explanations produced by our \HSCBM~model, we conduct a small-scale user study. We distributed a structured 
questionnaire to 21 colleagues from our institution who were not involved in this research and received 15 complete responses.

Each participant received one of three variants of the questionnaire containing eight randomly selected examples from the CONAN dataset. We chose the CONAN dataset for this study because it is one of the few publicly available datasets that provide labelled counter speech responses across multiple strategies, along with the preceding hate speech comments as context, which makes it particularly well-suited for evaluating model behaviour and interpretability in a real-world setting. Out of the initial seven classes, we selected four, i.e., \texttt{facts}, \texttt{humour}, \texttt{denouncing}, and \texttt{question}, to make the task more feasible for users. To this end, the class \texttt{hypocrisy} often included content similar in form to \texttt{question}, making it harder to distinguish. Other classes were excluded because they did not contain actual counter speech, such as \texttt{support}, which includes content in favour of the hate speech, and \texttt{unrelated}, which contains content unrelated to the hate speech.
For each of the four classes, we randomly selected two examples for a questionnaire. For every example, the file showed the hate speech (context), the counter speech, and the predicted class. Additionally, for each example, five adjectives were listed which our \HSCBM~model identified as the most relevant adjectives for the model’s prediction. Participants were asked to assess the relevance of each adjective by marking one of three columns: \textit{irrelevant}, \textit{maybe relevant}, or \textit{relevant}.

Across all collected responses, we observed that the majority of adjectives were rated as \textit{relevant} (252 adjectives), followed by \textit{maybe relevant} (180 adjectives) and \textit{irrelevant} (168 adjectives). Figure~\ref{tab:relevance_percentage} presents a breakdown of the results by class. For the classes \texttt{denouncing} and \texttt{facts}, the majority of adjectives were considered \textit{relevant} (51.3\% and 62.0\%, respectively), indicating a strong agreement between the model’s explanations and human assessment in these classes. In contrast, the classes \texttt{humor} and \texttt{question} show a higher proportion of adjectives rated as \textit{irrelevant} (42.7\% and 44.7\%, respectively), while fewer adjectives were marked as \textit{relevant} (30.0\% and 24.7\%, respectively). The category \textit{maybe relevant} is moderately represented in all classes (ranging between 25.3\% and 36.7\%), indicating a certain degree of uncertainty or borderline cases where users were unsure about the relevance of proposed adjectives. 

\begin{table}[h]
    \centering
    \caption{Per-class distribution of adjective relevance as rated by participants.}
    \label{tab:relevance_percentage}
    \begin{tabular}{lccc}
    \hline
    Class & Irrelevant (\%) & Maybe Relevant (\%) & Relevant (\%) \\
    \hline
    \texttt{denouncing} & 12.0 & 36.7 & 51.3 \\
    \texttt{facts}      & 12.7 & 25.3 & 62.0 \\
    \texttt{humor}      & 42.7 & 27.3 & 30.0 \\
    \texttt{question}   & 44.7 & 30.7 & 24.7 \\
    \hline
    \end{tabular}
\end{table}

\begin{figure}[t]
    \centering
    \includegraphics[width=1.0\textwidth]{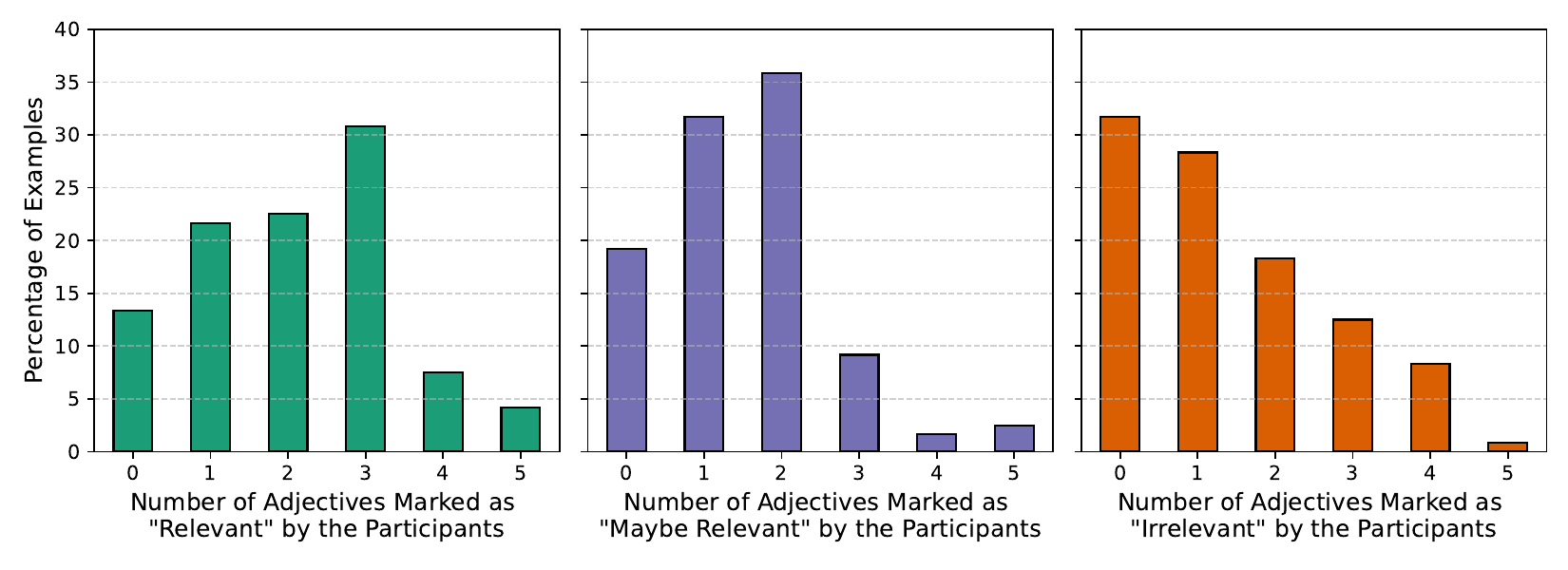}
    \caption{Distribution of user-assigned relevance ratings for the five adjectives per example across all annotated samples. Each histogram shows the number of examples that received 0 to 5 adjectives rated as (left) \textit{irrelevant}, (middle) \textit{maybe relevant}, and (right) \textit{relevant}. 
   } 
    \label{fig:userstudy}
\end{figure}

Analysing responses on a per-example basis (see Figure~\ref{fig:userstudy}) reveals that in only 13.3\% of cases (16 examples), none of the adjectives were rated as \textit{relevant}. The majority of examples had one (21.7\%), two (22.5\%), or three (30.8\%) adjectives rated as \textit{relevant}. A similar pattern is observed for the \textit{maybe relevant} category, though with slightly different proportions. In contrast, for the \textit{irrelevant} category, in 31.7\% of examples, none of the adjectives were rated as \textit{irrelevant} by participants.

These results confirm that the local explanations generated by our \HSCBM model are generally meaningful and interpretable to users, even with minimal context and no background in the specific domain. In addition to the large number of adjectives that were rated as \textit{relevant} or \textit{maybe relevant} (in total 72.0\% of rated adjectives), it is noteworthy that in only a few cases (13.0\% of examples) none of the top five adjectives were considered relevant for a given example.

\subsection{Sensitivity evaluation}\label{subsec:sensible}
We conducted three evaluations to test the sensitivity of our proposed model in terms of the selection of adjectives and prompts queried to LLMs.
In this section, we use GermEval and TSNH as our test datasets. 
Results are based on the performance of \HSCBM~using Llama 3.1 for adjective evaluation.

\paratitle{Sensitivity to perturbation of adjectives.}
The first evaluation aims to evaluate the sensitivity of \HSCBM~with respect to individual adjectives. To test the sensitivity, we hide individual adjectives from the bottleneck representation. We use the permutation importance method~\citep{breiman2001random} to assess whether specific adjectives are essential for the predictions of an already trained and fixed model, or if their presence introduces misleading correlations. 

Given a test dataset (\ie~GermEval or TSNH), we first calculate the bottleneck representation derived from $\LLM$, denoted as $\cVec_i$ for an input text $t_i$.
We randomly permute the values of the investigated adjective across all the input texts while the values of all the other adjectives remain unchanged. 
This approach creates a corrupted version of the data while maintaining the overall distribution of the adjective’s relevance values, as observed during training. By introducing noise to the specific adjective under investigation, we can assess its contribution to the model’s predictions and determine whether it holds meaningful discriminative power or introduces misleading correlations (without the need to retrain the entire model, making this approach computationally efficient).

\begin{figure}[h!]
  \subfigure[\label{fig:feature_corr_tsnh}TSNH Task]{\includegraphics[width=0.5\linewidth]{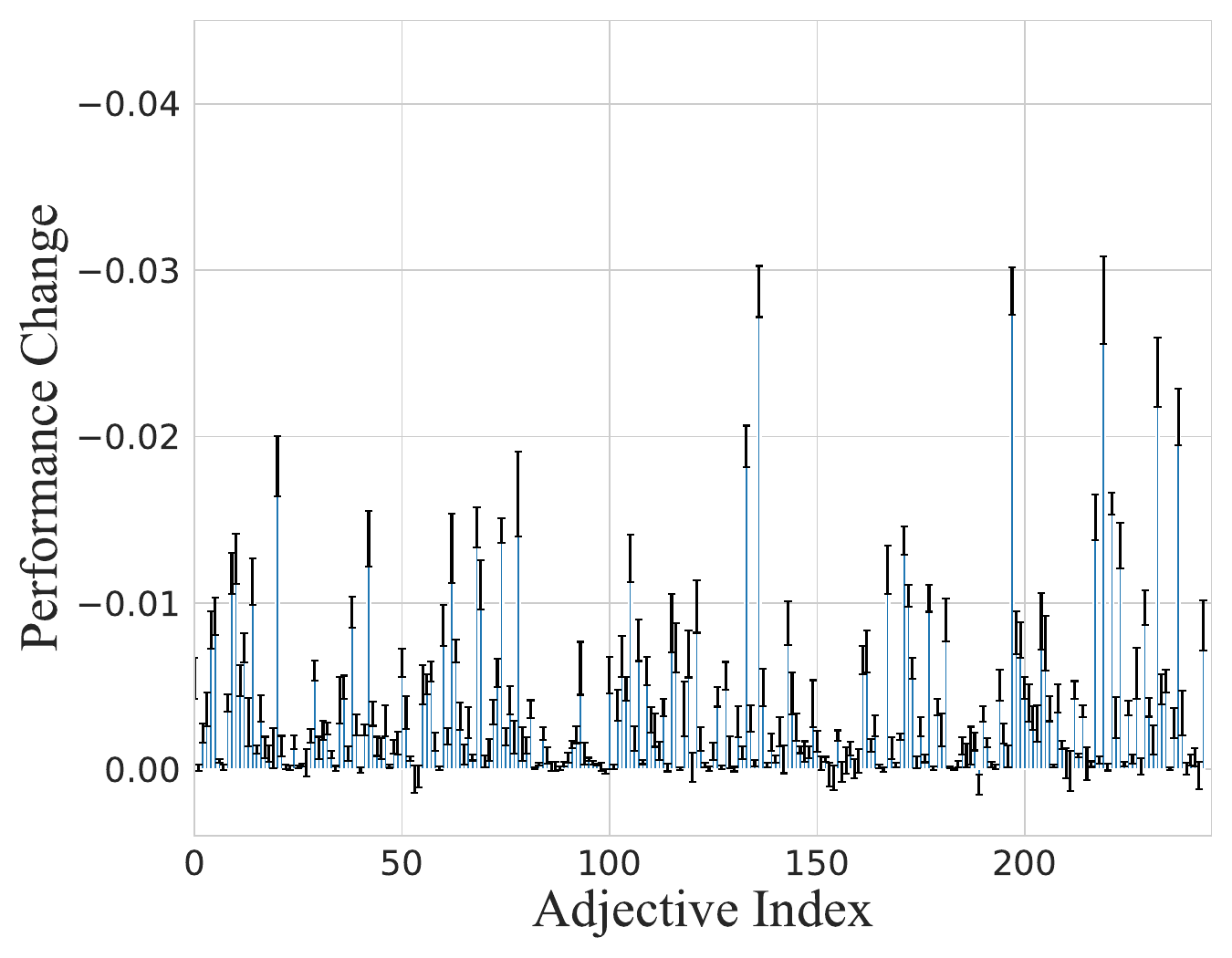}}
  \hspace{-1mm}
  \subfigure[\label{fig:feature_corr_germ}GermEval Task]{
    \includegraphics[width=0.5\linewidth]{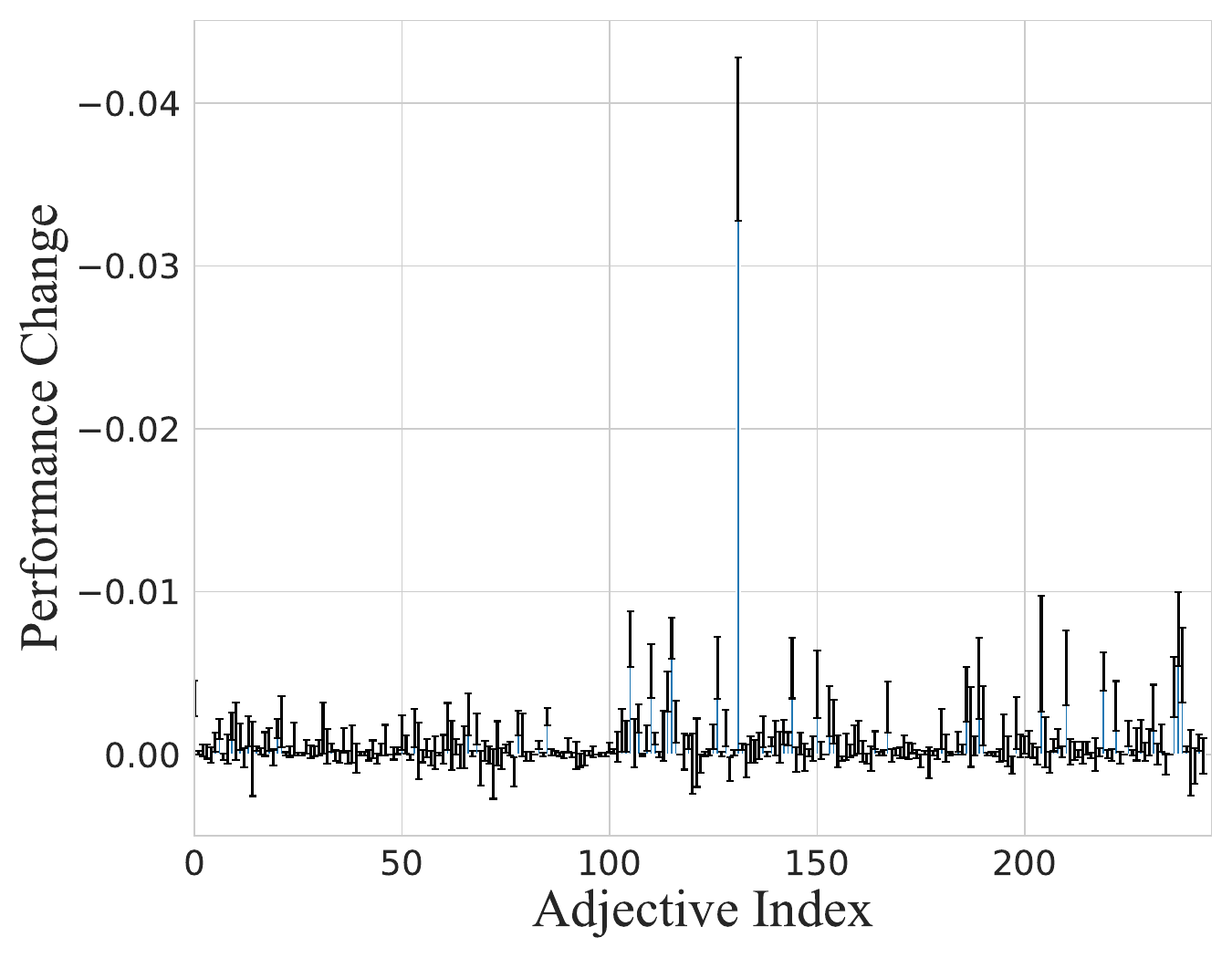}
  }
  \caption{Impact of an individual adjective on the model performance in terms of change in macro-F1 score. }
  \label{fig:feat_sensitiv}
\end{figure}

Figure~\ref{fig:feat_sensitiv} shows the mean performance variation in terms of the macro-F1 score along with its standard deviation after applying the permutation importance method for each adjective. To ensure robustness in our analysis, we performed 10 random permutations of the values for each adjective across the test dataset.
Overall, the permutation of individual adjective values causes only a 
marginal decrease in performance, \ie~$-9\times10^{-4}$ and $3.8\times10^{-3}$ on average for GermEval and TSNH datasets, respectively.
Nevertheless, one particular adjective, \ie~``insulting'' (or ``beleidigend'' in German), stands out for the GermEval dataset in Figure~\ref{fig:feat_sensitiv}b. When its values were permuted, the performance dropped on average $-0.037\pm 0.005$. This is noteworthy because the task involved recognizing offensive content, and the presence of the adjective ``insulting'' appears to be a strong feature in distinguishing offensive from non-offensive sentences

\begin{figure}[h!]
  \subfigure[\label{fig:set_corr_tsnh}TSNH Task]{\includegraphics[width=0.5\linewidth]{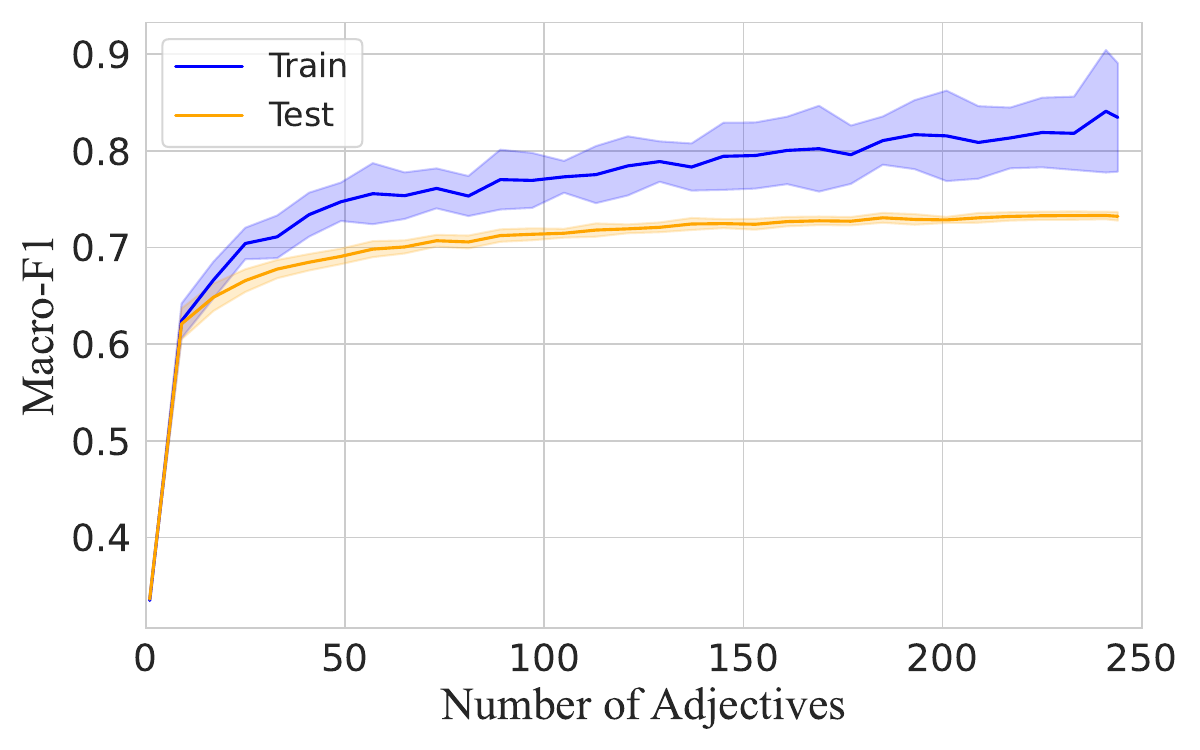}}
  \hspace{-2mm}
  \subfigure[\label{fig:set_corr_germ}Germeval Task]{
    \includegraphics[width=0.5\linewidth]{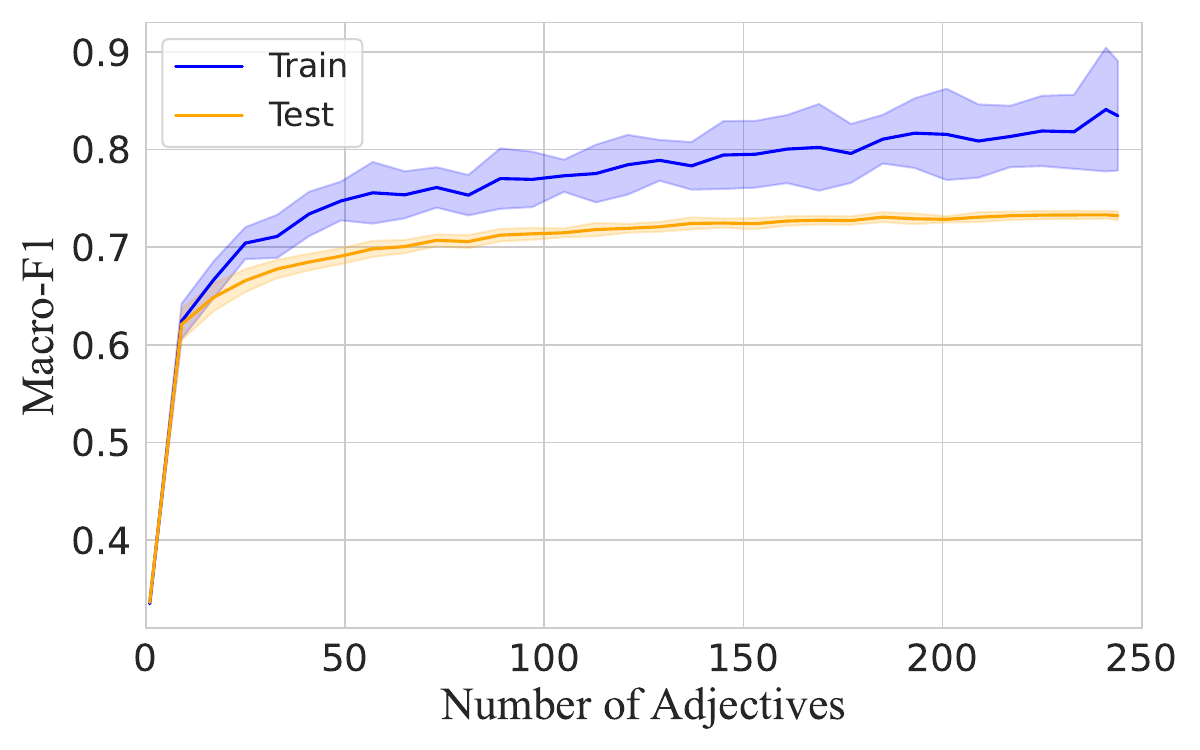}
  }
  \caption{Impact of differently large adjective sets (bottleneck layer dimensions) on the training (blue) and test (yellow) performance of \HSCBM~in terms of macro-F1 (averaged over 100 randomly selected subsets for each evaluation iteration). The shaded area represents the standard deviation.}
  \label{fig:feature_selection}
\end{figure}

\paratitle{Sensitivity to the amount of adjectives.} The second evaluation investigates how the number of adjectives from $\adjSet$ used to construct the bottleneck representation influences the learning and generalization capabilities of \HSCBM.
To systematically explore this dependency, we create bottleneck representations using progressively larger subsets of adjectives, starting with just one adjective and increasing incrementally until reaching the full set $\adjSet$, \ie~$\vert\adjSet\vert$.
In each iteration, we randomly construct 100 subsets and train \HSCBM~with a bottleneck representation relying on these adjectives and evaluate its performance. 
Figure~\ref{fig:feature_selection} illustrates how the number of adjectives affects the training and test performance in terms of macro-F1 (averaged over the 100 subsets for each iteration of the evaluation). 

The results show that for both datasets, the test performance stabilizes at around 100 adjectives and does not increase considerably thereafter. However, during training, performance continues to improve with larger adjective sets due to their increased descriptive power, which could indicate a certain degree of overfitting. Furthermore, the stagnation of the testing performance may also be an indicator of an increasing redundancy in the adjective set.

\paratitle{Sensitivity to prompt definition.}
Prompts used in our approach contain a prefix that provides some contextual information, such as the persona profile. Defining this persona profile should guide the LLM to adapt its answers to align with the characteristics of a human with that profile, \eg~a social psychologist~\citep{KongSH2024,giorgi2024human}.
In addition to investigating different personas, we adopt a classic ICL strategy by providing adjectives' definitions as context to examine our models' performance in the few-shot setting. 

To investigate the the impact of persona profiles on our \HSCBM~model, we define eight distinct personas, along with a setting in which no persona is provided to the LLM. 
\tableautorefname~\ref{tab:persona_description} lists the persona profiles evaluated in our experiments. First, we examine the differences in adjective scores across the various persona profiles, and then we assess their effect on the overall performance of \HSCBM.
\begin{table}[h]
    \centering
    \caption{Persona profiles utilized to evaluate the sensitivity of our \HSCBM~model to the prompt selection.}
    \label{tab:persona_description}
    \begin{tabular}{cp{11.5cm}}
    \hline
         ID & Persona Description  \\
         \hline
         1 & ``You are an expert in social psychology. When you are asked a question, you prefer to give short, concrete answers.''\\
          2 & ``You are an expert in social psychology.''\\
        3 & \textbf{No persona profile defined.}\\
        4 &  ``You are a linguist.''\\
        5 & ``You are a content moderator.''\\
        6 & ``You are a psychologist.''\\
        7 & ``You are a social media expert.''\\
        8 & ``You are a political scientist.''\\
        9 & ``You are a sociologist.''\\
        \hline
    \end{tabular}
\end{table}

\begin{figure}[h!]
    \hspace{-5mm}
  \subfigure[\label{fig:corr_tsnh}TSNH Task]{\includegraphics[width=0.53\linewidth]{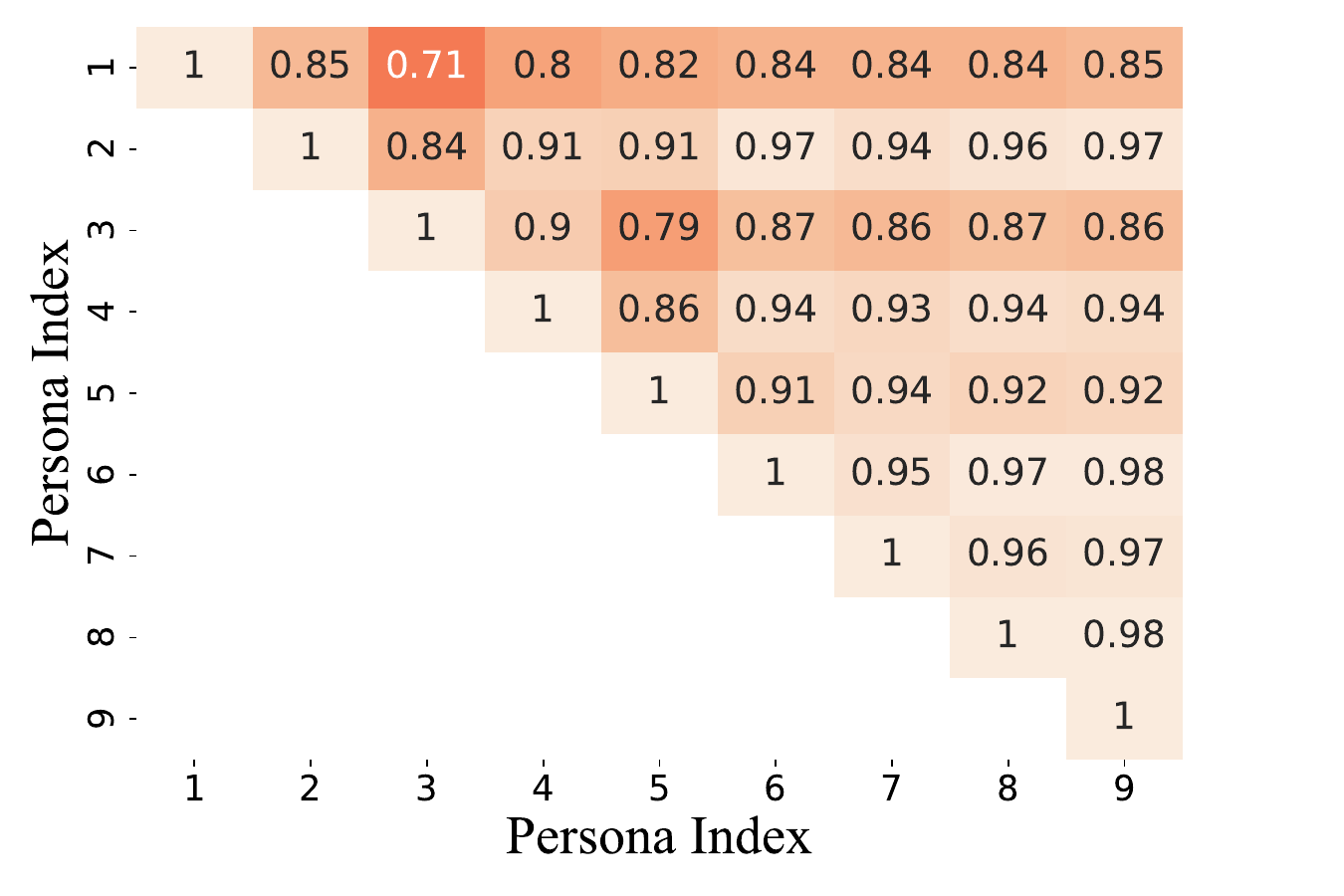}}
  \hspace{-10mm}
  \subfigure[\label{fig:corr_germ}Germeval Task]{
    \includegraphics[width=0.53\linewidth]{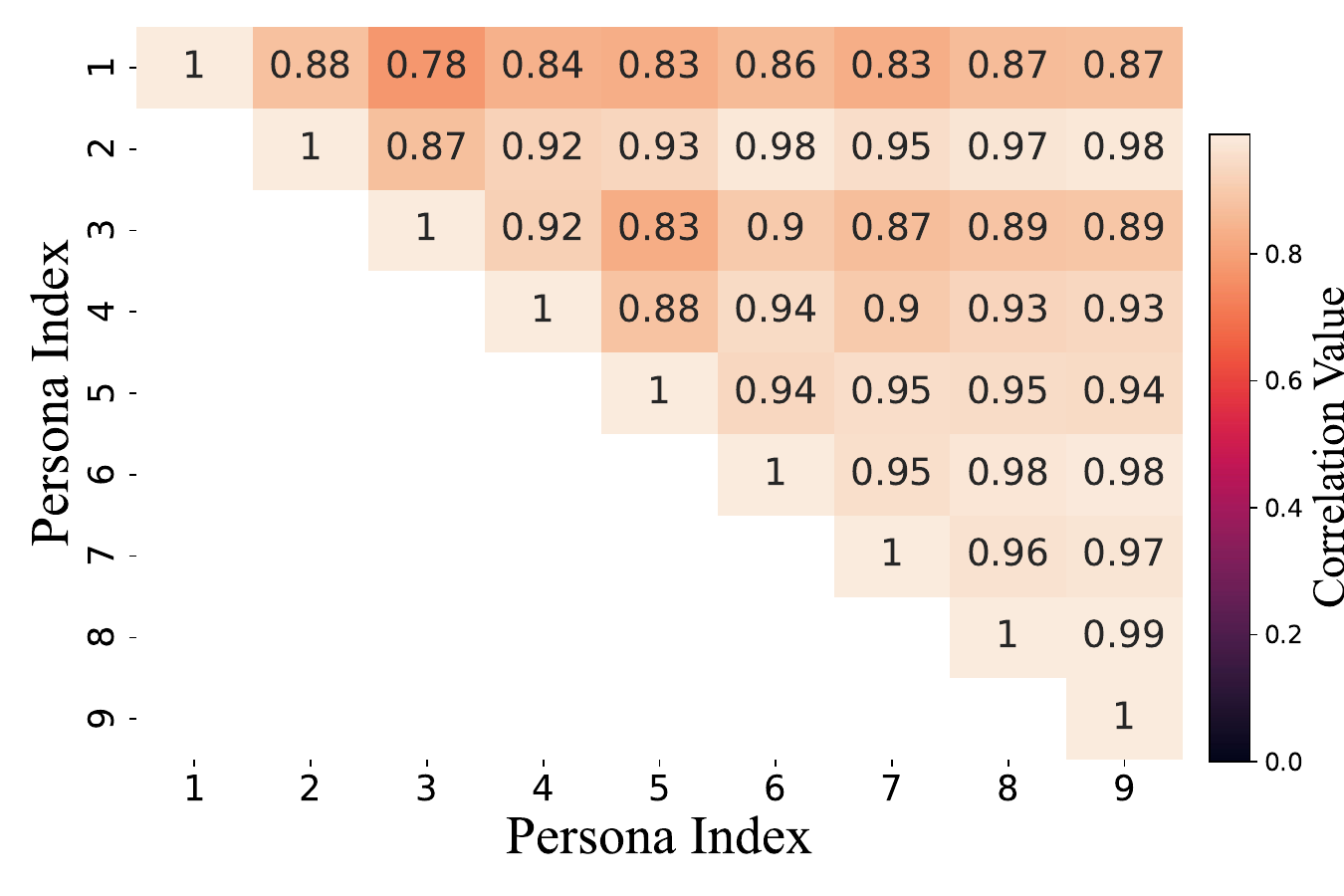}
  }
  \caption{Pairwise Pearson correlation between the concept bottleneck representations of \HSCBM~using nine different persona profiles in the prompt definition. Brighter colours indicate a higher correlation.}
  \label{fig:persona_corre}
\end{figure}

Figure~\ref{fig:persona_corre} shows the pairwise Pearson correlation between the adjective scores of \HSCBM~using nine different persona profiles in the prompt definition, shown as a heatmap.
The correlations are obtained with a p-value $\ll 0.001$  due to the large number of observations. We observe very high positive correlations (0.90 and 0.91 on average for TSNH and Germeval datasets, respectively) between any 
two persona profiles. The lowest correlation coefficients are typically observed in comparisons where no persona profile is defined (persona index 3), but they still show a high positive correlation (with the lowest correlation coefficient being $r=0.71$). 
These results show that the selection of different persona profiles has little impact on the concept bottleneck representations.
Moreover, the very high correlation between persona 1 and persona 2 shows that the representation is also robust to the length of LLM responses (persona 1 explicitly asks for short and concrete answers while persona 2 does not).

\begin{figure}[t!]
  \subfigure[\label{fig:prompt_sens_corr_tsnh}TSNH Task]{\includegraphics[width=0.5\linewidth]{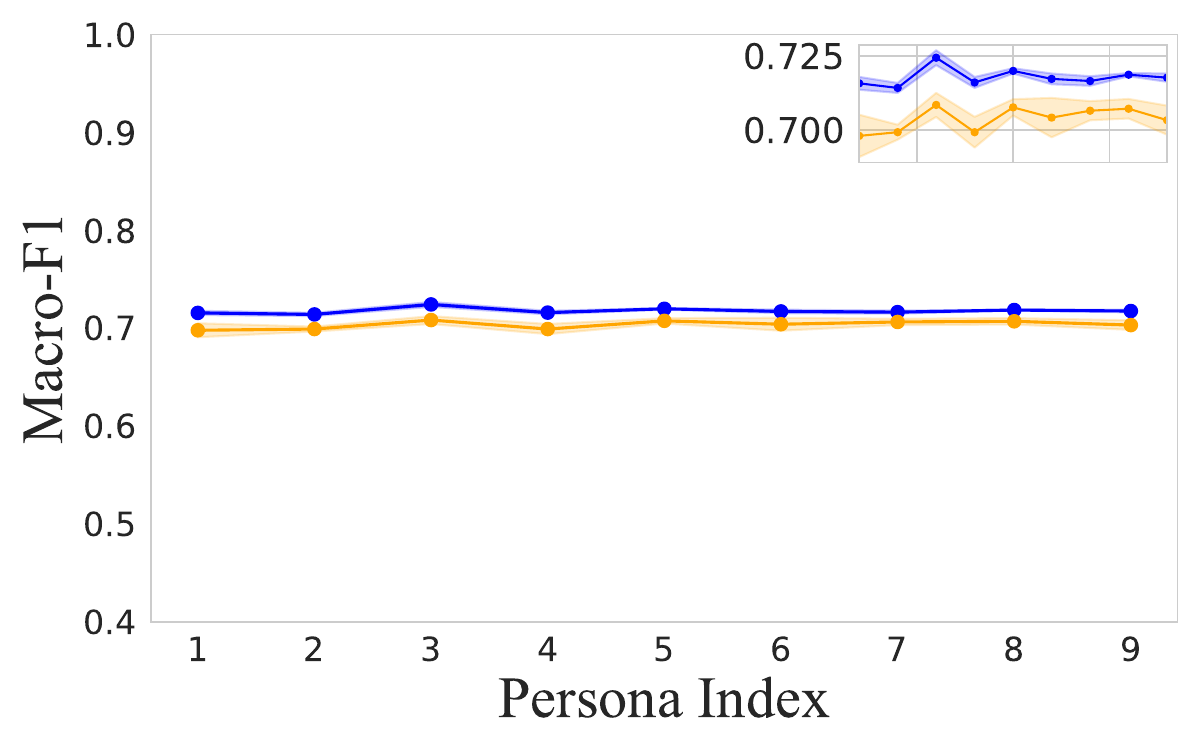}}
  \hspace{-3mm}
  \subfigure[\label{fig:prompt_sens_corr_germ}GermEval Task]{
    \includegraphics[width=0.5\linewidth]{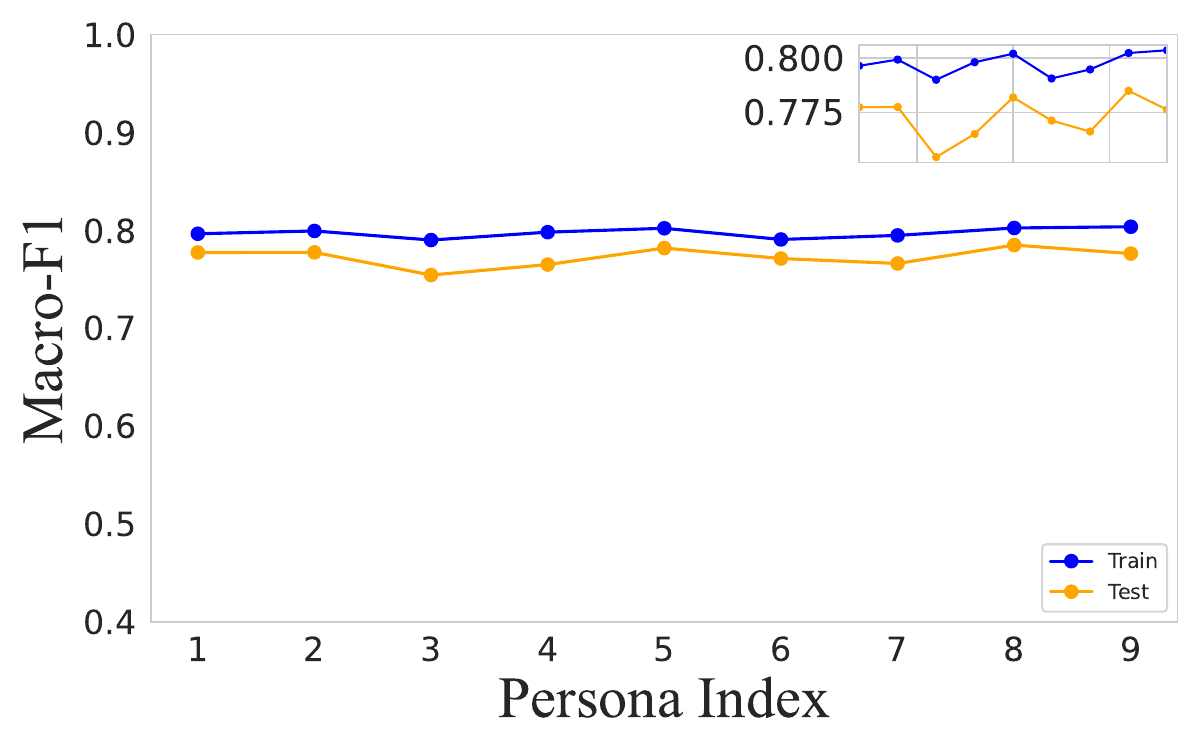}
  }
  \caption{Impact of the nine different persona profiles on model performance (blue train, yellow test performance).}
  \label{fig:f1_prompt_sensibility}
\end{figure}

Figure~\ref{fig:f1_prompt_sensibility} illustrates the performance in terms of macro-F1 when different persona profiles are utilized. Only small variations are observed in performance across the different persona profile definitions. Interestingly, providing no persona at all slightly improves performance for TSNH but decreases performance on GermEval, showing a certain dataset dependency. 

In the performance evaluation presented in \tableautorefname~\ref{tab:class_result}, we utilized zero-shot prompting in our proposed models to evaluate the relevance of adjectives to input texts (concept bottleneck representation). 
In addition, we evaluate their performance when the deployed LLM is prompted using ICL, by providing the definition of the queried adjective as contextual information. We obtain the definition of each adjective by prompting GPT 4o.

Similar to the previous investigation of persona profiles, to compare zero-shot and ICL prompting, we first examine the differences in the values of the concept bottleneck representations (i.e., adjective scores) using Pearson correlation, and then compare the final prediction performances in terms of macro-F1. 
Using the GermEval and TSNH datasets, we observe strong to moderate positive correlations of $0.630$ ($p\ll 0.001$) and 0.587 ($p\ll 0.001$), respectively. This suggests that the two prompting techniques capture a similar underlying semantic structure in the concept bottleneck representations.
In addition, we obtain macro-F1 scores of $0.778$ and $0.702$ for GermEval and TSNH, respectively, which are slightly lower than the scores achieved with zero-shot prompting ($0.781$ and $0.739$).

While the ICL prompting strategy we employ for the \HSCBM~model is relatively simple, we acknowledge that more sophisticated prompting techniques may further enhance the effectiveness of concept evaluation. However, in our experiments, adding more detailed information about the evaluated concepts to the prompts did not yield improvements in macro-F1. Moreover, incorporating more complex ICL prompts significantly slowed down the computation of representations. This is because the number of queries sent to LLMs for each text is equal to the number of concepts, and each of these queries requires processing all the context contained in the prompt.

%
%

\subsection{Automatic concept generation}
\label{subsec:concept_gen}

We explore the feasibility of automatically constructing the adjective lexicon to investigate the scalability and domain transferability of our manually curated concept 
set. To this end, we prompt GPT 4o to 
automatically generate adjectives suitable for distinguishing counter 
and non-counter speech (the two classes of the TSNH dataset). We conduct four rounds of iterative prompt refinements. After the initial set of adjective proposals, we instruct the LLM to augment the list with additional relevant adjectives that may have been missed. 
Across the four iterations, we collect a total of 375 adjectives, which, after removing duplicates, results in 321 unique adjectives.

We assess the semantic overlap of this automatically constructed set with our expert-curated set. 
To enable a more robust semantic comparison, we expand both sets using synonyms from WordNet, allowing us to account for adjectives that are semantically similar but lexically different. This semantic matching reveals that 23.8\% of the expert-curated adjectives are present in the generated set.

To evaluate the practical utility of the automatically generated concepts, 
we train the \HSCBM~model on the Germeval and TSNH datasets, following the same five-fold 
cross-validation strategy. The models trained on the automatically generated concepts achieved a macro-F1 score of 0.778 ($\pm$ 0.001) and 0.720 ($\pm$ 0.007) for Germeval and TSNH, respectively. This represents only 0.3\% lower than the performance obtained with the expert-curated adjective set for Germeval and 1.9\% for TSNH (see \tableautorefname~\ref{tab:class_result}). This small performance drop, despite the limited semantic overlap between the two sets, highlights the robustness of our approach to variation in concept sets. These results suggest that automatically derived concepts, even though they are not fully aligned with those proposed by domain experts, can still lead to effective classification.
Overall, these results demonstrate the potential of our approach to scale effectively and transfer easily to other domains without substantial loss in performance (i.e., by using LLMs to generate concept sets).

\section{Discussion}
Our approach addresses a crucial gap in the literature by providing human-interpretable explanations at both local and global levels. While prior work has primarily focused on post-hoc XAI methods or attention-based approaches that provide local explanations, these have been criticized in the literature for lacking faithfulness \citep{wilming2024gecobenchgendercontrolledtextdataset, slack2020fooling, jain2019attention}. In contrast, our method integrates interpretability directly into the model architecture via concepts, enabling more transparent and faithful reasoning not only about individual predictions (local explanations) but also about the overall model behaviour (global explanations). This interpretability on two levels, i.e., local and global, is essential for fostering trust, accountability, and transparency in NLP models.

Our work contributes to the growing body of research on concept-based interpretability in NLP. In contrast to previous CBM approaches, which rely on abstract concepts that may represent various topics or linguistic patterns, our use of human-understandable adjectives as interpretable units grounds the model's reasoning in concepts that are easier for users to understand and evaluate.

Our approach demonstrates that the CBM architecture can be effectively applied to the domain of hate and counter speech recognition, a task where transparency is particularly important due to its sensitive and subjective nature. Our results show that explanations based on adjectives are not only informative but also align well with human assessment, as confirmed by our user study.

Overall, the main contributions of this work are: i) evidence that the CBM architecture is useful for hate and counter speech classification; ii) empirical proof that adjectives serving as concepts are a highly promising representation for the investigated NLP tasks; iii) validation that the proposed CBM architecture achieves a high level of transparency without any trade-off in performance.

Despite the high performance and human interpretability of our approach, we identify several limitations that will be addressed in future work. First, in this study, we pre-defined the lexicon of descriptive adjectives with the assistance of domain experts (social scientists). Our approach is easily adaptable to new domains, tasks, and languages by simply modifying the lexicon. Future work should explore the potential for automating the creation of this lexicon as well as the extension of the evaluation towards a broader set of languages. Second, we focused on zero-shot prompts to demonstrate the effectiveness of our method. Investigating the use of few-shot or more elaborate prompting strategies may further enhance the performance of adjective evaluation. 

\section{Conclusion}
\label{sec:conclusion}
In this work, we proposed ``Speech Concept Bottleneck Models'' (\HSCBM), a novel representation and classification model based on CBM methodology to effectively address NLP tasks that require understanding abstract and subtle concepts, including hate speech and counter speech recognition. 
\HSCBM~effectively leverages the latest advancements in LLMs by distilling knowledge from these complex models in a deterministic fashion. 
Specifically, we proposed using adjectives as bottleneck concepts and leveraging LLMs to encode input texts into latent abstract embeddings. 
Furthermore, instead of leveraging 
a simple MLP as a classifier on top of the adjective-based bottleneck representations, we introduce a sigmoid-activated dense layer that functions as a relevance gate. To further enhance the interpretability of our \HSCBM~model, we propose a novel regularization term that guides the model to focus on the most discriminative adjectives for the target classes. 

To evaluate the performance and interpretability of our proposed method and its variants, we conducted experiments on five publicly available benchmark datasets focusing on different NLP tasks and covering two languages. 
Our empirical experiments demonstrate that recognising hate and counter speech does not require raw input features; instead, our proposed adjective-based representation is sufficient to achieve and even surpass state-of-the-art performance. 
Furthermore, we empirically demonstrated that existing methods, such as transformers, can also benefit from our adjective-based bottleneck representations, as they provide complementary information. The combination of our adjective-based representation with transformer embeddings, as proposed in \HSCBMT, often exceeds the performance of fine-tuned transformers and even outperforms large transformer models on certain datasets.
These findings highlight that our approach holds strong potential for broader applications, as the adjective lexicon can be adapted to different domains, tasks, and languages.


\bibliographystyle{elsarticle-harv} 
\bibliography{bibs_IPM}

\begin{thebibliography}{71}
\expandafter\ifx\csname natexlab\endcsname\relax\def\natexlab#1{#1}\fi
\providecommand{\url}[1]{\texttt{#1}}
\providecommand{\href}[2]{#2}
\providecommand{\path}[1]{#1}
\providecommand{\DOIprefix}{doi:}
\providecommand{\ArXivprefix}{arXiv:}
\providecommand{\URLprefix}{URL: }
\providecommand{\Pubmedprefix}{pmid:}
\providecommand{\doi}[1]{\href{http://dx.doi.org/#1}{\path{#1}}}
\providecommand{\Pubmed}[1]{\href{pmid:#1}{\path{#1}}}
\providecommand{\bibinfo}[2]{#2}
\ifx\xfnm\relax \def\xfnm[#1]{\unskip,\space#1}\fi
\bibitem[{Antverg and Belinkov(2022)}]{AntvergB22}
\bibinfo{author}{Antverg, O.}, \bibinfo{author}{Belinkov, Y.}, \bibinfo{year}{2022}.
\newblock \bibinfo{title}{{On the Pitfalls of Analyzing Individual Neurons in Language Models}}, in: \bibinfo{booktitle}{Proc.\ 10th International Conference on Learning Representations ({ICLR})}, \bibinfo{publisher}{OpenReview.net}.
\bibitem[{Arya et~al.(2019)Arya, Bellamy, Chen et~al.}]{arya2019one}
\bibinfo{author}{Arya, V.}, \bibinfo{author}{Bellamy, R.K.}, \bibinfo{author}{Chen, P.Y.}, et~al., \bibinfo{year}{2019}.
\newblock \bibinfo{title}{{One explanation does not fit all: A toolkit and taxonomy of ai explainability techniques}}.
\newblock \bibinfo{journal}{arXiv preprint arXiv:1909.03012} .
\bibitem[{Bastings et~al.(2019)Bastings, Aziz and Titov}]{BastingsAT19}
\bibinfo{author}{Bastings, J.}, \bibinfo{author}{Aziz, W.}, \bibinfo{author}{Titov, I.}, \bibinfo{year}{2019}.
\newblock \bibinfo{title}{{Interpretable Neural Predictions with Differentiable Binary Variables}}, in: \bibinfo{booktitle}{Proc.\ 57th Annual Meeting of the Association for Computational Linguistics ({ACL})}, \bibinfo{publisher}{Association for Computational Linguistics}. pp. \bibinfo{pages}{2963--2977}.
\newblock \DOIprefix\doi{10.18653/v1/P19-1284}.
\bibitem[{Benesch et~al.(2016)Benesch, Ruths, Dillon, Saleem and Wright}]{benesch2016counterspeech}
\bibinfo{author}{Benesch, S.}, \bibinfo{author}{Ruths, D.}, \bibinfo{author}{Dillon, K.P.}, \bibinfo{author}{Saleem, H.M.}, \bibinfo{author}{Wright, L.}, \bibinfo{year}{2016}.
\newblock \bibinfo{title}{Counterspeech on twitter: A field study}.
\newblock \bibinfo{journal}{A report for public safety Canada under the Kanishka project} , \bibinfo{pages}{1--39}.
\bibitem[{Bibal et~al.(2022)Bibal, Cardon, Alfter et~al.}]{bibal2022attention}
\bibinfo{author}{Bibal, A.}, \bibinfo{author}{Cardon, R.}, \bibinfo{author}{Alfter, D.}, et~al., \bibinfo{year}{2022}.
\newblock \bibinfo{title}{{Is attention explanation? an introduction to the debate}}, in: \bibinfo{booktitle}{Proceedings of the 60th Annual Meeting of the Association for Computational Linguistics (Volume 1: Long Papers)}, pp. \bibinfo{pages}{3889--3900}.
\bibitem[{Bilewicz and Soral(2020)}]{BS20}
\bibinfo{author}{Bilewicz, M.}, \bibinfo{author}{Soral, W.}, \bibinfo{year}{2020}.
\newblock \bibinfo{title}{{Hate Speech Epidemic. The Dynamic Effects of Derogatory Language on Intergroup Relations and Political Radicalization}}.
\newblock \bibinfo{journal}{Political Psychology} \bibinfo{volume}{41}, \bibinfo{pages}{3--33}.
\newblock \DOIprefix\doi{https://doi.org/10.1111/pops.12670}.
\bibitem[{B{\"o}ck et~al.(2024)B{\"o}ck, Slijep{\v{c}}evi{\'c} and Zeppelzauer}]{bock2024exploring}
\bibinfo{author}{B{\"o}ck, A.J.}, \bibinfo{author}{Slijep{\v{c}}evi{\'c}, D.}, \bibinfo{author}{Zeppelzauer, M.}, \bibinfo{year}{2024}.
\newblock \bibinfo{title}{{Exploring the Plausibility of Hate and Counter Speech Detectors With Explainable Ai}}, in: \bibinfo{booktitle}{2024 International Conference on Content-Based Multimedia Indexing (CBMI)}, \bibinfo{organization}{IEEE}. pp. \bibinfo{pages}{1--8}.
\bibitem[{Bornheim et~al.(2021)Bornheim, Grieger and Bialonski}]{bornheim-etal-2021-fhac}
\bibinfo{author}{Bornheim, T.}, \bibinfo{author}{Grieger, N.}, \bibinfo{author}{Bialonski, S.}, \bibinfo{year}{2021}.
\newblock \bibinfo{title}{{{FHAC} at {G}erm{E}val 2021: Identifying {G}erman toxic, engaging, and fact-claiming comments with ensemble learning}}, in: \bibinfo{editor}{Risch, J.}, \bibinfo{editor}{Stoll, A.}, \bibinfo{editor}{Wilms, L.}, \bibinfo{editor}{Wiegand, M.} (Eds.), \bibinfo{booktitle}{Proceedings of the GermEval 2021 Shared Task on the Identification of Toxic, Engaging, and Fact-Claiming Comments}, \bibinfo{publisher}{Association for Computational Linguistics}. pp. \bibinfo{pages}{105--111}.
\newblock \URLprefix \url{https://aclanthology.org/2021.germeval-1.16/}.
\bibitem[{Breiman(2001)}]{breiman2001random}
\bibinfo{author}{Breiman, L.}, \bibinfo{year}{2001}.
\newblock \bibinfo{title}{{Random forests}}.
\newblock \bibinfo{journal}{Machine learning} \bibinfo{volume}{45}, \bibinfo{pages}{5--32}.
\bibitem[{Chan et~al.(2025)Chan, Chen, Cheng and Huang}]{chan2024don}
\bibinfo{author}{Chan, B.J.}, \bibinfo{author}{Chen, C.T.}, \bibinfo{author}{Cheng, J.H.}, \bibinfo{author}{Huang, H.H.}, \bibinfo{year}{2025}.
\newblock \bibinfo{title}{Don't do rag: When cache-augmented generation is all you need for knowledge tasks}, in: \bibinfo{booktitle}{Companion Proceedings of the ACM on Web Conference 2025}, \bibinfo{publisher}{Association for Computing Machinery}. p. \bibinfo{pages}{893–897}.
\newblock \URLprefix \url{https://doi.org/10.1145/3701716.3715490}, \DOIprefix\doi{10.1145/3701716.3715490}.
\bibitem[{Chung et~al.(2021)Chung, Guerini and Agerri}]{chung-etal-2021-multilingual}
\bibinfo{author}{Chung, Y.L.}, \bibinfo{author}{Guerini, M.}, \bibinfo{author}{Agerri, R.}, \bibinfo{year}{2021}.
\newblock \bibinfo{title}{{Multilingual Counter Narrative Type Classification}}, in: \bibinfo{booktitle}{Proc.\ 8th Workshop on Argument Mining}, \bibinfo{publisher}{Association for Computational Linguistics}. pp. \bibinfo{pages}{125--132}.
\newblock \DOIprefix\doi{10.18653/v1/2021.argmining-1.12}.
\bibitem[{Chung et~al.(2019)Chung, Kuzmenko, Tekiroglu et~al.}]{Chung_2019}
\bibinfo{author}{Chung, Y.L.}, \bibinfo{author}{Kuzmenko, E.}, \bibinfo{author}{Tekiroglu, S.S.}, et~al., \bibinfo{year}{2019}.
\newblock \bibinfo{title}{{{CONAN}-{CO}unter {NA}rratives through Nichesourcing: a Multilingual Dataset of Responses to Fight Online Hate Speech}}, in: \bibinfo{booktitle}{Proc.\ 57th Annual Meeting of the Association for Computational Linguistics (ACL)}, \bibinfo{publisher}{Association for Computational Linguistics}. pp. \bibinfo{pages}{2819--2829}.
\newblock \DOIprefix\doi{10.18653/v1/P19-1271}.
\bibitem[{Cobbe(2021)}]{cobbe_algorithmic_2021}
\bibinfo{author}{Cobbe, J.}, \bibinfo{year}{2021}.
\newblock \bibinfo{title}{{Algorithmic Censorship by Social Platforms: Power and Resistance}}.
\newblock \bibinfo{journal}{Philosophy \& Technology} \bibinfo{volume}{34}, \bibinfo{pages}{739--766}.
\newblock \DOIprefix\doi{10.1007/s13347-020-00429-0}.
\bibitem[{Conneau et~al.(2020)Conneau, Khandelwal, Goyal et~al.}]{Conneau2019UnsupervisedCR}
\bibinfo{author}{Conneau, A.}, \bibinfo{author}{Khandelwal, K.}, \bibinfo{author}{Goyal, N.}, et~al., \bibinfo{year}{2020}.
\newblock \bibinfo{title}{{Unsupervised Cross-lingual Representation Learning at Scale}}, in: \bibinfo{booktitle}{Proc.\ 58th Annual Meeting of the Association for Computational Linguistics (ACL)}, \bibinfo{publisher}{Association for Computational Linguistics}. pp. \bibinfo{pages}{8440--8451}.
\newblock \DOIprefix\doi{10.18653/v1/2020.acl-main.747}.
\bibitem[{Dalvi et~al.(2019)Dalvi, Durrani, Sajjad et~al.}]{DalviDSBBG19}
\bibinfo{author}{Dalvi, F.}, \bibinfo{author}{Durrani, N.}, \bibinfo{author}{Sajjad, H.}, et~al., \bibinfo{year}{2019}.
\newblock \bibinfo{title}{{What Is One Grain of Sand in the Desert? Analyzing Individual Neurons in Deep {NLP} Models}}, in: \bibinfo{booktitle}{Proc.\ 33rd {AAAI} Conference on Artificial Intelligence ({AAAI})}, \bibinfo{publisher}{{AAAI} Press}. pp. \bibinfo{pages}{6309--6317}.
\newblock \DOIprefix\doi{10.1609/AAAI.V33I01.33016309}.
\bibitem[{Das et~al.(2022)Das, Gupta, Kovatchev et~al.}]{DasGKLL22}
\bibinfo{author}{Das, A.}, \bibinfo{author}{Gupta, C.}, \bibinfo{author}{Kovatchev, V.}, et~al., \bibinfo{year}{2022}.
\newblock \bibinfo{title}{{{P}roto{TE}x: Explaining Model Decisions with Prototype Tensors}}, in: \bibinfo{booktitle}{Proc.\ 60th Annual Meeting of the Association for Computational Linguistics (ACL)}, \bibinfo{publisher}{Association for Computational Linguistics}. pp. \bibinfo{pages}{2986--2997}.
\newblock \DOIprefix\doi{10.18653/v1/2022.acl-long.213}.
\bibitem[{{De la Peña Sarracén} and Rosso(2023)}]{DELAPENASARRACEN2023103433}
\bibinfo{author}{{De la Peña Sarracén}, G.L.}, \bibinfo{author}{Rosso, P.}, \bibinfo{year}{2023}.
\newblock \bibinfo{title}{{Systematic keyword and bias analyses in hate speech detection}}.
\newblock \bibinfo{journal}{Information Processing \& Management} \bibinfo{volume}{60}, \bibinfo{pages}{103433}.
\newblock \DOIprefix\doi{https://doi.org/10.1016/j.ipm.2023.103433}.
\bibitem[{Devlin et~al.(2019)Devlin, Chang, Lee et~al.}]{devlin-etal-2019-bert}
\bibinfo{author}{Devlin, J.}, \bibinfo{author}{Chang, M.W.}, \bibinfo{author}{Lee, K.}, et~al., \bibinfo{year}{2019}.
\newblock \bibinfo{title}{{{BERT}: Pre-training of Deep Bidirectional Transformers for Language Understanding}}, in: \bibinfo{booktitle}{Proc.\ 2019 Conference of the North {A}merican Chapter of the Association for Computational Linguistics: Human Language Technologies (NAACL-HCT)}, \bibinfo{publisher}{Association for Computational Linguistics}. pp. \bibinfo{pages}{4171--4186}.
\newblock \DOIprefix\doi{10.18653/v1/N19-1423}.
\bibitem[{Dimitrov et~al.(2024)Dimitrov, Alam, Hasanain et~al.}]{dimitrov2024semeval}
\bibinfo{author}{Dimitrov, D.}, \bibinfo{author}{Alam, F.}, \bibinfo{author}{Hasanain, M.}, et~al., \bibinfo{year}{2024}.
\newblock \bibinfo{title}{{Semeval-2024 task 4: Multilingual detection of persuasion techniques in memes}}, in: \bibinfo{booktitle}{Proceedings of the 18th International Workshop on Semantic Evaluation (SemEval-2024)}, pp. \bibinfo{pages}{2009--2026}.
\bibitem[{Dubey et~al.(2024)Dubey, Jauhri, Pandey et~al.}]{dubey2024llama}
\bibinfo{author}{Dubey, A.}, \bibinfo{author}{Jauhri, A.}, \bibinfo{author}{Pandey, A.}, et~al., \bibinfo{year}{2024}.
\newblock \bibinfo{title}{{The llama 3 herd of models}}.
\newblock \bibinfo{journal}{arXiv preprint arXiv:2407.21783} .
\bibitem[{Durrani et~al.(2020)Durrani, Sajjad, Dalvi et~al.}]{DurraniSDB20}
\bibinfo{author}{Durrani, N.}, \bibinfo{author}{Sajjad, H.}, \bibinfo{author}{Dalvi, F.}, et~al., \bibinfo{year}{2020}.
\newblock \bibinfo{title}{{Analyzing Individual Neurons in Pre-trained Language Models}}, in: \bibinfo{booktitle}{Proc.\ 2020 Conference on Empirical Methods in Natural Language Processing ({EMNLP})}, \bibinfo{publisher}{Association for Computational Linguistics}. pp. \bibinfo{pages}{4865--4880}.
\newblock \DOIprefix\doi{10.18653/v1/2020.emnlp-main.395}.
\bibitem[{Garland et~al.(2020)Garland, Ghazi-Zahedi, Young et~al.}]{garland-etal-2020-countering}
\bibinfo{author}{Garland, J.}, \bibinfo{author}{Ghazi-Zahedi, K.}, \bibinfo{author}{Young, J.G.}, et~al., \bibinfo{year}{2020}.
\newblock \bibinfo{title}{{Countering hate on social media: Large scale classification of hate and counter speech}}, in: \bibinfo{booktitle}{Proceedings of the Fourth Workshop on Online Abuse and Harms}, \bibinfo{publisher}{Association for Computational Linguistics}, \bibinfo{address}{Online}. pp. \bibinfo{pages}{102--112}.
\newblock \DOIprefix\doi{10.18653/v1/2020.alw-1.13}.
\bibitem[{Giorgi et~al.(2025)Giorgi, Cima, Fagni, Avvenuti and Cresci}]{giorgi2024human}
\bibinfo{author}{Giorgi, T.}, \bibinfo{author}{Cima, L.}, \bibinfo{author}{Fagni, T.}, \bibinfo{author}{Avvenuti, M.}, \bibinfo{author}{Cresci, S.}, \bibinfo{year}{2025}.
\newblock \bibinfo{title}{Human and llm biases in hate speech annotations: A socio-demographic analysis of annotators and targets}, in: \bibinfo{booktitle}{Proceedings of the International AAAI Conference on Web and Social Media}, pp. \bibinfo{pages}{653--670}.
\bibitem[{Guillaume et~al.(2022)Guillaume, Duch{\^e}ne and Dehak}]{guillaume2022hate}
\bibinfo{author}{Guillaume, P.}, \bibinfo{author}{Duch{\^e}ne, C.}, \bibinfo{author}{Dehak, R.}, \bibinfo{year}{2022}.
\newblock \bibinfo{title}{{Hate speech and toxic comment detection using transformers}}.
\newblock \bibinfo{journal}{EPITA Speech and Language Recognition Group (ESLR)} .
\bibitem[{Guo et~al.(2023)Guo, Hu, Mu et~al.}]{GuoHMS2024}
\bibinfo{author}{Guo, K.}, \bibinfo{author}{Hu, A.}, \bibinfo{author}{Mu, J.}, et~al., \bibinfo{year}{2023}.
\newblock \bibinfo{title}{{An Investigation of Large Language Models for Real-World Hate Speech Detection}}, in: \bibinfo{booktitle}{Proc.\ 2023 International Conference on Machine Learning and Applications ({ICMLA})}, \bibinfo{publisher}{{IEEE}}. pp. \bibinfo{pages}{1568--1573}.
\newblock \DOIprefix\doi{10.1109/ICMLA58977.2023.00237}.
\bibitem[{Han et~al.(2020)Han, Wallace and Tsvetkov}]{HanWT20}
\bibinfo{author}{Han, X.}, \bibinfo{author}{Wallace, B.C.}, \bibinfo{author}{Tsvetkov, Y.}, \bibinfo{year}{2020}.
\newblock \bibinfo{title}{{Explaining Black Box Predictions and Unveiling Data Artifacts through Influence Functions}}, in: \bibinfo{booktitle}{Proc.\ 58th Annual Meeting of the Association for Computational Linguistics ({ACL})}, \bibinfo{publisher}{Association for Computational Linguistics}. pp. \bibinfo{pages}{5553--5563}.
\newblock \DOIprefix\doi{10.18653/v1/2020.acl-main.492}.
\bibitem[{Hao et~al.(2024)Hao, Wu, Pan et~al.}]{HaoWP24}
\bibinfo{author}{Hao, G.}, \bibinfo{author}{Wu, J.}, \bibinfo{author}{Pan, Q.}, et~al., \bibinfo{year}{2024}.
\newblock \bibinfo{title}{{Quantifying the uncertainty of LLM hallucination spreading in complex adaptive social networks}}.
\newblock \bibinfo{journal}{Scientific Reports} \bibinfo{volume}{14}.
\newblock \DOIprefix\doi{10.1038/s41598-024066708-4}.
\bibitem[{Hengle et~al.(2024)Hengle, Padhi, Singh et~al.}]{hengle-etal-2024-intent}
\bibinfo{author}{Hengle, A.}, \bibinfo{author}{Padhi, A.}, \bibinfo{author}{Singh, S.}, et~al., \bibinfo{year}{2024}.
\newblock \bibinfo{title}{{Intent-conditioned and Non-toxic Counterspeech Generation using Multi-Task Instruction Tuning with {RLAIF}}}, in: \bibinfo{booktitle}{Proceedings of the 2024 Conference of the North American Chapter of the Association for Computational Linguistics: Human Language Technologies (Volume 1: Long Papers)}, \bibinfo{publisher}{Association for Computational Linguistics}. pp. \bibinfo{pages}{6716--6733}.
\newblock \DOIprefix\doi{10.18653/v1/2024.naacl-long.374}.
\bibitem[{Hinton(2012)}]{hinton2012rmsprop}
\bibinfo{author}{Hinton, G.E.}, \bibinfo{year}{2012}.
\newblock \bibinfo{title}{{RMSprop: Divide the gradient by a running average of recent gradient magnitudes}}.
\newblock \bibinfo{howpublished}{\url{https://www.cs.toronto.edu/~hinton/absps/rmsprop.pdf}}.
\newblock \bibinfo{note}{Accessed: 2025-03-20}.
\bibitem[{Jahan and Oussalah(2023)}]{JahanO23}
\bibinfo{author}{Jahan, M.S.}, \bibinfo{author}{Oussalah, M.}, \bibinfo{year}{2023}.
\newblock \bibinfo{title}{{A systematic review of hate speech automatic detection using natural language processing}}.
\newblock \bibinfo{journal}{Neurocomputing} \bibinfo{volume}{546}, \bibinfo{pages}{126232}.
\newblock \DOIprefix\doi{https://doi.org/10.1016/j.neucom.2023.126232}.
\bibitem[{Jin et~al.(2024)Jin, Wanner and Shvets}]{jin-etal-2024-gpt}
\bibinfo{author}{Jin, Y.}, \bibinfo{author}{Wanner, L.}, \bibinfo{author}{Shvets, A.}, \bibinfo{year}{2024}.
\newblock \bibinfo{title}{{{GPT}-{H}ate{C}heck: Can {LLM}s Write Better Functional Tests for Hate Speech Detection?}}, in: \bibinfo{booktitle}{Proc.\ 2024 Joint International Conference on Computational Linguistics, Language Resources and Evaluation (LREC-COLING)}, \bibinfo{publisher}{ELRA and ICCL}. pp. \bibinfo{pages}{7867--7885}.
\bibitem[{Kalyan(2024)}]{kalyan2023survey}
\bibinfo{author}{Kalyan, K.S.}, \bibinfo{year}{2024}.
\newblock \bibinfo{title}{{A survey of {GPT-3} family large language models including {ChatGPT} and {GPT-4}}}.
\newblock \bibinfo{journal}{Natural Language Processing Journal} \bibinfo{volume}{6}, \bibinfo{pages}{100048}.
\newblock \DOIprefix\doi{https://doi.org/10.1016/j.nlp.2023.100048}.
\bibitem[{Kirk et~al.(2023)Kirk, Yin, Vidgen and R{\"o}ttger}]{kirk-etal-2023-semeval}
\bibinfo{author}{Kirk, H.}, \bibinfo{author}{Yin, W.}, \bibinfo{author}{Vidgen, B.}, \bibinfo{author}{R{\"o}ttger, P.}, \bibinfo{year}{2023}.
\newblock \bibinfo{title}{{{S}em{E}val-2023 Task 10: Explainable Detection of Online Sexism}}, in: \bibinfo{editor}{Ojha, A.K.}, \bibinfo{editor}{Do{\u{g}}ru{\"o}z, A.S.}, \bibinfo{editor}{Da~San~Martino, G.}, \bibinfo{editor}{Tayyar~Madabushi, H.}, \bibinfo{editor}{Kumar, R.}, \bibinfo{editor}{Sartori, E.} (Eds.), \bibinfo{booktitle}{Proceedings of the 17th International Workshop on Semantic Evaluation (SemEval-2023)}, \bibinfo{publisher}{Association for Computational Linguistics}, \bibinfo{address}{Toronto, Canada}. pp. \bibinfo{pages}{2193--2210}.
\newblock \DOIprefix\doi{10.18653/v1/2023.semeval-1.305}.
\bibitem[{Koh et~al.(2020)Koh, Nguyen, Tang et~al.}]{KohNTMPKL20}
\bibinfo{author}{Koh, P.W.}, \bibinfo{author}{Nguyen, T.}, \bibinfo{author}{Tang, Y.S.}, et~al., \bibinfo{year}{2020}.
\newblock \bibinfo{title}{{Concept Bottleneck Models}}, in: \bibinfo{booktitle}{Proc.\ 37th International Conference on Machine Learning ({ICML})}, \bibinfo{publisher}{{PMLR}}. pp. \bibinfo{pages}{5338--5348}.
\bibitem[{Kong et~al.(2024)Kong, Zhao, Chen et~al.}]{KongSH2024}
\bibinfo{author}{Kong, A.}, \bibinfo{author}{Zhao, S.}, \bibinfo{author}{Chen, H.}, et~al., \bibinfo{year}{2024}.
\newblock \bibinfo{title}{{Better Zero-Shot Reasoning with Role-Play Prompting}}, in: \bibinfo{booktitle}{Proc.\ 2024 Conference of the North American Chapter of the Association for Computational Linguistics: Human Language Technologies}, \bibinfo{publisher}{Association for Computational Linguistics}. pp. \bibinfo{pages}{4099--4113}.
\newblock \DOIprefix\doi{10.18653/v1/2024.naacl-long.228}.
\bibitem[{Labadie~Tamayo et~al.(2023)Labadie~Tamayo, Chulvi-Ferriols and Rosso}]{labadie2023everybody}
\bibinfo{author}{Labadie~Tamayo, R.}, \bibinfo{author}{Chulvi-Ferriols, M.A.}, \bibinfo{author}{Rosso, P.}, \bibinfo{year}{2023}.
\newblock \bibinfo{title}{{Everybody hurts, sometimes overview of hurtful humour at iberlef 2023: Detection of humour spreading prejudice in {T}witter}}.
\newblock \bibinfo{journal}{Procesamiento del lenguaje natural} , \bibinfo{pages}{383--395}.
\bibitem[{Lee et~al.(2022)Lee, Na, Song et~al.}]{LeeNSSP22}
\bibinfo{author}{Lee, H.}, \bibinfo{author}{Na, Y.J.}, \bibinfo{author}{Song, H.}, et~al., \bibinfo{year}{2022}.
\newblock \bibinfo{title}{{{ELF}22: A Context-based Counter Trolling Dataset to Combat {I}nternet Trolls}}, in: \bibinfo{booktitle}{Proc.\ 13th Language Resources and Evaluation Conference}, \bibinfo{publisher}{European Language Resources Association}. pp. \bibinfo{pages}{3530--3541}.
\bibitem[{Li et~al.(2024a)Li, Fan, Atreja et~al.}]{LiFAH24}
\bibinfo{author}{Li, L.}, \bibinfo{author}{Fan, L.}, \bibinfo{author}{Atreja, S.}, et~al., \bibinfo{year}{2024}a.
\newblock \bibinfo{title}{{``HOT" {ChatGPT}: The Promise of {ChatGPT} in Detecting and Discriminating Hateful, Offensive, and Toxic Comments on Social Media}}.
\newblock \bibinfo{journal}{ACM Trans. Web} \bibinfo{volume}{18}.
\newblock \DOIprefix\doi{10.1145/3643829}.
\bibitem[{Li et~al.(2024b)Li, Cao, Kang et~al.}]{LiCKYCJT23}
\bibinfo{author}{Li, Q.}, \bibinfo{author}{Cao, Y.}, \bibinfo{author}{Kang, J.}, et~al., \bibinfo{year}{2024}b.
\newblock \bibinfo{title}{{{LaFFi}: Leveraging Hybrid Natural Language Feedback for Fine-tuning Language Models}}.
\newblock \bibinfo{journal}{CoRR} \bibinfo{volume}{abs/2401.00907}.
\newblock \DOIprefix\doi{10.48550/ARXIV.2401.00907}, \href{http://arxiv.org/abs/2401.00907}{{\tt arXiv:2401.00907}}.
\bibitem[{Liu et~al.(2019)Liu, Ott, Goyal et~al.}]{liu2019robertarobustlyoptimizedbert}
\bibinfo{author}{Liu, Y.}, \bibinfo{author}{Ott, M.}, \bibinfo{author}{Goyal, N.}, et~al., \bibinfo{year}{2019}.
\newblock \bibinfo{title}{{{RoBERTa}: A Robustly Optimized {BERT} Pretraining Approach}}.
\newblock \bibinfo{journal}{CoRR} \bibinfo{volume}{abs/1907.11692}.
\newblock \href{http://arxiv.org/abs/1907.11692}{{\tt arXiv:1907.11692}}.
\bibitem[{Ludan et~al.(2024)Ludan, Lyu, Yang et~al.}]{LudanLYD24}
\bibinfo{author}{Ludan, J.M.}, \bibinfo{author}{Lyu, Q.}, \bibinfo{author}{Yang, Y.}, et~al., \bibinfo{year}{2024}.
\newblock \bibinfo{title}{{Interpretable-by-Design Text Understanding with Iteratively Generated Concept Bottleneck}}.
\newblock \href{http://arxiv.org/abs/2310.19660}{{\tt arXiv:2310.19660}}.
\bibitem[{Madsen et~al.(2022)Madsen, Reddy and Chandar}]{madsen2022post}
\bibinfo{author}{Madsen, A.}, \bibinfo{author}{Reddy, S.}, \bibinfo{author}{Chandar, S.}, \bibinfo{year}{2022}.
\newblock \bibinfo{title}{{Post-hoc interpretability for neural {NLP}: A survey}}.
\newblock \bibinfo{journal}{ACM Computing Surveys} \bibinfo{volume}{55}, \bibinfo{pages}{1--42}.
\bibitem[{Masud et~al.(2022)Masud, Bedi, Khan et~al.}]{MasudBKA22}
\bibinfo{author}{Masud, S.}, \bibinfo{author}{Bedi, M.}, \bibinfo{author}{Khan, M.A.}, et~al., \bibinfo{year}{2022}.
\newblock \bibinfo{title}{{Proactively Reducing the Hate Intensity of Online Posts via Hate Speech Normalization}}, in: \bibinfo{booktitle}{Proc.\ 28th {ACM} {SIGKDD} Conference on Knowledge Discovery and Data Mining (KDD)}, \bibinfo{publisher}{{ACM}}. pp. \bibinfo{pages}{3524--3534}.
\newblock \DOIprefix\doi{10.1145/3534678.3539161}.
\bibitem[{Masud et~al.(2021)Masud, Dutta, Makkar et~al.}]{MasudDMJGD021}
\bibinfo{author}{Masud, S.}, \bibinfo{author}{Dutta, S.}, \bibinfo{author}{Makkar, S.}, et~al., \bibinfo{year}{2021}.
\newblock \bibinfo{title}{{Hate is the New Infodemic: {A} Topic-aware Modeling of Hate Speech Diffusion on {Twitter}}}, in: \bibinfo{booktitle}{Proc.\ 37th {IEEE} International Conference on Data Engineering ({ICDE})}, \bibinfo{publisher}{{IEEE}}. pp. \bibinfo{pages}{504--515}.
\newblock \DOIprefix\doi{10.1109/ICDE51399.2021.00050}.
\bibitem[{Mathew et~al.(2019a)Mathew, Dutt, Goyal et~al.}]{MathewDGM19}
\bibinfo{author}{Mathew, B.}, \bibinfo{author}{Dutt, R.}, \bibinfo{author}{Goyal, P.}, et~al., \bibinfo{year}{2019}a.
\newblock \bibinfo{title}{{Spread of Hate Speech in Online Social Media}}, in: \bibinfo{booktitle}{Proc.\ 10th ACM Conference on Web Science ({WebSci})}, \bibinfo{publisher}{Association for Computing Machinery}. pp. \bibinfo{pages}{173--182}.
\newblock \DOIprefix\doi{10.1145/3292522.3326034}.
\bibitem[{Mathew et~al.(2019b)Mathew, Saha, Tharad et~al.}]{mathew2018thou}
\bibinfo{author}{Mathew, B.}, \bibinfo{author}{Saha, P.}, \bibinfo{author}{Tharad, H.}, et~al., \bibinfo{year}{2019}b.
\newblock \bibinfo{title}{{Thou Shalt Not Hate: Countering Online Hate Speech}}, in: \bibinfo{booktitle}{Proc. \ 13th International Conference on Web and Social Media ({ICWSM})}, \bibinfo{publisher}{{AAAI} Press}. pp. \bibinfo{pages}{369--380}.
\bibitem[{Miller(2019)}]{miller2019explanation}
\bibinfo{author}{Miller, T.}, \bibinfo{year}{2019}.
\newblock \bibinfo{title}{{Explanation in artificial intelligence: Insights from the social sciences}}.
\newblock \bibinfo{journal}{Artificial intelligence} \bibinfo{volume}{267}, \bibinfo{pages}{1--38}.
\bibitem[{Molnar(2022)}]{molnar2022interpretable}
\bibinfo{author}{Molnar, C.}, \bibinfo{year}{2022}.
\newblock \bibinfo{title}{Interpretable Machine Learning: A Guide for Making Black Box Models Explainable}.
\newblock \bibinfo{publisher}{Christoph Molnar}.
\newblock \URLprefix \url{https://books.google.at/books?id=hVv_zgEACAAJ}.
\bibitem[{Oikarinen et~al.(2023)Oikarinen, Das, Nguyen et~al.}]{OikarinenDNW23}
\bibinfo{author}{Oikarinen, T.P.}, \bibinfo{author}{Das, S.}, \bibinfo{author}{Nguyen, L.M.}, et~al., \bibinfo{year}{2023}.
\newblock \bibinfo{title}{{Label-free Concept Bottleneck Models}}, in: \bibinfo{booktitle}{Proc.\ 11th International Conference on Learning Representations ({ICLR})}, \bibinfo{publisher}{OpenReview.net}.
\bibitem[{Ousidhoum et~al.(2019)Ousidhoum, Lin, Zhang et~al.}]{OusidZSY19}
\bibinfo{author}{Ousidhoum, N.}, \bibinfo{author}{Lin, Z.}, \bibinfo{author}{Zhang, H.}, et~al., \bibinfo{year}{2019}.
\newblock \bibinfo{title}{{Multilingual and Multi-Aspect Hate Speech Analysis}}, in: \bibinfo{booktitle}{Proc.\ 2019 Conference on Empirical Methods in Natural Language Processing and 9th International Joint Conference on Natural Language Processing ({EMNLP-IJCNLP})}, \bibinfo{publisher}{Association for Computational Linguistics}. pp. \bibinfo{pages}{4674--4683}.
\newblock \DOIprefix\doi{10.18653/V1/D19-1474}.
\bibitem[{Pawar et~al.(2018)Pawar, Agrawal, Joshi et~al.}]{PawarAJGR18}
\bibinfo{author}{Pawar, R.}, \bibinfo{author}{Agrawal, Y.}, \bibinfo{author}{Joshi, A.}, et~al., \bibinfo{year}{2018}.
\newblock \bibinfo{title}{{Cyberbullying Detection System with Multiple Server Configurations}}, in: \bibinfo{booktitle}{Proc.\ 2018 {IEEE} International Conference on Electro/Information Technology ({EIT})}, \bibinfo{publisher}{{IEEE}}. pp. \bibinfo{pages}{90--95}.
\newblock \DOIprefix\doi{10.1109/EIT.2018.8500110}.
\bibitem[{Pen et~al.(2024)Pen, Teo and Wang}]{10605280}
\bibinfo{author}{Pen, H.}, \bibinfo{author}{Teo, N.}, \bibinfo{author}{Wang, Z.}, \bibinfo{year}{2024}.
\newblock \bibinfo{title}{Comparative analysis of hate speech detection: Traditional vs. deep learning approaches}, in: \bibinfo{booktitle}{2024 IEEE Conference on Artificial Intelligence (CAI)}, pp. \bibinfo{pages}{332--337}.
\newblock \DOIprefix\doi{10.1109/CAI59869.2024.00070}.
\bibitem[{Poudhar et~al.(2024)Poudhar, Konstas and Abercrombie}]{poudhar-etal-2024-strategy}
\bibinfo{author}{Poudhar, A.}, \bibinfo{author}{Konstas, I.}, \bibinfo{author}{Abercrombie, G.}, \bibinfo{year}{2024}.
\newblock \bibinfo{title}{{A Strategy Labelled Dataset of Counterspeech}}, in: \bibinfo{editor}{Chung, Y.L.}, \bibinfo{editor}{Talat, Z.}, \bibinfo{editor}{Nozza, D.}, \bibinfo{editor}{Plaza-del Arco, F.M.}, \bibinfo{editor}{R{\"o}ttger, P.}, \bibinfo{editor}{Mostafazadeh~Davani, A.}, \bibinfo{editor}{Calabrese, A.} (Eds.), \bibinfo{booktitle}{Proceedings of the 8th Workshop on Online Abuse and Harms (WOAH 2024)}, \bibinfo{publisher}{Association for Computational Linguistics}. pp. \bibinfo{pages}{256--265}.
\newblock \DOIprefix\doi{10.18653/v1/2024.woah-1.20}.
\bibitem[{Ramaswamy et~al.(2023)Ramaswamy, Kim, Fong et~al.}]{RKFR23}
\bibinfo{author}{Ramaswamy, V.V.}, \bibinfo{author}{Kim, S.S.Y.}, \bibinfo{author}{Fong, R.}, et~al., \bibinfo{year}{2023}.
\newblock \bibinfo{title}{{Overlooked Factors in Concept-Based Explanations: Dataset Choice, Concept Learnability, and Human Capability}}, in: \bibinfo{booktitle}{Proc.\ 2023 {IEEE/CVF} Conference on Computer Vision and Pattern Recognition ({CVPR})}, \bibinfo{publisher}{{IEEE}}. pp. \bibinfo{pages}{10932--10941}.
\newblock \DOIprefix\doi{10.1109/CVPR52729.2023.01052}.
\bibitem[{Ribeiro et~al.(2018)Ribeiro, Singh and Guestrin}]{ribeiro2018anchors}
\bibinfo{author}{Ribeiro, M.T.}, \bibinfo{author}{Singh, S.}, \bibinfo{author}{Guestrin, C.}, \bibinfo{year}{2018}.
\newblock \bibinfo{title}{{Anchors: High-Precision Model-Agnostic Explanations}}, in: \bibinfo{booktitle}{Proc.\ 32nd AAAI Conference on Artificial Intelligence (AAAI), 13th Innovative Applications of Artificial Intelligence Conference (IAAI), and 8th AAAI Symposium on Educational Advances in Artificial Intelligence (EAAI)}, \bibinfo{publisher}{{AAAI} Press}. pp. \bibinfo{pages}{1527--1535}.
\newblock \DOIprefix\doi{10.1609/AAAI.V32I1.11491}.
\bibitem[{Risch et~al.(2021)Risch, Stoll, Wilms et~al.}]{wiegand2018overview}
\bibinfo{author}{Risch, J.}, \bibinfo{author}{Stoll, A.}, \bibinfo{author}{Wilms, L.}, et~al., \bibinfo{year}{2021}.
\newblock \bibinfo{title}{{Overview of the {G}erm{E}val 2021 Shared Task on the Identification of Toxic, Engaging, and Fact-Claiming Comments}}, in: \bibinfo{booktitle}{Proc.\ {GermEval} 2021 Shared Task on the Identification of Toxic, Engaging, and Fact-Claiming Comments}, \bibinfo{publisher}{Association for Computational Linguistics}. pp. \bibinfo{pages}{1--12}.
\bibitem[{Roy et~al.(2023)Roy, Harshvardhan, Mukherjee et~al.}]{RHMS2023}
\bibinfo{author}{Roy, S.}, \bibinfo{author}{Harshvardhan, A.}, \bibinfo{author}{Mukherjee, A.}, et~al., \bibinfo{year}{2023}.
\newblock \bibinfo{title}{{Probing {LLM}s for hate speech detection: strengths and vulnerabilities}}, in: \bibinfo{editor}{Bouamor, H.}, \bibinfo{editor}{Pino, J.}, \bibinfo{editor}{Bali, K.} (Eds.), \bibinfo{booktitle}{Findings of the Association for Computational Linguistics: EMNLP 2023}, \bibinfo{publisher}{Association for Computational Linguistics}. pp. \bibinfo{pages}{6116--6128}.
\newblock \DOIprefix\doi{10.18653/v1/2023.findings-emnlp.407}.
\bibitem[{Saleh et~al.(2023)Saleh, Alhothali and and}]{Saleh31122023}
\bibinfo{author}{Saleh, H.}, \bibinfo{author}{Alhothali, A.}, \bibinfo{author}{and, K.M.}, \bibinfo{year}{2023}.
\newblock \bibinfo{title}{{Detection of Hate Speech using BERT and Hate Speech Word Embedding with Deep Model}}.
\newblock \bibinfo{journal}{Applied Artificial Intelligence} \bibinfo{volume}{37}, \bibinfo{pages}{2166719}.
\newblock \DOIprefix\doi{10.1080/08839514.2023.2166719}.
\bibitem[{Sen et~al.(2024)Sen, Das and Sen}]{sen2024hatetinyllm}
\bibinfo{author}{Sen, T.}, \bibinfo{author}{Das, A.}, \bibinfo{author}{Sen, M.}, \bibinfo{year}{2024}.
\newblock \bibinfo{title}{{HateTinyLLM : Hate Speech Detection Using Tiny Large Language Models}}.
\newblock \bibinfo{journal}{CoRR} \bibinfo{volume}{abs/2405.01577}.
\newblock \DOIprefix\doi{10.48550/ARXIV.2405.01577}.
\bibitem[{Slack et~al.(2020)Slack, Hilgard, Jia et~al.}]{slack2020fooling}
\bibinfo{author}{Slack, D.}, \bibinfo{author}{Hilgard, S.}, \bibinfo{author}{Jia, E.}, et~al., \bibinfo{year}{2020}.
\newblock \bibinfo{title}{{Fooling lime and shap: Adversarial attacks on post hoc explanation methods}}, in: \bibinfo{booktitle}{Proceedings of the AAAI/ACM Conference on AI, Ethics, and Society}, pp. \bibinfo{pages}{180--186}.
\bibitem[{Sun et~al.(2024)Sun, Oikarinen, Ustun et~al.}]{SunOUW24}
\bibinfo{author}{Sun, C.}, \bibinfo{author}{Oikarinen, T.P.}, \bibinfo{author}{Ustun, B.}, et~al., \bibinfo{year}{2024}.
\newblock \bibinfo{title}{{Concept Bottleneck Large Language Models}}.
\newblock \bibinfo{journal}{CoRR} \bibinfo{volume}{abs/2412.07992}.
\newblock \DOIprefix\doi{10.48550/ARXIV.2412.07992}.
\bibitem[{Tan et~al.(2024)Tan, Cheng, Wang et~al.}]{TanCWYLL24}
\bibinfo{author}{Tan, Z.}, \bibinfo{author}{Cheng, L.}, \bibinfo{author}{Wang, S.}, et~al., \bibinfo{year}{2024}.
\newblock \bibinfo{title}{{Interpreting Pretrained Language Models via Concept Bottlenecks}}, in: \bibinfo{booktitle}{Proc.\ 28th Pacific-Asia Conference on Knowledge Discovery and Data Mining ({PAKDD})}, \bibinfo{publisher}{Springer}. pp. \bibinfo{pages}{56--74}.
\newblock \DOIprefix\doi{10.1007/978-981-97-2259-4_5}.
\bibitem[{Tjuatja et~al.(2024)Tjuatja, Chen, Wu et~al.}]{TjuatjaCWTN24}
\bibinfo{author}{Tjuatja, L.}, \bibinfo{author}{Chen, V.}, \bibinfo{author}{Wu, T.}, et~al., \bibinfo{year}{2024}.
\newblock \bibinfo{title}{{Do LLMs Exhibit Human-like Response Biases? A Case Study in Survey Design}}.
\newblock \bibinfo{journal}{Transactions of the Association for Computational Linguistics} \bibinfo{volume}{12}, \bibinfo{pages}{1011--1026}.
\newblock \DOIprefix\doi{10.1162/tacl_a_00685}.
\bibitem[{Touvron et~al.(2023)Touvron, Martin, Stone et~al.}]{touvron2023llama}
\bibinfo{author}{Touvron, H.}, \bibinfo{author}{Martin, L.}, \bibinfo{author}{Stone, K.R.}, et~al., \bibinfo{year}{2023}.
\newblock \bibinfo{title}{{{Llama} 2: Open Foundation and Fine-Tuned Chat Models}}.
\newblock \bibinfo{journal}{ArXiv} \bibinfo{volume}{abs/2307.09288}, \bibinfo{pages}{1--77}.
\newblock \URLprefix \url{https://api.semanticscholar.org/CorpusID:259950998}.
\bibitem[{Vidgen et~al.(2020)Vidgen, Hale, Guest et~al.}]{vidgen-etal-2020-detecting}
\bibinfo{author}{Vidgen, B.}, \bibinfo{author}{Hale, S.}, \bibinfo{author}{Guest, E.}, et~al., \bibinfo{year}{2020}.
\newblock \bibinfo{title}{{Detecting {E}ast {A}sian Prejudice on Social Media}}, in: \bibinfo{booktitle}{Proc.\ 4th Workshop on Online Abuse and Harms}, \bibinfo{publisher}{Association for Computational Linguistics}. pp. \bibinfo{pages}{162--172}.
\newblock \DOIprefix\doi{10.18653/v1/2020.alw-1.19}.
\bibitem[{Wang et~al.(2020)Wang, Liu, Ouyang et~al.}]{WangLOS20}
\bibinfo{author}{Wang, S.}, \bibinfo{author}{Liu, J.}, \bibinfo{author}{Ouyang, X.}, et~al., \bibinfo{year}{2020}.
\newblock \bibinfo{title}{{Galileo at SemEval-2020 Task 12: Multi-lingual Learning for Offensive Language Identification Using Pre-trained Language Models}}, in: \bibinfo{booktitle}{Proc.\ 14th Workshop on Semantic Evaluation ({SemEval@COLING})}, \bibinfo{publisher}{International Committee for Computational Linguistics}. pp. \bibinfo{pages}{1448--1455}.
\newblock \DOIprefix\doi{10.18653/V1/2020.SEMEVAL-1.189}.
\bibitem[{Wiegreffe and Pinter(2019)}]{jain2019attention}
\bibinfo{author}{Wiegreffe, S.}, \bibinfo{author}{Pinter, Y.}, \bibinfo{year}{2019}.
\newblock \bibinfo{title}{{Attention is not not Explanation}}, in: \bibinfo{editor}{Inui, K.}, \bibinfo{editor}{Jiang, J.}, \bibinfo{editor}{Ng, V.}, \bibinfo{editor}{Wan, X.} (Eds.), \bibinfo{booktitle}{Proceedings of the 2019 Conference on Empirical Methods in Natural Language Processing and the 9th International Joint Conference on Natural Language Processing (EMNLP-IJCNLP)}, \bibinfo{publisher}{Association for Computational Linguistics}. pp. \bibinfo{pages}{11--20}.
\newblock \DOIprefix\doi{10.18653/v1/D19-1002}.
\bibitem[{Wilming et~al.(2024)Wilming, Dox, Schulz, Oliveira, Clark and Haufe}]{wilming2024gecobenchgendercontrolledtextdataset}
\bibinfo{author}{Wilming, R.}, \bibinfo{author}{Dox, A.}, \bibinfo{author}{Schulz, H.}, \bibinfo{author}{Oliveira, M.}, \bibinfo{author}{Clark, B.}, \bibinfo{author}{Haufe, S.}, \bibinfo{year}{2024}.
\newblock \bibinfo{title}{Gecobench: {A} gender-controlled text dataset and benchmark for quantifying biases in explanations}.
\newblock \bibinfo{journal}{CoRR} \bibinfo{volume}{abs/2406.11547}.
\newblock \DOIprefix\doi{10.48550/ARXIV.2406.11547}, \href{http://arxiv.org/abs/2406.11547}{{\tt arXiv:2406.11547}}.
\bibitem[{Yang et~al.(2023)Yang, Panagopoulou, Zhou et~al.}]{YangPZJCY23}
\bibinfo{author}{Yang, Y.}, \bibinfo{author}{Panagopoulou, A.}, \bibinfo{author}{Zhou, S.}, et~al., \bibinfo{year}{2023}.
\newblock \bibinfo{title}{{Language in a Bottle: Language Model Guided Concept Bottlenecks for Interpretable Image Classification}}, in: \bibinfo{booktitle}{Proc.\ 2023 {IEEE/CVF} Conference on Computer Vision and Pattern Recognition ({CVPR})}, \bibinfo{publisher}{{IEEE}}. pp. \bibinfo{pages}{19187--19197}.
\newblock \DOIprefix\doi{10.1109/CVPR52729.2023.01839}.
\bibitem[{Yu et~al.(2022)Yu, Blanco and Hong}]{yu-etal-2022-hate}
\bibinfo{author}{Yu, X.}, \bibinfo{author}{Blanco, E.}, \bibinfo{author}{Hong, L.}, \bibinfo{year}{2022}.
\newblock \bibinfo{title}{{Hate Speech and Counter Speech Detection: Conversational Context Does Matter}}, in: \bibinfo{booktitle}{Proc.\ 2022 Conference of the North American Chapter of the Association for Computational Linguistics: Human Language Technologies (NAACL-HLT)}, \bibinfo{publisher}{Association for Computational Linguistics}. pp. \bibinfo{pages}{5918--5930}.
\newblock \DOIprefix\doi{10.18653/v1/2022.naacl-main.433}.
\bibitem[{Zhou et~al.(2023)Zhou, Li, Li et~al.}]{zhou2023comprehensive}
\bibinfo{author}{Zhou, C.}, \bibinfo{author}{Li, Q.}, \bibinfo{author}{Li, C.}, et~al., \bibinfo{year}{2023}.
\newblock \bibinfo{title}{{A Comprehensive Survey on Pretrained Foundation Models: A History from {BERT} to {ChatGPT}}}.
\newblock \bibinfo{journal}{CoRR} \bibinfo{volume}{abs/2302.09419}, \bibinfo{pages}{1--99}.
\newblock \DOIprefix\doi{10.48550/ARXIV.2302.09419}, \href{http://arxiv.org/abs/2302.09419}{{\tt arXiv:2302.09419}}.

\end{thebibliography}

\clearpage
\appendix

\section{Selected Set of Adjectives $\adjSet$}
\label{app:lexicon_list}


\fontsize{6.4pt}{7.pt}\selectfont
\begin{longtable}{p{2.3cm}p{1.8cm}|p{1.6cm}p{1.8cm}|p{1.6cm}p{2.0cm}}
   \hline
\textbf{German} & \textbf{English}  & \textbf{German} & \textbf{English}& \textbf{German} & \textbf{English} \\\hline
beschimpfend & abusive & freundlich& friendly & rücksichtslos & ruthless  \\
akzeptierend & accepting  & frustriert& frustrated & sarkastisch   & sarcastic \\
entgegenkommend & accommodating   & lustig & funny & satirisch& satirical \\
bewundernd & admiring   & aggressiv  & aggressive & einfühlsam & sensitive \\
altersdiskriminierend   & age discriminatory & schadenfroh & gleefully& sensibilisiernd  & sensitizing \\
geschlechter- \newline diskriminierend & gender discriminatory &  amüsiert   & amused& sexistisch & sexist \\
gewaltverherrlichend   & glorifying violence  & belästigend & harassing& sexualisierend   & sexualizing \\
belustigend& amusing & schädlich & harmful  & beschämend & shameful  \\
gemeinschaftsstärkend & strength  community  & hasserfüllt & hateful  & schamlos & shameless \\
behindertenfeindlich & anti-disabled   & hilfsbereit & helpful  & scharfsinnig  & sharp  \\
antifeministisch   & anti-feminist   & homofeindlich  & homophobic & skeptisch& skeptical \\
antiziganistisch   & antigypsyist & anfeindend& hostile  & solidarisch   & solidaric \\
menschenfreundlich  & humane  & antisemitisch   & anti-Semitic & gehässig & spiteful  \\
religionsfeindlich & anti-religious  &  queerfeindlich  & anti-queer  & hassverbreitend  & spreading hate \\
minderheitenfeindlich  & hostile to minorities & erniedrigend   & humiliating   &  grausam & awful \\
anerkennend& appreciative & humorvoll & humorous & verärgert  & annoyed \\
argumentativ & argumentative   & verletzend& hurtful  & bitter& bitter  \\
konfliktschürend & stirring up conflict& ignorant  & ignorant & hartnäckig & stubborn  \\
ausbalancierend & balancing  & einwandsbereit & impeccable & unterstützend & supportive\\
herabsetzend & belittling & pietätlos & impious  & terrorisierend   & terrorizing \\
zurückschlagend  & striking back & unangemessen   & inappropriate & denkanstoßend & thought-provoking   \\
bodyshamend& body shaming & mutig & brave & bedrohend& threatening \\
gewaltaufrufend& inciting violence
& unverständlich & incomprehensible & tolerant & tolerant  \\
brückenbauend   & building bridges   & hetzerisch& inflammatory  & toxisch  & toxic  \\
schikanierend   & bullying   & informierend   & informative   & transfeindlich   & transphobic \\
unvoreingenommen & unbiased& informiert& informed & inakzeptabel  & unacceptable   \\
behutsam   & carefully  & unmenschlich   & inhuman  & gelassen   & calmly   \\
herausfordernd  & challenging& versöhnlich& conciliatory & unzivilisiert & uncivilized \\
aggressionsgeladen & charged with aggression & nachfragend & inquiring& verständnisvoll  & understanding  \\
chauvinistisch  & chauvinistic & unsensibel& insensitive   & unfair   & unfair \\
prüfend & checking   & heimtückisch   & insidious& unverzeihlich & unforgivable   \\
engagiert  & committed  & einsichtsvoll  & insightful & unnötig  & unnecessary \\
gemeinschaftsorientiert & community oriented & schmähend & insulting& unangenehm & unpleasant\\
menschenverachtend  & inhumane  & integrativ& integrative   & vorbehaltlos  & unreservedly   \\
beweiskräftig   & conclusive & einschüchternd & intimidating  & ungezügelt & unrestrained   \\
unverantwortlich & irresponsible  & intolerant& intolerant & unzufrieden   & unsatisfied \\
gewaltverurteilend & condemning violence& untersuchend   & investigating & aufgeregt& upset  \\
herablassend & condescending   &  verurteilend & condemning& stichhaltig   & valid  \\
konfliktreich   & conflictual& islamfeindlich & Islamophobic  & niederträchtig   & vile   \\
konfrontativ & confrontational & sachkundig& knowledgeable & gewaltbereit  & violent   \\
rücksichtsvoll  & considerate& ettiketierend  & labeling & vulgär   & vulgar \\
konstruktiv& constructive & objektiv  & lens& warmherzig & warm-hearted   \\
geringschätzend & contemptuous & bösartig  & malicious& abschwächend  & weakening \\
widersprechend  & contradictory   & vermittelnd & mediating& besorgt  & worried   \\
zuwiderlaufend  & contrary   & misanthropisch & misanthropic  & xenofeindlich & xenohostile \\
kontrastierend  & contrasting& misogyn   & misogynistic  & xenophob & xenophobic\\
umstritten & controversial   & spöttisch & mockingly& kreativ & creative \\
überzeugend& convincing & moderierend & moderating & provokativ & provocative \\
korrigierend & corrective & moralisierend  & moralizing & umsichtig& prudent   \\
toleranzfördernd & promoting tolerance    & motivierend & motivating & hinterfragend & questioning \\
kritisch   & critical   & negativ   & negative & rassistisch   & racist \\
grenzüberschreitend& cross-border & neutralisierend& neutralizing  & radikal  & radical   \\
vorurteilsabbauend & reducing prejudice   & gewaltfrei& non-violent   & rational & rational  \\
gefährlich & dangerous  & nuanzierend & nuanced  & räsonnierend  & reasoning \\
verleumderisch  & defamatory & obszön & obscene  &  neugierig  & curious\\
verteidigend & defensive  & kränkend  & offending& reflexiv & reflexive \\
degradierend & degrading  & anstößig  & offensive& widerlegend   & refuting  \\
entmenschlichend   & dehumanizing & aufgeschlossen & open minded   & ablehnend& rejecting \\
abwägend   & deliberative & unterdrückend  & oppressive & relativierend & relativizing   \\
demokratisch & democratically  & optimistisch   & optimistic & verwerflich   & reprehensible  \\
abfällig   & derogatory & empört & outraged & abstoßend& repulsive \\
verabscheuungswürdig & despicable & empörend  & outrageous & respektvoll   & respectful\\
zerstörerisch   & destructive& überdenkend & overthinking  & achtungsvoll  & respectfully   \\
entwertend & devaluing  & leidenschaftlich & passionate & respektierend & respecting\\
dialogorientiert   & dialogue-oriented  & passiv-aggressiv & passive-aggressive & umkehrend& reversing \\
meinungsverschieden& different opinions & erbärmlich& pathetic & revidierend   & revising  \\
differenzierend & differentiating & geduldig  & patient  & widerrufend   & revoking  \\
diplomatisch & diplomatic & friedlich & peaceful & grob& rough  \\
enttäuscht & disappointed & beharrlich& persistently  & aufklärerisch & enlightening   \\
diskreditierend & discreditable   & polemisch & polemical& gleichberechtigt & Equal rights   \\
diskriminierend & discriminatory  & höflich   & polite   & beweisführend & evidential\\
bedenkenausdrückend   & expressing concern & positiv   & positive & ausgrenzend   & exclusionary   \\
widerwärtig& disgusting & hasspredigend  & preaching hate   &  diskutierend & discussing  \\
aggressionsfördernd & promoting aggression  & respektlos & disrespectful    & ausdrucksstark   & expressive\\
vorurteilsbehaftet  & prejudiced & abweisend  & dismissive  & extremistisch & extremist \\
distanzierend   & distancing &   ablenkend  & distracting & faktenbasiert & fact-based\\
sich distanzierend & distancing yourself& dialogfördernd & promoting dialogue & faktenbezogen & factual   \\
unterscheidend  & distinctive& emotional & emotionally   & fair& fair   \\
demokratiefördernd  & promoting democracy & empathisch& empathetic & feministisch  & feminist  \\
emanzipatorisch  & emancipatory  & aufmunternd & encouraging   & effektiv & effectively \\
diversitätsbewusst & diversity conscious& erbaulich & edifying & störend & disturbing   \\
spaltend   & divisive   &   &  &  &  \\ \hline   
\end{longtable}
\normalsize
%

\section{Prompt decomposition for KV pre-computation}\label{app:prefix+suffix}

Below is an example of prompt splitting to accelerate the computation of probability values across combinations of text samples and adjectives ($t_i$, $a_j$). In grey, the prefix (invariant part) is highlighted, and the remaining corresponds to the suffix. We store and reuse the key (K) and value (V) submatrices of the transformer blocks for the prefix to save inference time.

\begin{tcolorbox}[colback=white, colframe=mygray, boxrule=0.7pt, title={Prompt splitting for $\Pr(a_j\mid t_i,\LLM)$ computation in ELF22.}, coltitle=black]
 \redhighlight{[INST] <<SYS>> \\
 You are an expert in social psychology. When you are asked a question, you prefer to give short, concrete answers\\
 <</SYS>>\\
 Consider carefully the following conversation between person ``A'' and ``B''.\\
 A:} \textbf{X}\\
 B: \textbf{Y}\\
 Tell me if the response of B seems to be ``\textbf{adjective}''. [/INST]
\end{tcolorbox}

\section{Prompting templates for concepts evaluation}\label{app:templatep}
\tableautorefname~\ref{tab:prompt_template}  shows the prompt templates employed for each of the explored datasets in our experiments. 
\begin{table}[h!]
    \fontsize{8.3pt}{9pt}\selectfont
    \centering
    \caption{Prompt templates employed for each task. Text in bold, \ie~ \textit{X, Y, adjective}, correspond to placeholders changing for every combination of $t_i$ and $a_j$ on each dataset.}
    \label{tab:prompt_template}
    \begin{tblr}{
                  colspec={p{1.2cm}|p{5.0cm}|p{5.5cm}},
                  hline{1,3, 4, 5, 6},
                  cell{3}{2} = {r=3}{m}, 
                }
         \textbf{Dataset}& \textbf{Persona} & \textbf{Interaction}  \\
         \hline
         GermEval&Sie sind ein Experte für Sozialpsychologie.  Wenn Ihnen eine Frage gestellt wird, geben Sie lieber kurze und konkrete Antworten.&Überlege genau und sage mir, ob das Adjektiv ``\textbf{adjective}'' den folgenden Satz in irgendeiner Weise beschreibt: ``\textbf{X}''.\\
         
         ELF22 & You are an expert in social psychology. When you are asked a question, you prefer to give short, concrete answers & {Consider carefully the following conversation between person ``A'' and ``B''.\\ A: \textbf{X} \\B: \textbf{Y}\\ Tell me if the response of B seems to be ``\textbf{adjective}''}.\\
         
         HS-CS \& \newline CONAN&& {Consider what a person ``A'' states around a ``CONTEXT''. \\CONTEXT: \textbf{X} \\ ``A'': \textbf{Y} \\ Tell me if the response of ``A'' seems to be ``\textbf{adjective}''.}\\
         
         TSNH&&{Tell me if adjective ``\textbf{adjective}'' describes the content below:\\``\textbf{X}''.}\\
    \end{tblr}
\end{table}     

\section{Definition of zero-shot prompting templates}\label{app:template_zeroshot}

\noindent
\revised{\paratitle{Zero-shot.}}
\revised{\tableautorefname~\ref{tab:prompt_template_zero} shows the prompt templates employed 
for zero-shot experiments (GPT 3.5, GPT 4o, Llama 3.1 from Category IV in \tableautorefname~\ref{tab:class_result}). For each dataset, the persona in the prompt templates is the same as in \tableautorefname~\ref{tab:prompt_template} to ensure comparability. The rest of the prompt (interaction) is adapted to suit the respective task of each dataset, \eg~by modifying the classes and including relevant context.} 
\begin{table}[!h]
    \fontsize{9pt}{10pt}\selectfont
    \centering
    \caption{Prompt templates for the zero-shot evaluation. 
    Elements highlighted with bold letters, \ie~ \textit{X, Y, adjective}, correspond to variables for every instance in each dataset.}
    \label{tab:prompt_template_zero}
    \begin{tblr}{
                  colspec={p{2cm}|p{9cm}},
                  hline{1,2,3, 4,5,6,7},
                  cell{2}{1} = {r=1}{m},
                }
         \textbf{Dataset} & \textbf{Interaction}  \\
         GermEval&Überlege genau und sage mir, ob das Wort ``Offensiv'' den folgenden Satz in irgendeiner Weise beschreibt: ``\textbf{X}''.\\
         
         ELF22 &{Consider the following conversation between person ``A'' and ``B'' carefully. \\ A: \textbf{X} \\B: \textbf{Y}\\ Tell me if the response of B seems to be trolling}.\\
         
         HS-CS&{Consider what a person ``A'' states around a ``CONTEXT''. \\CONTEXT: \textbf{X} \\ ``A'': \textbf{Y} \\ Tell me to which one of the following categories [ ``counter speech'', ``hate speech'', ``neither''] it corresponds.}\\

         CONAN& {Consider what a person ``A'' states around a ``CONTEXT''. \\CONTEXT: \textbf{X} \\ ``A'': \textbf{Y} \\ Tell me to which one of the following categories [ ``hypocrisy'', ``denouncing'', ``question'', ``unrelated'', ``humor'', ``facts'', ``support''] it corresponds.}\\
         
         TSNH&{Think carefully and tell me if the following sentence is counterspeech:\\``\textbf{X}''.}\\
    \end{tblr}
\end{table}    

\revised{The templates above defined are additionally employed when experimenting with the model with reasoning capabilities, i.e., GPT o3-mini (CoT) from Category IV in \tableautorefname~\ref{tab:class_result}}. 

\paratitle{In-context learning (ICL).} 
\revised{To realize our experiments using in-context learning, i.e., GPT 4o (ICL) from Category IV in \tableautorefname~\ref{tab:class_result}, we prompt the model with five examples for each class. These examples were randomly sampled following a uniform distribution and excluding those that exceeded 180 words to avoid overloading the prompt with information that could lead to contextual dilution.}

\revised{\tableautorefname~\ref{box:contexticl} shows the structure of the ICL prompt employed in those datasets where context is provided, i.e., ELF22, HS-CS, and CONAN, as well as in those without context, i.e., TSNH and GermEval. Light gray text is replaced with the information from the actual examples used during evaluation, and \redhighlight{CLASSES} is replaced by the list of classes corresponding to the explored dataset.}
\revised{
\begin{table}[!h]
    \centering
    \caption{\revised{Prompt example for ICL in our experimental setup. \redhighlight{CLASSES} represents a variable that is replaced with the list of classes corresponding to the explored datasets.}}
    \label{box:contexticl}
    \begin{tabular}{p{\textwidth}}
        \hline
        \multicolumn{1}{c}{\textbf{Template for datasets with context}} \\
        \hline
        You are an expert in social psychology. \\
        Always read the CONTEXT before the COMMENT and decide whether the comment is \redhighlight{CLASSES}. Reply with the single label — no extra words. \\
        \\
        \textbf{EXAMPLES} \\
        \texttt{<text\_icl\_begin>} \\
        CONTEXT: \textcolor{gray}{It is so that women ... }\\
        COMMENT: \textcolor{gray}{Simply pointing out that women ...} \\
        \texttt{<text\_icl\_end>} \\
        LABEL: \textcolor{gray}{POSITIVE} \\
        ... \\
        \textbf{END EXAMPLES}\\\\
        \hline
        \multicolumn{1}{c}{\textbf{Template for datasets without context}} \\
        \hline
        You are an expert in social psychology. \\
        Decide whether the TEXT is \redhighlight{CLASSES}. Reply with the single label — no extra words. \\
        \\
        \textbf{EXAMPLES} \\
        \texttt{<text\_icl\_begin>} \\
        TEXT: \textcolor{gray}{Simply pointing out that women ...} \\
        \texttt{<text\_icl\_end>} \\
        LABEL: \textcolor{gray}{POSITIVE} \\
        ... \\
        \textbf{END EXAMPLES} \\
        \hline
    \end{tabular}
\end{table}
}
\section{Local explanations from $\HSCBM$-R}\label{app:lime}

\tableautorefname~\ref{tab:local_exp_ext} provides additional local explanation results to \tableautorefname~\ref{tab:instances_descriptor}. It shows the top-valued adjectives provided by \HSCBM-R for predictions of individual samples from the CS-HS dataset, together with a comparison to LIME explanations for the XML-RoBERTa model. In LIME explanations colour encodes the relative relevance (stronger saturation means higher relevance) of words according to the LIME method.

\begin{table}[h!]
    \fontsize{7pt}{8pt}\selectfont
    \centering
    \caption{
    Top-weighted adjectives for individual input samples from each class of the HS-CS dataset provided by \HSCBM-R.}
    \label{tab:local_exp_ext}
    \begin{tblr}{
                  colspec={Q[t, 3.8em]|p{6.7cm}|p{4.5cm}},
                  hline{1,3, 4, 5, 6, 7, 8},
                  cell{2}{1} = {r=2}{m}, 
                  cell{4}{1} = {r=2}{m}, 
                  cell{6}{1} = {r=2}{m}, 
                }
         \textbf{Class}& \textbf{Input} & \textbf{Adjectives}  \\
         \hline
         Counter Speech& {CONTEXT: \colorbox{cs_color!2}{What} \colorbox{cs_color!40}{would} \colorbox{cs_color!30}{be} \colorbox{cs_color!50}{the} difference \colorbox{cs_color!15}{between} \colorbox{cs_color!50}{fishes} and birds \colorbox{cs_color!30}{in this} case?  \\
         COMMENT: \colorbox{cs_color!30}{Birds} eat fish. \colorbox{cs_color!5}{Fish} \colorbox{cs_color!5}{don}’\colorbox{cs_color!10}{t} eat birds }& {\fontsize{7pt}{11pt}\selectfont refuting, differentiating, \newline informative, evidential, \newline discussing, promoting dialogue,\newline exclusionary, humorous,\newline weakening, provocative} \\
         & {CONTEXT: \colorbox{cs_color!25}{Except} testicles are required \colorbox{cs_color!15}{for procreation} and foreskin \colorbox{cs_color!15}{is not}. Bad comparison. \\
         COMMENT: foreskin \colorbox{cs_color!15}{is} \colorbox{cs_color!10}{required} to have an \colorbox{cs_color!15}{actual} \colorbox{cs_color!50}{orgasm}.  \colorbox{cs_color!15}{procreation is} unnecessary.}&{\fontsize{7pt}{11pt}\selectfont refuting, exclusionary,  dialogue-oriented, differentiating, \newline incomprehensible, reversing,\newline different opinions, relativizing, \newline conflictual, moralizing}\\
         Hate Speech & {CONTEXT: Sorry before i have sex \colorbox{hs_color!10}{with women} i get them to sign contracts. \colorbox{hs_color!10}{Not} like you \colorbox{hs_color!40}{dumbfucks}.  \\
         COMMENT: \colorbox{hs_color!10}{Lmao} found the \colorbox{hs_color!25}{virigin}. } & {\fontsize{7pt}{11pt}\selectfont disturbing, discreditable, \newline hateful, vulgar,\newline irresponsible, threatening,\newline defamatory, intolerant,\newline sarcastic, argumentative}\\
         & {CONTEXT: \colorbox{hs_color!15}{Oh}, my \colorbox{hs_color!15}{God} not Piers \colorbox{hs_color!15}{fucking} Morgan. \colorbox{hs_color!25}{He}'s \colorbox{hs_color!15}{such a} \colorbox{hs_color!10}{c}**t.   \\
         COMMENT: \colorbox{hs_color!25}{He lacks} \colorbox{hs_color!15}{the} \colorbox{hs_color!10}{depth} \colorbox{hs_color!15}{and} warmth to be {a} \colorbox{hs_color!40}{cunt}. } & {\fontsize{7pt}{11pt}\selectfont passive-aggressive, rejecting,\newline devaluing, labeling,\newline hostile, polemical,\newline prejudiced, contrasting,\newline questioning, sarcastic}\\
         
         Neutral Speech& {CONTEXT: \colorbox{ns_color!15}{Shitting} on \colorbox{ns_color!40}{women} = \colorbox{ns_color!5}{men}'s \colorbox{ns_color!15}{rights} \\
         COMMENT: \colorbox{ns_color!35}{Satirizing} \colorbox{ns_color!5}{feminist} \colorbox{ns_color!50}{bullshit} =/= \colorbox{ns_color!15}{Shitting} on \colorbox{ns_color!40}{women}}&{\fontsize{7pt}{11pt}\selectfont inappropriate, exclusionary,\newline toxic, polemical,\newline conflictual, harmful,\newline labeling, dismissive,\newline gender discriminatory, contradictory}\\
         & {CONTEXT: \colorbox{ns_color!10}{He}'s had dozens of \colorbox{white}{women} come forward \colorbox{white}{giving} correlating \colorbox{white}{accounts} of fucked up, \colorbox{ns_color!50}{rapist} \colorbox{ns_color!15}{behaviour}.  \colorbox{ns_color!25}{Cosby} had \colorbox{ns_color!15}{the} same \colorbox{ns_color!15}{thing}.  \colorbox{ns_color!30}{Trump} is \colorbox{ns_color!15}{a} \colorbox{ns_color!50}{rapist}/\colorbox{ns_color!40}{violator}.  
  \\
         \colorbox{white}{COMMENT}: There have been anonymous accusations \colorbox{ns_color!15}{but} no one has \colorbox{ns_color!15}{come} forward. } & {\fontsize{7pt}{11pt}\selectfont exclusionary,	refuting,\newline rejecting, moralizing,\newline	worried, revising,\newline challenging, reflexive,\newline expressive,	sceptical}\\
    \end{tblr}
\end{table}  
\bigskip
\newpage

\end{document}